\documentclass{report}
\usepackage[final,nonatbib]{style/mystyle}
\usepackage{natbib}
\usepackage[utf8]{inputenc}
\usepackage{paralist}
\usepackage[dvipsnames]{xcolor}
\usepackage{url}
\usepackage{graphicx}
\usepackage{latexsym}
\usepackage{amsmath,bm}
\usepackage{hyperref}
\usepackage{amsmath}
\usepackage{tikz}
\usepackage[disable]{todonotes}
\usetikzlibrary{fadings}
\usetikzlibrary{mindmap,trees}
\usetikzlibrary{arrows,automata,shapes,positioning,shadows,trees}
\usetikzlibrary{shapes,snakes,shadows}
\usetikzlibrary{shapes.arrows}

\usetikzlibrary{calc,shapes, positioning}
\usetikzlibrary{decorations.text}
\usetikzlibrary{matrix,chains,positioning,decorations.pathreplacing,arrows}
\usetikzlibrary{bayesnet}
\usepackage[edges]{forest}
\usetikzlibrary{shadows.blur}
\usetikzlibrary{shapes.geometric}

\usepackage{pgfplots}
\usepackage{pgfplotstable}
\usepackage{pgfmath,pgffor}
\usepgfplotslibrary{colorbrewer}
\pgfplotsset{compat = 1.14, cycle list/Set1-8}
\usetikzlibrary{pgfplots.statistics, pgfplots.colorbrewer}
\pgfplotsset{compat=1.8}

\tikzstyle{edge}=[-latex',draw=black!90,shorten <=1pt,shorten >=1pt]
\tikzstyle{redge}=[latex'-,draw=black!90,shorten <=1pt,shorten >=1pt]
\tikzstyle{dedge}=[latex'-latex',draw=black!90,shorten <=1pt,shorten >=1pt]

\tikzstyle{block}=[draw, text width=5em,align=center,shape=rectangle, rounded corners, , align=center]
\tikzstyle{nobox}=[align=center]
\definecolor{emb}{RGB}{209,228,252}
\definecolor{hidden-blue}{RGB}{194,232,247}
\definecolor{hidden-orange}{RGB}{243,202,120}
\definecolor{hidden-yellow}{RGB}{242,244,193}
\definecolor{output-purple}{RGB}{219,203,231}
\definecolor{output-green}{RGB}{204,231,207}
\definecolor{hiddendraw}{RGB}{205, 44, 36}
\usepackage{amssymb}
\usepackage{array,multirow}
\usepackage{bbding}
\usepackage{pifont}
\usepackage{wasysym}
\usepackage{wrapfig}
\usepackage{subfigure}
\usepackage{caption}
\usepackage{mwe}
\usepackage{amsmath}
\usepackage{amssymb}
\usepackage{wasysym}
\usepackage{bbm}
\usepackage{CJKutf8}
\usepackage{mathtools}
\usepackage{makecell}

\DeclarePairedDelimiter\floor{\lfloor}{\rfloor}

\newcommand{\bleu}[1]{ \texttt{BLEU}}
\newcommand{\modelname}[1]{ \textsc{PlotMachines}}
\newcommand{\modelnameshort}[1]{\textsc{PM}}
\newcommand{\singlemem}[1]{\textsc{PM-Single}}
\newcommand{\dualmem}[1]{\textsc{PM-Full}}
\newcommand{\nomem}[1]{\textsc{PM-NoMem}}
\newcommand{\paratitle}[1]{\noindent\textbf{#1}\ \ }

\newcommand{\grover}[1]{ \textsc{Grover}}
\newcommand{\taskname}[1]{outline-conditioned story generation}
\makeatletter
\newcommand{\printfnsymbol}[1]{%
  \textsuperscript{\@fnsymbol{#1}}%
}

 \usepackage{algorithm}
 \usepackage{algorithmic}

\makeatother

\title{A Survey on Green Deep Learning}

 \author{Jingjing Xu\thanks{Equal Contribution}\\
ByteDance AI Lab \\
 \small{\texttt{xujingjing.melody@bytedance.com}}\\
 \small{\texttt{jingjingxu@pku.edu.cn}}\\
 \And
 Wangchunshu Zhou\footnotemark[1]\\
 ByteDance AI Lab  \\
 \small{\texttt{
zhouwangchunshu.7@bytedance.com}}\\
 \And
 Zhiyi Fu\footnotemark[1] \ \thanks{This work is done during internship at ByteDance AI Lab.} \\
 \ \ Peking University \ \ \\
 \small{\texttt{
ypfzy@pku.edu.cn}}
\And
 Hao Zhou \\
 ByteDance AI Lab \ \  \\
 \small{\texttt{
zhouhao.nlp@bytedance.com}} 
\And
 Lei Li \\
 University of California, Santa Barbara \\
 \small{\texttt{
lilei@ucsb.edu}}
 }
\date{}

\begin{document}
\maketitle
\begin{abstract}
In recent years, larger and deeper models are springing up and continuously pushing state-of-the-art (SOTA) results across various fields like natural language processing (NLP) and computer vision (CV). However, despite promising results, it needs to be noted that the computations required by SOTA models have been increased at an exponential rate. Massive computations not only have a surprisingly large carbon footprint but also have negative effects on research inclusiveness and deployment on real-world applications.

Green deep learning is an increasingly hot research field that appeals to researchers to pay attention to energy usage and carbon emission during model training and inference. The target is to yield novel results with lightweight and efficient technologies.  Many technologies can be used to achieve this goal, like model compression and knowledge distillation.  This paper focuses on presenting a systematic review of the development of Green deep learning technologies. We classify these approaches into four categories:  (1) compact networks, (2) energy-efficient training strategies, (3) energy-efficient inference approaches, and (4) efficient data usage. For each category, we discuss the progress that has been achieved and the unresolved challenges. 

\end{abstract}

\clearpage

\tableofcontents

\newpage
\chapter{Introduction}
\label{intro}

 Deep learning, based on deep neural networks, is part of machine learning methods. In this chapter, we first introduce the development of deep learning in section~\ref{sec:background}. Then, we elucidate what is Green deep learning, why Green deep learning matters, and how to evaluate the ``greenness'' of deep learning in section~\ref{sec:greenlearning}. 

 \section{Deep Learning}
 \label{sec:background}
 
 While a decade ago, artificial intelligence (AI) mainly focuses on shallow models, like structure perceptrons~\citep{DBLP:conf/naacl/McDonaldHM10,DBLP:conf/naacl/HuangFG12,DBLP:conf/acl/LiJ14} and conditional random fields~\citep{DBLP:conf/ijcnlp/GhoshJRT11,DBLP:journals/ftml/SuttonM12,DBLP:conf/iccv/0001JRVSDHT15}. These shallow models only require limited computations. Most AI approaches can be deployed on CPUs.   
 
 In recent years, powerful GPUs have become increasingly accessible, making it possible to deploy larger models, which accelerates the development of deep learning. The ideas of widely-used deep learning models have been proposed in the 1990s, such as convolutional neural networks (CNNs)~\citep{lecun1998gradient} and long-short term networks (LSTMs)~\citep{DBLP:journals/neco/HochreiterS97}. Confined by hardware capacity and large-scale data resources, these models began to be popular until the past few years.  \citet{DBLP:journals/jmlr/CollobertWBKKK11} proposed the first systematic deep learning framework for NLP tasks.  \citet{DBLP:conf/nips/KrizhevskySH12} proposed a convolution-based deep network, which ranked the first  in the image classification challenge. These studies are good pioneers that motivate AI participants to dive into deep learning. 

\begin{figure}[t]
    \centering
       
    \includegraphics[width=0.5\textwidth]{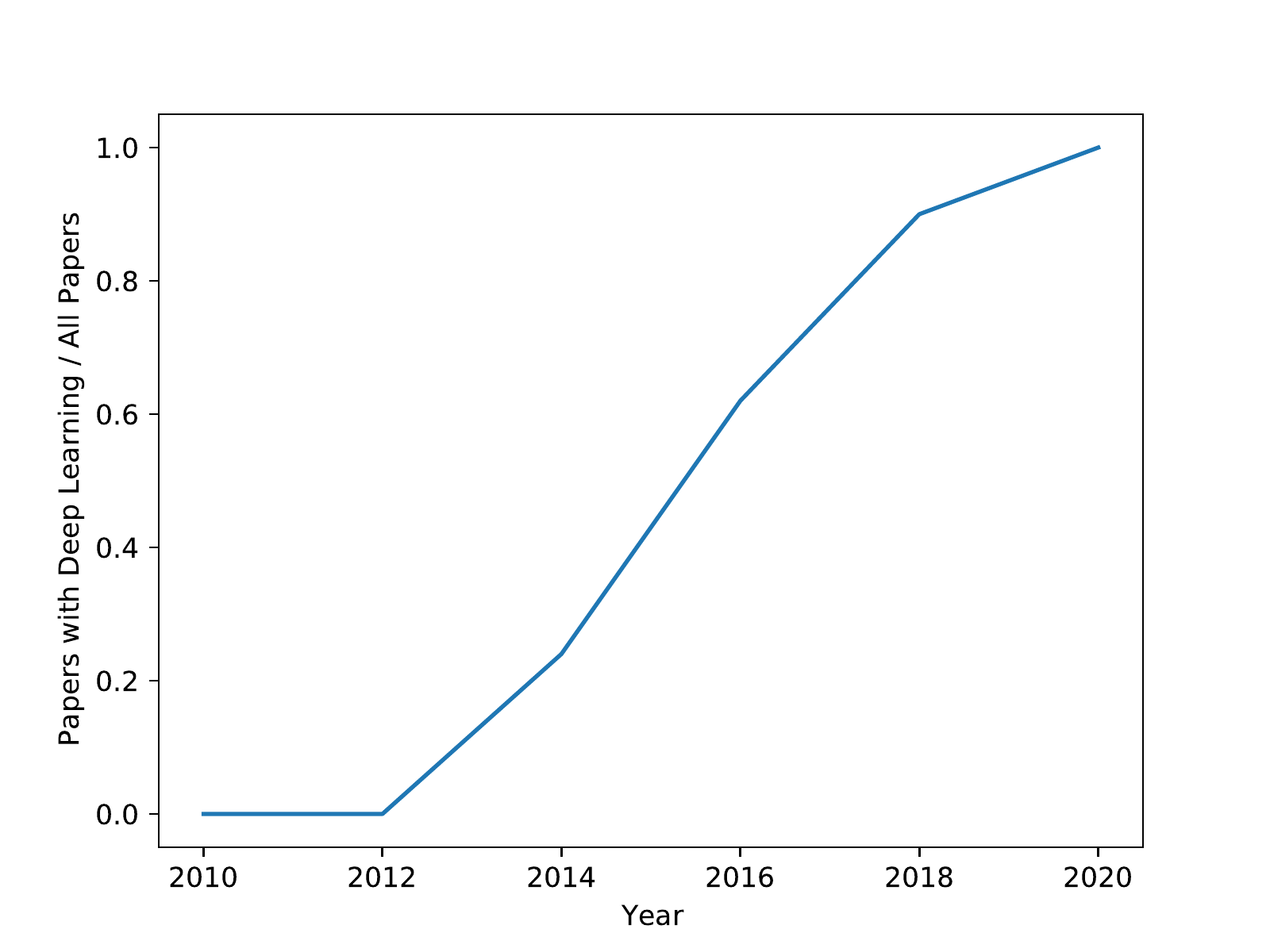}
    \caption{The ratio of papers using neural networks in ACL, a top-tier NLP conference, from 2010 to 2020. We manually count how many papers use neural networks among sampled 50 papers. }\label{fig:deeplearning}
\end{figure}
 
Deep learning methods currently have become a prime choice for AI.
The promising prospect of neural networks attracts more  AI participants to engage with deep learning, in the meanwhile, deep learning is highly competitive in yielding profits when applied to real-world applications. The industry continuously develops more efficient hardware and launches better programming platforms, such as Theano, Caffe, MxNet, Tensorflow, and Pytorch.  Advanced infrastructures further enable AI participants to develop stronger deep models. Therefore, deep learning takes a high-speed train since 2010. To visualize the transition process from shallow models to deep networks in the AI community,  we analyze papers from a top-tier AI conference, ACL, starting from 2010 to 2020. We randomly select 50 works from each year and manually count the number of papers using deep learning. As we can see from Figure~\ref{fig:deeplearning}, the number of papers using neural networks is growing fast from 2012 to 2020.  All papers adopt deep networks as backbone since 2020. From then, the AI field is fully entered into the age of deep learning. 

\begin{figure}[b]
    \centering
    \includegraphics[width=0.7\textwidth]{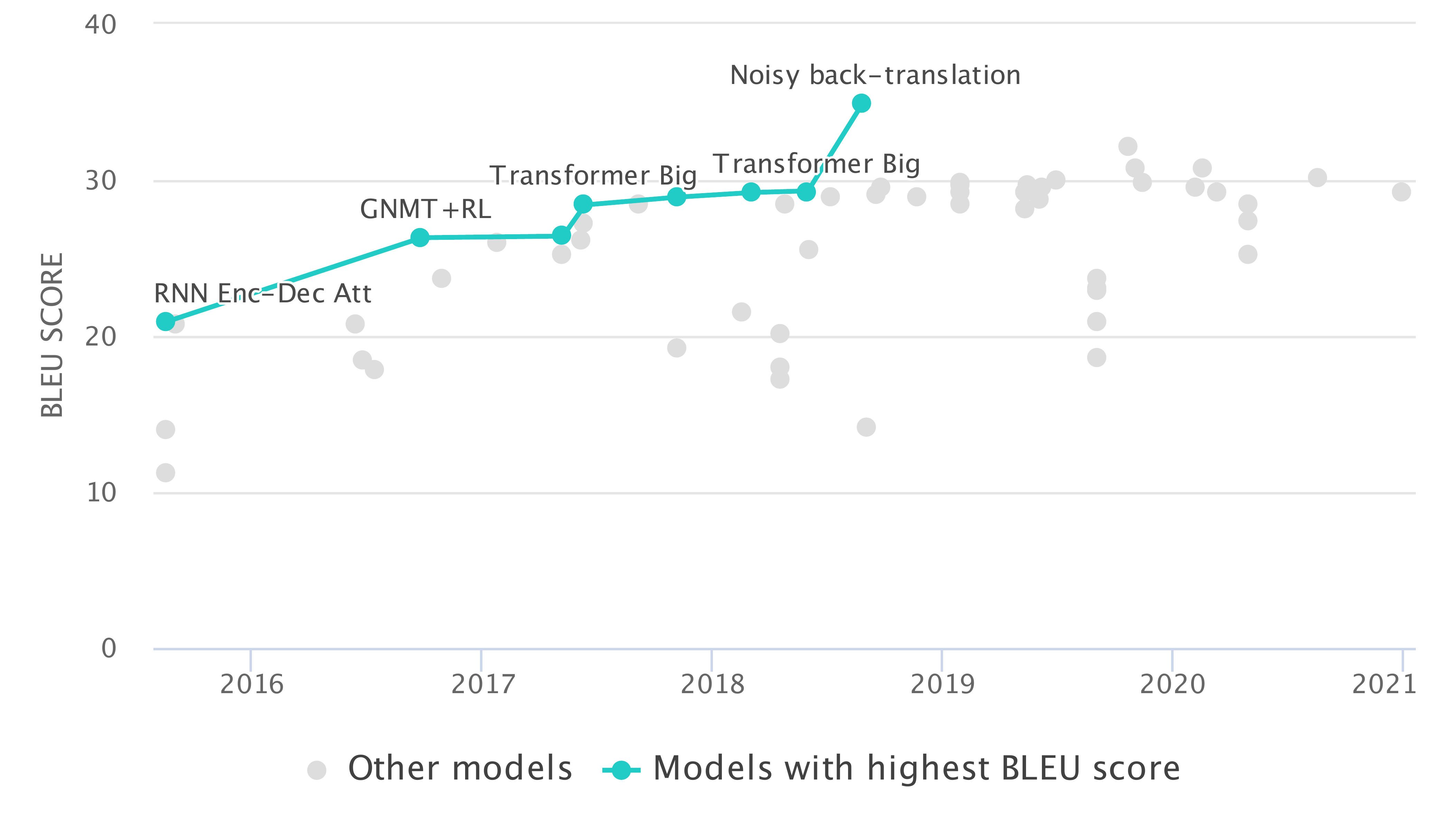}
    \caption{Results of WMT English-German translation from 2016 to 2020.  As we can see, the recent published results are reaching into the ceiling. The data is collected from \url{https://paperswithcode.com}. }
    \label{fig:ende}
\end{figure}

\begin{figure}[t]
    \centering
    \includegraphics[width=\textwidth]{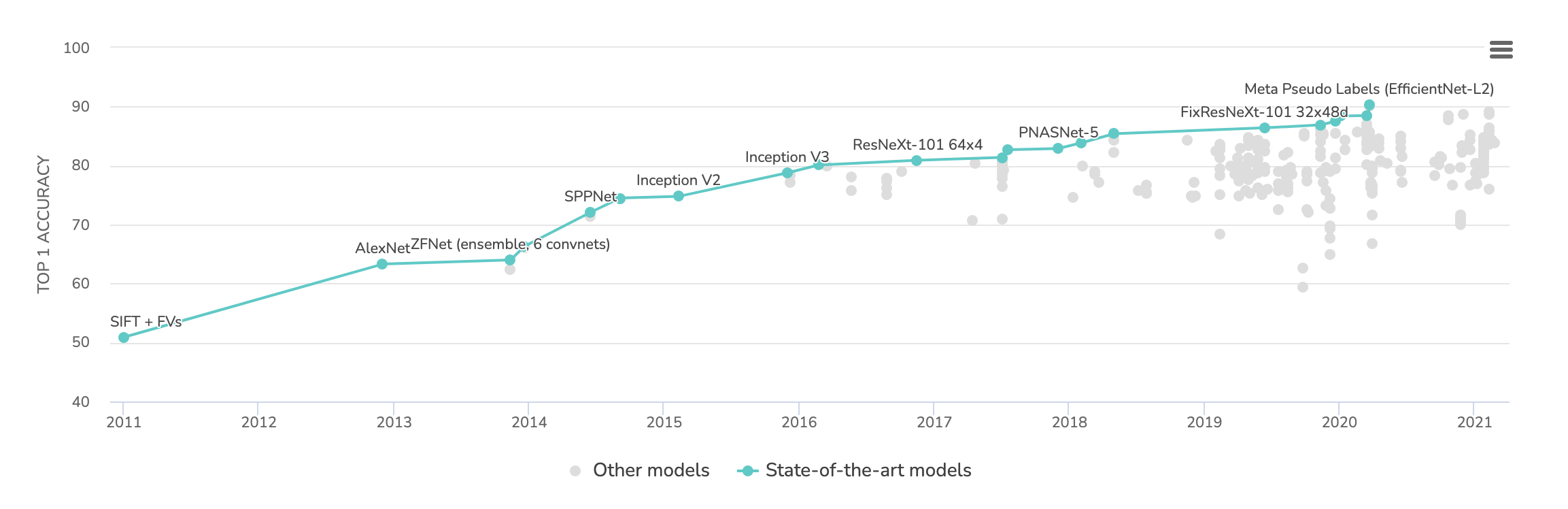}
    \caption{The image classification results on ImageNet from  2011 to 2021. As we can see, the recent published results are reaching into the ceiling. The data is collected from  \url{https://paperswithcode.com}. } 
    \label{fig:imagenet}
\end{figure}

In the age of deep learning, a hot direction is to obtain SOTA results.  Following~\cite{DBLP:journals/cacm/SchwartzDSE20}, we call such research trend as \textcolor{red}{\textbf{Red AI}}. Recently,  researchers have noticed that it was harder to gain an advantage over SOTA results.  For traditional AI fields, like CV and NLP, the improvements achieved by new AI models/algorithms are diminishing. Many popular research benchmarks are reaching their performance ceiling. Figure~\ref{fig:ende} and~\ref{fig:imagenet} list several examples showing how the returns of deep learning are diminishing over time.

The trend of red AI requires massive computations to achieve better results.  For example, as reported in \cite{DBLP:journals/cacm/SchwartzDSE20}, the amount of computations used to train deep learning models has increased 300,000x in 6 years.  These computations not only cause expensive financial costs but also contribute to an excessive carbon footprint. The former harms AI inclusiveness and the latter harms our environment. We classify the computation source required by deep learning into the following three categories: model size, parameter tuning, and training data. 

1) With access to large-scale data resources, increasing model size is the simplest way to improve results. For example, on WMT English-German translation, the performance can be increased from 27.3 to 28.4 while the size of the machine translation models is increased from 60M (Transformer-base) to 180M (Transformer-large). To achieve better results, more and more AI participants would like to increase model size as much as possible, especially for rich organizations. For example, researchers from OpenAI first pre-trained a large-scale text generation model, called GPT-3 with 175B parameters.  It shows that a super-large model can generate human-like texts. However, according to~\cite{DBLP:conf/acl/StrubellGM19}, training GPT-3 can emit almost 500M carbons, almost emissions of five cars in their lifetime. These studies are key milestones in a long run. However, we believe that larger models are not always better if we consider ``invisible'' computation cost. We are still concerned about ``the crazy love'' to super-large models no matter whether the increased computations bring significant benefits. In addition, bigger models largely increase the burden of inference serving. Amazon estimates that 90\% of production ML infrastructure costs are for
inference, not training~\citep{DBLP:journals/corr/abs-1901-10008}. 

2) Model experiments are also an overlooked computation consumer. To verify the effectiveness of a new model/algorithm, AI participants usually conduct massive experiments, including model/algorithm implementation, baseline re-implementation, and hyper-parameter tuning.  First, baseline re-implementation is an redundant computation source. For example, the original Transformer paper has 18K citations. Assume each citation represents a single implementation.  Each re-implementation takes 100 hours on a single GPU (following the cost of running a Transformer-base model on English-German translation). It means that only baseline re-implementation on a single dataset can take 1.8M GPU hours.  In addition,  hyper-parameter tuning is an overlooked computation source.  We design a simple questionnaire to ask the ratio of experiments experiments for hyper-parameter tuning while developing a new model/algorithm, answered by 64 AI specialists, including researchers and engineers. All of the surveyed choose tuning hyper-parameters. To be specific, 10.7\%, 32.1\%, 35.7\%, 21.4\% individuals take 80\%-100\%, 50\%-80\%, 30\%-50\%, 0\%-30\% experiments to tune hyper-parameters. In sum, 42.8\% individuals spend over 50\% experiments for hyper-parameter tuning.

3) Starting from shallow models, it is popular to increase the amount of training data to achieve better generalization ability, especially in semi-supervised settings. One recent hot topic is pre-training a super-large model on billions of raw data.  In the NLP field, ELMo~\citep{DBLP:conf/naacl/PetersNIGCLZ18} is the first well-known work to explore large-scale pre-training.  Following ELMo, BERT pre-trains a Transformer encoder on 3 billion word pieces.  Researchers from OpenAI recently proposed GPT-3, a generative model pre-trained on 45TB data. These massive training examples largely increase the training costs compared to previous shallow models.

In all, the trend of Red AI brings heavy computation costs. These computations exacerbate the research inequality, making it difficult to involve all researchers in Red AI. Furthermore, massive computation requirements bring huge carbon emissions.  To address these problems, \textcolor{green}{Green} deep learning, or Green AI, was first proposed by~\cite{DBLP:journals/cacm/SchwartzDSE20} to encourage the AI community to focus more on energy costs.

 \section{Green Deep Learning}
 \label{sec:greenlearning}

In this section, we mainly describe what is Green deep learning, how to evaluate ``greenness'' in deep learning, and  why Green deep learning matters.

\subsection{Definition}
Green learning, a term first proposed by~\citet{DBLP:journals/cacm/SchwartzDSE20}, is gaining mounting attention. Formally, \textcolor{green}{Green} deep learning, or Green AI, appeals to researchers to obtain novel results without increasing computational cost rather, ideally reducing it. Unlike Red AI pushing state-of-the-art results at any cost, Green deep learning encourages AI participants to achieve comparable or better results using as few computations as possible.

\subsection{Measure}

In Green deep learning, computations are important evaluation metrics. Currently, the whole community lacks a comprehensive and widely-accepted measure to evaluate computations because multiple aspects can attribute to computations, including model size, training examples, and so on. A comprehensive measure is expected for a fair comparison. Here we list several computation measures and discuss their merits and demerits. 

\paragraph{Running time} Some studies adopt the total training time as a kind of computations measure. If all models/algorithms adopt the same hardware and software settings, it is the most natural measure to evaluate training/inference computations. However, since running time heavily relies on infrastructure settings, it is not suitable for comparing models running on different infrastructures. Even so, we still encourage AI participants to report the running time for an intuitive understanding. 

\paragraph{Carbon emission} Carbon emission is the most direct approach to evaluate environmental effects. In order to quantify carbon emissions,~\citet{DBLP:journals/corr/abs-1910-09700} used $CO_2$-\textit{equivalents} ($CO_2eq$) as the amount of $CO_2$ which would have the equivalent global warming impact. However, the main challenge of this measure lies in accurate estimation. First, computations via electricity consumption are easily influenced by local infrastructures. Furthermore, it is hard for AI participants to estimate the amount of $CO_2$ if they do not run experiments on well-known cloud platforms.  Therefore, it is also not suitable as a standard metric to compare different models running on different regions and different computing infrastructures.

\paragraph{Model size}  The model size is also an important factor in deciding training and inference costs. We encourage researchers to report model size to corporate with other measures in practice.

\paragraph{FLOPs} Floating-Point Operations (FLOPs) count the number of works required for running a model when executing a specific instance. Previous studies usually adopt this metric to evaluate efficiency.  FLOPs are almost independent of hardware and software platforms, being the simplest measure to conduct a fair comparison between different models.  However, FLOPs are theoretical values, and there is a gap between FLOPs and running time. In addition to the total amount of works (FLOPs), the degree of parallelism also affects the running time. 

According to these measures, we summarize evaluation strategies towards fair comparison and intuitive understanding. 
 
\paragraph{Fair measure} Generally speaking, AI participants prefer well-performing models/algorithms with fewer computations. Therefore, it is an important question to fairly compare computations required for training and inference. We strongly suggest reporting FLOPs during model training and inference.  Last but not least, to evaluate the wasted and redundant computations required for developing a new model/algorithm, we also encourage researchers to report the total FLOPs during all experiments, including but not limited to parameter tuning and baseline implementation. 

\paragraph{Intuitive understanding} To increase the intuitively understanding about computations, we encourage researchers to report running time, model size, and carbon emission as optional results. 

\subsection{Broader Impact}
First, Green deep learning can help deep learning empower real society applications better. In the past decade, deep learning continuously pushed state-of-the-art results on research benchmarks, like machine translation, image classification, and so on. In the research community, the cost of deep learning seems to be nothing compared to energy consumption of all human activities. Nowadays, deep learning is widely applied to real society tasks, such as auto-driving, face recognition, drug discovery, and so on. Once deep learning is involved in large-scale applications, the cost of deep learning will be multiplied hundreds of millions of times.  Furthermore, some edge applications, like mobiles with extremely few computation resources, also require Green deep learning.  Therefore, Green deep learning is a necessary research direction in the future.

Second, Green deep learning can largely improve AI inclusiveness. We note the contributions of rich organizations for pushing higher results on many downstream tasks. Meanwhile, we also notice the dilemma of researchers from academics and developing countries on engaging Red AI research. Most researchers only have limited computations, which could not support them to develop super-large models with state-of-the-art results. Unfortunately, compared to ideas with state-of-the-art results, novel and innovative ideas without state-of-the-art results are losing their sounds.  For example, news media would like to report studies with state-of-the-art results. The attractive propaganda of No.1 also pushes the rich organizations to pour more money on super-big models. These cases may confuse researchers on how they engage in deep learning research without strong financial support. We argue that state-of-the-art results are good, but not the only criteria to evaluate the quality of new models/algorithms.  We encourage rich organizations to continuously explore data and model boundaries, also encourage the AI community to pay attention to innovative ideas.  In fact, deep learning struggles many years until it outperforms shallow models with feature engineers. We believe that the development of AI should be diverse. Green deep learning can improve AI inclusion and motivate more AI participants to explore deep learning possibilities.




\section{Outline of the Survey}

It is a long-term goal to develop tiny yet strong networks for all AI researchers and engineers. Driven by this target, several popular tiny networks have been proposed~\citep{DBLP:journals/corr/HowardZCKWWAA17,DBLP:conf/cvpr/Chollet17, DBLP:conf/icml/TanL19}. For example, MobileNet proposed by~\citet{DBLP:journals/corr/HowardZCKWWAA17} is an efficient architecture based on depthwise separable convolution. Similar idea has been adopted at Xception~\citep{DBLP:conf/cvpr/Chollet17}. Recently, to explore extremely tiny networks, advanced training/inference/network surgery methods have been proposed. For example,   EdgeBERT~\citep{DBLP:conf/micro/TambeHPJYDSWR0W21} is proposed to build an extremely tiny network that can run on IoT devices. It adopts advanced methods like quantization, pruning, early exit to further reduce model 
parameters and running computations.

In this survey, we give a systematic review of Green deep learning technologies. We first build a green technology taxonomy and then classify the related technologies into four categories, including compact networks, energy-saving training strategies, energy-saving  inference, and efficient data usage. In each category, we review the current progress on Green technologies and explore potential issues.

It is important to note that building a Green technology taxonomy is challenging since there lacks a unified standard measurement. For example, BERT requires massive computations during training. If we only consider training costs, BERT can not be treated as a Green technology. However, BERT can improve downstream performance with fewer training examples. If we consider its transfer ability, BERT  is absolutely a Green technology. Therefore, whether a technology is defined as Green or not is open to doubt. We will try our best to avoid giving a biased definition. If a technology has the potential to reduce the costs of deep learning, we will include it in the green technology taxonomy. We review Green deep learning technologies in the following categories:
\begin{itemize}

 \item \textbf{Compact Architecture Design.} This part focuses on small networks. We split this chapter into two sub-chapters, i.e., component design, and component assembling. The component design focuses on subtle components with competitive results but much fewer computations. Component assembling describes how to build a network efficiently. 

\item \textbf{Energy-efficient Training Strategies.}  Previous studies have proposed several efficient training approaches. In this survey, we classify these studies into four categories, including initialization, normalization, progressive training, and efficient AutoML. 

\item \textbf{Energy-efficient Inference. } In this chapter, we describe approaches that aim to get a smaller yet comparable network from a larger network for efficient inference, including model pruning, low-rank factorization, quantization, distillation.

\item \textbf{Efficient Data Usage.} This chapter lists algorithms that leverage training data efficiently. We focus on two popular directions: active learning and pre-trained models as few-shot learners. 

\end{itemize}

\clearpage

\chapter{Compact Architecture}
\label{cha:network}

\begin{figure*}[thp]
  \centering
\begin{forest}
  forked edges,
  for tree={
    grow=east,
    reversed=true, 
    anchor=base west,
    parent anchor=east,
    child anchor=west,
    base=left,
    font=\footnotesize,
    rectangle,
    draw=hiddendraw,
    rounded corners, 
    align=left,
    minimum width=2.5em,
    minimum height=1.2em,
    s sep=6pt,
    inner xsep=3pt,
    inner ysep=1pt,
  },
  where level=1{font=\scriptsize}{},
  where level=2{font=\scriptsize}{},
  where level=3{font=\scriptsize}{},
  where level=4{font=\scriptsize}{},
  where level=5{font=\scriptsize}{},
  [Compact Architecture Design
    [Component \\ 
    Design
      [Compact Convolution
        [Depth-wise Separable Convolution]
        [Fire Convolution]
        [Flattened Convolution]
        [Shrinked Convolution]
      ]
      [Efficient Attention
        [Sparse Attention]
        [Attention Approximation]
      ]
      [Lightweight Softmax]
      [Compact Embedding]
    ]
    [Component \\ 
    Assembling
      [Memory Sharing]
      [Static Weight Sharing
        [Cross-layer Parameter Sharing]
        [Cross-data Parameter Sharing]
      ]
      [Dynamic Weight Sharing
        [Cascading]
        [Early Exit]
        [Skipping]
        [Mixture of Experts (MoE)]
      ]
      [Deployment Weight Sharing]
    ]
    [Compact-architecture \\ Search]
  ]
\end{forest}
\caption{Taxonomy of compact architecture design with representative examples.}
\label{fig:taxonomy_of_compact_architecture_design}
\end{figure*}
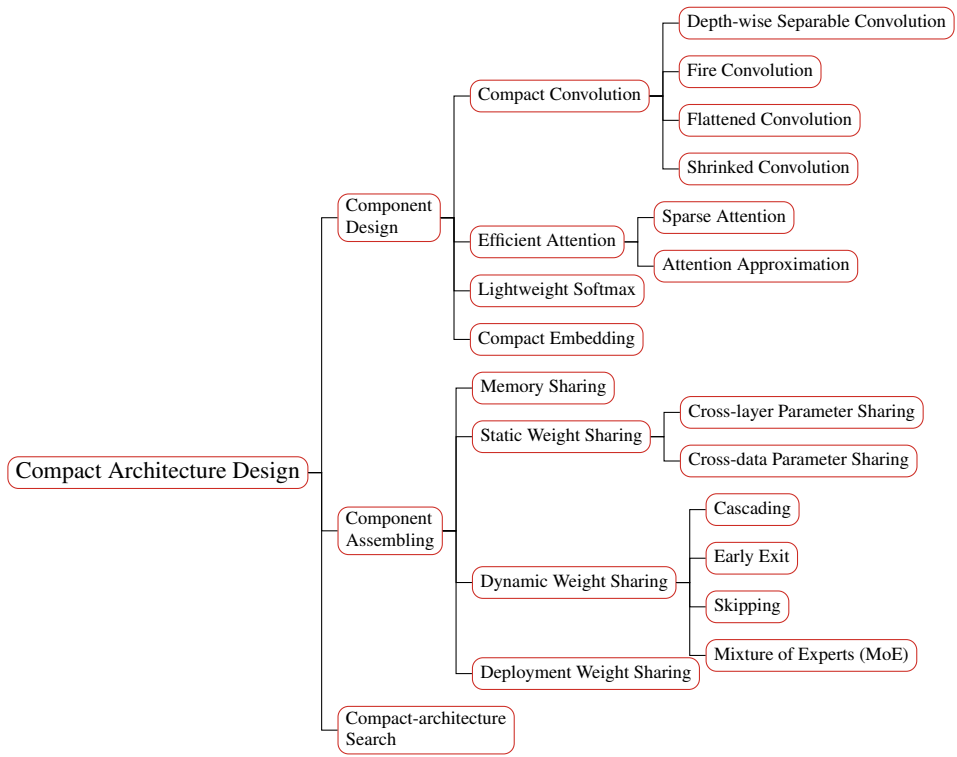

 Developing efficient neural networks has been a long-standing goal towards Green AI.  In this survey, we define Green networks as neural networks that are efficient in terms of computational costs. We can generate compact networks via subtle design, model surgery, and network search. Subtle design means that we can manually define efficient architectures requiring fewer computations. Model surgery means that we can generate compact architectures from a larger model via parameter reduction. In this chapter, we focus on architectures with subtle design and leave the details of network surgery to Chapter 4. An overview of this section is shown in Figure~\ref{fig:taxonomy_of_compact_architecture_design}.



\section{Component Design}
In this section, we describe efficient variants of popular components, including convolution, attention, softmax, and embedding.

\subsection{Compact Convolution}
Starting from AlexNet~\citep{DBLP:conf/nips/KrizhevskySH12}, it has been a hot direction to build deeper and larger CNNs to achieve better performance~\citep{DBLP:journals/corr/IandolaMKGDK14,DBLP:journals/corr/SimonyanZ14a, DBLP:conf/cvpr/SzegedyLJSRAEVR15,DBLP:conf/cvpr/HeZRS16,DBLP:conf/aaai/SzegedyIVA17}. Currently, even a simple CNN baseline contains hundreds of layers and thousands of channels. To reduce deployment costs, previous studies proposed efficient variants. Here we list several widely-used variants.





\paragraph{Depthwise Separable Convolution}
This architecture has been adopted in Xception~\citep{DBLP:conf/cvpr/Chollet17} and MobileNet~\citep{DBLP:journals/corr/HowardZCKWWAA17}.   Depthwise separable convolution contains two components: depthwise convolution and pointwise convolution. The depthwise convolution applies a single filter for each input channel. The pointwise convolution is a kind of $1 \times 1$ convolution. Following this research line, many advanced variants have been proposed~\citep{DBLP:journals/corr/abs-1811-07083,DBLP:conf/cvpr/SandlerHZZC18}. For example, ~\citet{DBLP:conf/iccvw/WangLF17} also proposed a factorized convolution by unravelling the standard convolution and arranging the spatial convolution sequentially. 

\paragraph{Fire Convolution}
~\citet{DBLP:journals/corr/IandolaMAHDK16} proposed a vision model SqueezeNet. The fire module is the key building block. It is similar with depthwise separable convolutions. A fire module contains two components: a squeeze convolution layer with $1 \times 1$ filters and an expand layer with a mixture of $1 \times 1$ and $3 \times 3$ convolution filters.

\paragraph{Flattened Convolution}
It is proposed by~\citet{DBLP:journals/corr/JinDC14} to decrease the redundancy of the filters. It separates the 3D convolution filters into three consecutive 1D filters: convolution across channels (lateral), vertical, and horizontal direction. 


\paragraph{Shrinked Convolution}
Traditional convolutions usually have fixed hyper-parameter settings, like the number of filters.  Different from these models,  MobileNet~\citep{DBLP:journals/corr/HowardZCKWWAA17} adopts a dynamic setting, also called shrinked convolution. It introduces a width multiplier to thin a network uniformly at each layer. The standard convolutions with $M$ input channels and $N$ output channels become a shrinked convolution with $\alpha M$ input channels and $\alpha N$ output channels where $\alpha \leq 1$.






\subsection{Efficient Attention}

Attention~\citep{DBLP:journals/corr/BahdanauCB14} is first proposed for handling long-distance dependency in machine translation. Currently, it has been widely used in tasks like summarization, natural language understanding, and so on. The key idea is to dynamically attend all tokens at each step. All tokens can be directly aligned together, which can address long-distance dependencies to some extent. Since any two tokens have an attention score, the required computations grow quadratically with the input length. To address this problem, previous studies proposed several efficient attention variants.  Currently, dop-product-based attention~\citep{DBLP:conf/nips/VaswaniSPUJGKP17} becomes the dominant choice for NLP and CV applications. For simplification, we refer to attention as dot-product attention. 

We roughly classify these variants into two categories: sparse attention that 
reduces the span of attention, and attention approximation with different attention estimation formats. Let us review the original self-attention definition. Formally, given a sequence of hidden vectors $x$, we can map it into different representation space $Q$, $K$, and $V$. Then, attention takes $Q$, $K$, and $V$ as inputs and is responsible for generating vector via the following equation:
\begin{equation}
    \text{Attention}(Q,K,V)=\text{softmax} (\frac{QK^T}{\sqrt{d_k}}) V
    \label{eq:attention}
\end{equation}
where $Q$, $K$, $V$ are 3-dimension tensors, with dimensions of sequence length, head number, hidden dimension. The computations mainly come from $(\frac{QK^T}{\sqrt{d_k}})$ and softmax operations.  This section describes several approaches to reduce dot-product computations. We leave the details of efficient softmax variants in Section 2.1.3.

\paragraph{Sparse Attention}
As we can see from Eq.~\ref{eq:attention},  all tokens are required to be attended at each step. Several approaches are proposed to reduce attention length by only attending local tokens at each step.  A natural solution is to reduce the number of attended tokens by assigning some tokens with zero weights. It is the idea of sparse attention~\citep{DBLP:conf/icml/MartinsA16,DBLP:journals/corr/abs-1904-10509,DBLP:conf/emnlp/CorreiaNM19,DBLP:conf/acl/DaiYYCLS19, DBLP:conf/nips/ZaheerGDAAOPRWY20}.  ~\citet{DBLP:conf/icml/MartinsA16} proposed Sparsemax, which added a $L_2$ regularization to encourage the attention matrix to be sparse. Sparsemax has been applied to various architectures~\citep{DBLP:conf/nips/NiculaeB17,DBLP:conf/naacl/MarufMH19,DBLP:conf/acl/PetersNM19}. ~\citet{DBLP:journals/corr/abs-1904-10509} introduced heuristic rules and defined two sparse attention variants. One attends previous $l$ tokens. The other splits a sequence into different spans where each span has $l$ tokens. Each head attends to every $l$ tokens where $l$ is much smaller than the length of inputs.  
~\citet{DBLP:conf/acl/SukhbaatarGBJ19} believed that the naive sparse attention was somehow arbitrary. They found that some attention heads focused on the recent tokens, while other heads took information from the whole context. Motivated by this, they proposed adaptive attention by learning the attention span of each head. An similar idea is proposed by~\citet{DBLP:conf/emnlp/CorreiaNM19}. They introduced an adaptive sparse attention approach. They replaced full softmax operations with $\alpha$ softmax that allowed low-scoring words to receive precisely zero weight. In addition to attention length, the attention head is also an important sparse factor. ~\citet{DBLP:conf/acl/VoitaTMST19} found that only a subset of heads matter and the rest can be pruned. 

\paragraph{Attention Approximation} ~\citet{DBLP:conf/iclr/KitaevKL20} proposed an efficient attention model Reformer. It approximates the dot-product attention computation by one that uses locality-sensitive hashing, reducing the complexity from $O(L^2)$ to $O(L \log L)$, where $L$ is the length of the sequence. ~\citet{DBLP:journals/corr/abs-2009-14794} further proposed a more efficient model Performer with an unbiased positive random feature map estimator. Compared to the original attention, Performer is a linear architecture with compatible performance.

\subsection{Lightweight Softmax} 
Softmax layer is a necessary component for deep learning. The key idea is to normalize a vector to a probability distribution of possible labels. Traditional softmax is computed as:
\begin{equation}
    p(y_i| x) = \frac{\exp(h_i\cdot w)}{\sum_{j=1}\exp(h_j \cdot w)}
\end{equation}
where $y_i$ is the $i$-th label and $w$ are learnable parameters. $h_i$ is the $i$-th dimension of hidden vector. $x$ is the input sequence. The denominator requires the dot-product over label candidates. If the task has a large label set, the denominator will require large computations. Since the complexity of softmax is proportional to the number of labels and sequence generation~\citep{DBLP:journals/corr/abs-1301-3781,DBLP:conf/nips/MikolovSCCD13} tasks usually have a large vocabulary size, we take sequence generation as an example to show several lightweight variants with fewer computations. In sequence generation tasks, token vocabulary is equal to the label set.  Formally, given a hidden vector $h$ and all token embeddings, the softmax for sequence generation is computed as:
\begin{equation}
    p(y_i|x) = \frac{\exp( h^T  v_i)}{\sum_{j=1}\exp(h^T v_j)}
    \label{eq:softmax}
\end{equation}
where $x$ is the input and $v_i$ is the embedding of the $i$-th token.   Eq.~\ref{eq:softmax} shows that the softmax layer introduces embeddings for all tokens and requires the inner-product between hidden vector and all embeddings. A large vocabulary will require many computations.  Therefore, several efficient softmax variants have been proposed.

\paragraph{Fewer Parameters} To reduce memory usage, ~\citet{DBLP:conf/eacl/PressW17} proposed to tie input embeddings and embeddings in Eq.~\ref{eq:softmax}. They conducted experiments on machine translation. Results show that weight sharing can reduce the size of neural translation models without harming translation results. In addition, reducing the number of labels is another important research direction. Recently, several studies have been ~\citep{DBLP:conf/aaai/KimJSR16,DBLP:conf/acl/Costa-JussaF16} proposed to generate a sequence in character level, rather than in word level. The number of characters is largely less than that of words and the computations for softmax can be largely reduced. Similarly, ~\citet{DBLP:journals/corr/JozefowiczVSSW16} implemented character-based softmax on language modeling, which achieved promising results. It is important to note that these character-based methods also bring longer sequences. Current sequence generation models usually adopt auto-regressive generation frameworks. The longer sequence brings higher decoding costs. In all, it should be considered case-by-case whether character-based methods reduce the whole decoding cost. Recently, a trade-off is achieved by sub-word level vocabularies~\citep{DBLP:conf/acl/SennrichHB16a}. Sub-word level vocabularies have a tradeoff granularity between character vocabularies and word vocabularies. Sub-word level vocabularies have more tokens than character-level vocabulary but also have much shorter segmented sequences. Therefore, sub-word level vocabularies become the popular choice for almost all sequence generation tasks. 

\paragraph{Fewer Computations} We classify softmax variants with fewer computations into five categories: hierarchical softmax, softmax with dynamic embeddings, sampling-based softmax, hashing-based softmax, and normalization-based softmax. Hierarchical softmax  (H-Softmax)~\citep{DBLP:conf/aistats/MorinB05,DBLP:conf/nips/MnihH08} is a kind of softmax variant. To be specific, it formulates a label set as a tree and all labels in the set is the leaf node.  The complexity can be dropped from $O(N)$ to $O(log(N))$ where $N$ is the size of the label set. In this way, the traditional one single probability over labels is decomposed into a product of a sequence of probability over each tree layer. The regular softmax can be regarded as a tree of depth $1$, with all labels as leaf nodes.   The second research direction focuses on dynamic label embeddings~\citep{DBLP:conf/acl/ChenGA16}. The intuition is that not all labels require the same parameter size. It assigns variable parameter sizes for different labels. In particular, the approach assigns more parameters to frequent labels. The embedding size affects the computation costs. Therefore, this kind of method can reduce the computations required by softmax operations. In addition, sampling-based softmax aims to estimate the full softmax computations with sampled label candidates. The key idea is to sample several label embeddings to estimate all embeddings~\citep{DBLP:conf/aistats/BengioS03,DBLP:journals/tnn/BengioS08,DBLP:conf/acl/JeanCMB15}. Hoever, it only reduces training costs while the full softmax is still be computed to obtain a variance-free result during inference. Hashing-based softmax is another kind of estimation variant.  ~\citet{DBLP:journals/corr/VijayanarasimhanSMY14} proposed a fast locality-sensitive hashing technique to approximate the actual dot-product.  Normalization-based softmax~\citep{DBLP:conf/acl/DevlinZHLSM14,DBLP:conf/naacl/AndreasK15} aims to avoid explicit denominator. The target is to output a vector as close as the probability distribution with the sum being 1.

\subsection{Compact Embeddings} 

Building token embeddings is the first step for NLP tasks. The parameters of embeddings are decided by vocabulary size and embedding length. How to reduce embedding parameters is an important and interesting topic. Learning compact token vectors is related to learning compressed neural networks. There have been several techniques for learning compact neural networks, like pruning, knowledge distillation, low-rank approximation, and quantization. In this part, we only focus on related approaches for compact embeddings. We classify these approaches into four categories: reuse-based approaches, knowledge-distillation-based approaches, low-rank-based approaches, fine-grained vocabularies.

Reuse-based approaches focus on compositional embeddings~\citep{DBLP:conf/acl/FaruquiTYDS15, DBLP:conf/acl/ChenMXLJ16,DBLP:conf/iclr/ShuN18, DBLP:conf/naacl/JoshiCLWZ19,DBLP:conf/kdd/ShiMNY20}. For example,  ~\citet{DBLP:conf/acl/FaruquiTYDS15} aimed to represent each token embedding as a sparse linear combination of basis vectors. The size of basis vectors is much less than token embeddings. Similar idea has been proposed by~\citet{DBLP:conf/acl/ChenMXLJ16}. They split the vocabulary into two parts. One part is a base set containing frequent tokens with fixed size (e.g., 8K), the other part is a set of rare tokens whose embeddings are encoded by the base set's embeddings. Following these studies, ~\citet{DBLP:conf/iclr/ShuN18} adopted the quantization approach to construct embeddings with few basis vectors.  Recently,  this idea has been adapted to other fields beyond NLP, like recommendation systems~\citep{DBLP:conf/kdd/ShiMNY20}.

In addition, traditional compression approaches have been applied to compress embeddings. ~\citet{DBLP:conf/cikm/MouJXL0J16} used knowledge distillation to transfer knowledge from a big token embedding layer into a smaller embedding layer. ~\citet{DBLP:conf/nips/ChenSLCH18} used vocabulary-partition (block) based low-rank matrix approximation to reduce parameter size. ~\citet{DBLP:journals/corr/abs-1803-05651}  used 1-2 bits per parameter, rather than traditional 32-bits, for token embedding.  Vocabulary size is also an important factor in deciding embedding size. Therefore, fine-grained vocabularies have been proposed to reduce the vocabulary length, like character-level vocabulary~\citep{DBLP:conf/aaai/KimJSR16}, subword-level vocabulary~\citep{DBLP:conf/acl/SennrichHB16a}, and byte-level vocabulary~\citep{DBLP:conf/aaai/WangCG20}. 

\section{Component Assembling}
This part presents several component assembling solutions for efficient architecture design.  Many widely-used architectures are efficient component assembling solutions.  CNNs and LSTMs are representative models. A single filter in CNNs can handle all input spans. LSTMs adopt the same parameters for all steps.  The key idea of efficient component assembling lies in \textit{sharing}. We classify these assembling solutions into four categories: memory sharing, static weight sharing, dynamic weight sharing, and deployment weight sharing.

\subsection{Memory Sharing}

Memory sharing is a common technique to store a large model on devices with limited memories.  A natural idea is to share the same storage among intermediate forward vectors~\citep{DBLP:journals/corr/PleissCHLMW17} or backward vectors~\citep{DBLP:journals/corr/ChenXZG16,DBLP:conf/nips/GruslysMDLG16}. There are also some reversible models~\citep{DBLP:conf/nips/GomezRUG17, DBLP:conf/nips/MacKayVBG18} where the activation of each layer can be reconstructed from the next layer to reduce memory requirements during the backward process. The models do not need to save intermediate activation vectors. 
Since several vectors share the same storage space, recomputation is necessary in some cases. To achieve a trade-off between efficient memory usage and fewer computations, some studies~\citep{DBLP:conf/ppopp/WangYZWLSXK18} proposed to combine memory sharing with liveness analysis~\citep{DBLP:conf/ics/WangWXXY16}. During graph computation, GPUs adopt liveness analysis to create tensors and free tensors. For large intermediate tensors, frequent allocation/deallocation operations are time-consuming. Therefore, the runtime can be reduced by directly reusing memory segments from a huge pre-allocated memory pool. In addition to memory optimization on a single node, ~\citet{DBLP:conf/sc/RajbhandariRRH20} further explored memory sharing on distributed settings across multiple computation nodes.



\subsection{Static Weight Sharing}
Unlike memory sharing, static weight sharing aims at exploring how to reuse weights for a neural network. The difference between weights and intermediate vectors is that weights are fixed during inference and shared by all examples. To save memory, many models choose to reuse parameters across different layers or different tasks.

\subsubsection{Cross-layer Parameter Sharing}

Cross-layer parameter sharing is a common technique for parameter efficiency. The idea of sharing parameters across layers has been well explored~\citep{DBLP:conf/iclr/DehghaniGVUK19,DBLP:conf/nips/BaiKK19,DBLP:conf/iclr/LanCGGSS20}. \citet{DBLP:conf/iclr/SavareseM19} proposed a parameter sharing scheme that defined a global bank of templates. The parameters of each layer of a CNN come from the linear combination of these templates.  ~\citet{DBLP:conf/iclr/DehghaniGVUK19} proposed a model, called Universal Transformer, where all layers shared the same parameters. Following these study, ~\citet{DBLP:conf/iclr/LanCGGSS20} applied cross-layer sharing mechanism on pre-train/fine-tune settings and ~\citet{DBLP:journals/corr/abs-2104-06022} proposed diverse sharing strategies. Recently, ~\citet{plummer2020neural} adopted the idea of network architecture search to automatically learn how to share parameters between all layers in a network. 

\subsubsection{Cross-task Parameter Sharing} 

Cross-task parameter sharing is also a popular solution to handle multi-task, multi-domain, or multi-lingual problems~\citep{DBLP:journals/corr/RamsundarKRWKP15,DBLP:conf/acl/DuongCBC15,DBLP:conf/acl/SogaardG16,DBLP:conf/emnlp/HashimotoXTS17,DBLP:conf/iclr/YangSC17,DBLP:journals/jmlr/RaffelSRLNMZLL20}. The key idea of cross-task is to enable all tasks (or languages/domains) to share parameters. Multi-task learning~\citep{ruder2019neural} has two popular implementations, including hard and soft parameter sharing. Compared to soft parameter sharing where different tasks do not have shared networks, hard parameter sharing uses fewer parameters. Therefore, we only focus on hard parameter sharing in this work.

To be specific, cross-task sharing is initially implemented by sharing the hidden layers between all tasks, while keeping several task-specific output layers~\citep{DBLP:conf/iclr/YangSC17,DBLP:conf/icml/HoulsbyGJMLGAG19,DBLP:journals/jmlr/RaffelSRLNMZLL20}. For the CV field, multi-task solutions often share CNN layers. For the NLP filed, in addition to naive sharing, researchers also focus on finding better parameter reusing solutions for different tasks. For example, ~\citet{DBLP:conf/acl/SogaardG16} found that low-level tasks, e.g., part-of-speech tagging, should share parameters at lower layers. Motivated by these findings, ~\citet{DBLP:conf/emnlp/HashimotoXTS17} proposed a parameter-sharing network across multiple NLP tasks. Currently, the trend of developing large-scale models encourages researchers to directly use one single model to support multiple tasks. T5~\citep{DBLP:journals/jmlr/RaffelSRLNMZLL20} is one  representative model.

Multilingual is also a special cross-task variant. At the early stage of multilingual models, researchers usually choose to share a part of parameters across different languages~\citep{DBLP:conf/naacl/FiratCB16,DBLP:conf/acl/UpadhyayFDR16,DBLP:conf/coling/BlackwoodBW18}. Recently, multilingual approaches usually treated all languages equally and mixed them together to train a single model~\citep{DBLP:journals/corr/HaNW16,DBLP:journals/csl/FiratCSYB17,DBLP:journals/tacl/JohnsonSLKWCTVW17,DBLP:journals/corr/abs-2010-11125}. More recently, adapter-based solutions have been widely used for modeling task-specific features beyond shared parameters~\citep{DBLP:conf/icml/HoulsbyGJMLGAG19,DBLP:conf/emnlp/BapnaF19,DBLP:conf/emnlp/PfeifferRPKVRCG20}.

\subsection{Dynamic Weight Sharing}

Static parameter sharing usually relies on pre-specified networks. Researchers define heuristic rules based on shared features to decide which layers/components should be shared by different inputs/tasks. Although this solution is natural, hard sharing usually fails in handling tasks that are not closely related. Dynamic solutions are proposed to decide which layers/components should be shared among different input samples. Specifically, dynamic networks are neural networks with dynamic computational graphs where the computational topology or parameters are decided on the fly. Therefore, this kind of network can reduce computation costs and improve the adaptiveness of networks. In this survey, we describe the overview of general dynamic architectures. If you are interested in other dynamic features, you can find surveys focusing on dynamic networks~\citep{DBLP:journals/corr/abs-2102-04906}. 
Networks with dynamic architecture can be classified into the following classes:
\begin{itemize}
    \item \textit{Cascading-style Networks}. Multiple basic networks are cascaded in a directed acyclic graph (DAG) in a from-small-to-big manner, where the model first executes smaller networks, then larger networks. If a smaller network can handle the input sample, the model will stop the execution process and does not run execute models. 
    
    \item \textit{Early-exit-style Networks}. A single network contains multiple internal classifiers, allowing ``easy'' samples to exit at shallow layers. The difference with cascading-style networks lies in early-exiting networks feed the output of previous layer to the next layer while  cascading-style networks cut off this information flow and every network only takes raw samples as inputs. 
    
    
    \item  \textit{Skipping-style Networks}. It accelerates inference by either skipping certain layers, or skipping unimportant input spans in the whole input sequence.
    
    \item \textit{Mixture-of-experts-style Networks}. Multiple experts are provided as candidates in the same block. Only a small part of experts are used in each block for inference.
    
\end{itemize}

\textbf{Cascading-style dynamic networks}  have a historical development~\citep{viola2001rapid,lienhart2002extended,viola2004robust}. The authors cascade architectures are originally proposed for unbalanced binary classification tasks. They cascaded multiple basic models and fed the input to the next model only if the current model was not confident of its prediction. For example, \citet{DBLP:conf/codes/ParkKKKKYY15} cascaded two VGG networks in a small-first manner to obtain a better trade-off between classification accuracy and energy consumption. The smaller model can handle most samples, which largely reduce inference costs.  \citet{DBLP:conf/icml/BolukbasiWDS17} cascaded AlexNet, GoogLeNet, and ResNet together. \citet{DBLP:conf/uai/WangLCTYG18} introduced a cost-aware objective for jointly training criterion functions among basic models. More recently, \citet{DBLP:journals/corr/abs-2012-14682} proposed a dynamic framework for accelerating the inference of pre-trained language models, CascadeBERT, which dynamically selected proper-sized and complete models in a cascading manner. 

\begin{table*}
\renewcommand\arraystretch{1.5}
\caption{An overview of widely-used confidence criteria deciding whether the forward process should be terminated in cascading-style and early-exiting-style networks. In the Formulation column, $\mathbbm{1}(\cdot)\in\{0,1\}$ indicates  action \{``continue'', ``terminate''\}. $\alpha$ is the threshold. $\lambda$ and $\tau$ are hyper-parameters. $\text{MLP}(\cdot)$ is a learnable module.
}
\label{tab:dynamic_criteria_sum}
\centering
\resizebox{\textwidth}{!}{
\begin{tabular}{|c|c|c|}
\hline \textbf{Criterion} & \textbf{Descriptions} & \textbf{Formulation} \\
\hline 

\multicolumn{3}{|c|}{\textit{Confidence-based Criterion}} \\
\hline
\makecell*[c]{Score margin \\ \citep{DBLP:conf/codes/ParkKKKKYY15}} & 
\makecell*[c]{The gap between the largest and the second  \\ largest values among the predicted probability distribution.} &
\makecell*[c]{$\mathbbm{1}(\bm{\hat y}^{1\rm{st}}-\bm{\hat y}^{2\rm{nd}}<\alpha)$} \\
\hline 
\makecell*[c]{Entropy \\ \citep{DBLP:conf/icpr/Teerapittayanon16} \\ \citep{DBLP:conf/acl/LiuZWZDJ20} \\ \citep{DBLP:conf/acl/LiSSYQH20} } & 
\makecell*[c]{The entropy or normalized entropy of \\ the predicted probability distribution.} & 
\makecell*[c]{$\mathbbm{1}(H(\bm{\hat y})>\alpha)$} \\
\hline 
\makecell*[c]{Max probability \\ \citep{DBLP:conf/icml/KayaHD19} \\ \citep{DBLP:conf/nips/WangLHSYH20}} & 
\makecell*[c]{The maximum predicted probability.} & 
\makecell*[c]{$\mathbbm{1}(\max(\bm{\hat y})<\alpha)$} \\
\hline 

\multicolumn{3}{|c|}{\textit{Counting-based Criterion}} \\
\hline 
\makecell*[c]{Patience \\ \citep{DBLP:conf/nips/ZhouXGM0W20}} & 
\makecell*[c]{The number of identical predictions.} & 
\makecell*[c]{
$cnt_i^{cls}=\begin{cases} 
cnt_{i-1}+1 & \arg\max(\bm{\hat y}_i)=\arg\max(\bm{\hat y}_{i-1}) \\ 
0 & \arg\max(\bm{\hat y}_i) \ne \arg\max(\bm{\hat y}_{i-1}) \lor i=0
\end{cases}$ \\
$cnt_i^{reg}=\begin{cases} 
cnt_{i-1}+1 & |\bm{\hat{y}}_i-\bm{\hat{y}}_{i-1}|<\tau \\ 
0 & |\bm{\hat{y}}_i-\bm{\hat{y}}_{i-1}|\ge \tau \lor i=0 
\end{cases}$ \\
$\mathbbm{1}(cnt_i<\alpha)$} \\
\hline

\makecell*[c]{Voting \\ \citep{DBLP:journals/corr/abs-2105-13792}} & 
\makecell*[c]{The number of most predictions.} & 
\makecell*[c]{
$V_i=\max_c\{\sum_{l=1}^i\mathbbm{1}(\arg\max(\bm{\hat y}_i)=y_c)\}/i^\lambda$ \\
$\mathbbm{1}(V_i)<\alpha$} \\
\hline

\multicolumn{3}{|c|}{\textit{Learning-based Criterion}} \\
\hline 
\makecell*[c]{After-prediction \\ \citep{DBLP:conf/icml/BolukbasiWDS17} \\ \citep{DBLP:conf/uai/WangLCTYG18}  \\ \citep{DBLP:journals/corr/abs-2104-08803}} & 
\makecell*[c]{Take the predicted probability distribution  as input \\ and generate the label \\ deciding whether to execute the forward process. } &
\makecell*[c]{$\text{MLP}(\bm{\hat y})$} \\
\hline 

\makecell*[c]{Before-prediction \\ \citep{DBLP:conf/iclr/ElbayadGGA20} \\ \citep{DBLP:conf/eacl/XinTYL21}} &
\makecell*[c]{Take features as input and generate the label \\ deciding whether to execute the forward process.} &
\makecell*[c]{$\text{MLP}(\bm{h})$} \\

\hline
\end{tabular}}
\end{table*}


\textbf{Early-exiting-style dynamic networks} might be the most popular dynamic architecture nowadays~\citep{DBLP:conf/icpr/Teerapittayanon16,DBLP:conf/icml/BolukbasiWDS17,DBLP:journals/corr/abs-2103-01148,DBLP:conf/iclr/HuangCLWMW18,DBLP:conf/cvpr/YangHCSDH20,DBLP:journals/corr/abs-2105-15075}. With multiple internal classifiers on intermediate layers, a network is capable to give intermediate predictions and make decisions about whether to execute the forward process or not. If the answer is yes, current state would be fed to the next layer. Otherwise, the network outputs the intermediate prediction as the final prediction. Different from cascading architectures that cuts off the information flow between networks, early-exiting networks reuse the feature computed by previous layer. For example, BranchyNet~\citep{DBLP:conf/icpr/Teerapittayanon16} inserted several branch classifiers into a CNN to speedup inference. 
MSDNet~\citep{DBLP:conf/iclr/HuangCLWMW18} designed an exquisite two-dimensional multi-scale architecture to enable early exiting along two dimensions and RANet~\citep{DBLP:conf/cvpr/YangHCSDH20} further utilized the spatial redundancy for image classification.

In addition to the CV field, existing studies also apply early-existing-style dynamic networks to the NLP field~\citep{DBLP:conf/acl/XinTLYL20, DBLP:conf/acl/LiuZWZDJ20, DBLP:conf/acl/Zhu20}. 
The \textit{Two Stage} fine-tuning is the most representative approach to train early-existing-style dynamic networks in NLP
where the backbone is fine-tuned with the final classifier in the first stage, and the intermediate classifiers are fine-tuned in the second stage. In addition, joint training is also a trend to tune all parameters together including basic backbones and intermediate classifiers~\citep{DBLP:conf/acl/SchwartzSSDS20,DBLP:conf/naacl/LiaoZRSSH21,DBLP:journals/corr/abs-2101-09755}. In addition to training algorithms, recent researchers also focus on  criterion design. For example, ~\citet{DBLP:conf/nips/ZhouXGM0W20} and \cite{DBLP:journals/corr/abs-2105-13792} utilized counting-based criteria to support early exiting. 
Without relying on heuristic criteria, several approaches~\citep{DBLP:conf/eacl/XinTYL21,DBLP:journals/corr/abs-2104-08803} directly learned the criteria by introducing a small module to decide whether to execute the forward process. 
Due to the simplicity of single-step prediction, dynamic networks are widely applied to classification models. Recently, ~\citet{DBLP:conf/iclr/ElbayadGGA20} and ~\citet{DBLP:conf/acl/LiSSYQH20} extended multi-exit design to translation tasks and sequence labeling tasks.
Dynamic halting is a special case of early exiting, where the parameters across layers are shared and, therefore, the final classifier can also be shared. Specifically, these networks infer samples through a shared layer iteratively, rather than infer through multiple stacked individual layers. One representative network is proposed by \citet{DBLP:journals/corr/Graves16}. They proposed the adaptive computation time (ACT) mechanism for recurrent models to automatically decide how many times (iterations) each input symbol or token should be computed. 
Following this work, the ACT mechanism has been applied to various architectures, like ResNets and Transformers. For example, SACT~\citep{DBLP:conf/cvpr/FigurnovCZZHVS17} performed dynamic halting in two dimensions, including the coarse ACT among multiple layers within the same block and the fine-grained ACT on all spatial positions. Universal Transformer~\citep{DBLP:conf/iclr/DehghaniGVUK19} shared all layers within the encoder (or decoder) in Transformer. 

\paragraph{Discussion} 
In cascading-style or early-exiting style dynamic networks,  the key question is to figure out how confident the intermediate classifier is. 
Previous studies proposed various criteria for judging the reliability of an intermediate prediction. We categorize them into types 
as shown in Table~\ref{tab:dynamic_criteria_sum}. \textit{Score margin} is the gap between the largest and the second largest scores in the predicted probability distribution~\citep{DBLP:conf/codes/ParkKKKKYY15}. \textit{Entropy-based criterion} is based on the entropy of the predicted probability distribution~\citep{DBLP:conf/icpr/Teerapittayanon16,DBLP:conf/acl/LiSSYQH20, DBLP:conf/acl/LiuZWZDJ20}. The model executes the forward process only if the entropy is larger than the pre-defined threshold. \textit{Max-probability based criterion} is the gap between the max value of the predicted probability distribution and the pre-defined threshold~\citep{DBLP:conf/icml/KayaHD19,DBLP:conf/nips/WangLHSYH20}. \textit{Patience-based criterion} terminates the forward process only if the model generates continuously identical predictions~\citep{DBLP:conf/nips/ZhouXGM0W20}.
\textit{Voting-based criterion} is inspired by the ensemble technique, which terminates the forward process if the most of historic predictions reach an agreement~\citep{DBLP:journals/corr/abs-2105-13792}.  \textit{After-prediction based criterion} and \textit{before-prediction based criterion} introduce additional learning functions to learn whether to execute the  forward process. The only difference lies in that after-prediction based criterion uses the prediction distribution as inputs and before-prediction based criterion uses the naive hidden vector as inputs. 

\textbf{Skipping-style dynamic networks} skip some computations during forward process. These dynamic networks are capable to obtain higher efficiency. This dynamic solution has been widely-use in various models, like SkipNet~\citep{DBLP:conf/eccv/WangYDDG18}, ConvNet-AIG~\citep{DBLP:journals/ijcv/VeitB20}, and BlockDrop~\citep{DBLP:conf/cvpr/WuNKRDGF18}. They introduced additional policy networks responsible for deciding to skip certain layers or not. The main formula for those dynamic networks could be summarized as:
\begin{align}
 \text{SkipNet:}&\quad \bm{x}_{l+1} = z_l \* F_l(\bm{x}_l) + (1-z_l) \* \bm{x}_l\\
 \text{ConvNet-AIG:}&\quad \bm{x}_{l+1} = z_l \* F_l(\bm{x}_l) + \bm{x}_l\\
 \text{BlockDrop:}&\quad \bm{x}_{l+1} = z_l \* F_l(\bm{x}_l) +  \bm{x}_l
\end{align}
where $\bm{x}_l$ is the input of the $l$-th residual unit, $F_l(\cdot)$ is the network layers within the $l$-th residual unit except skip connection, and $z_l\in\{0,1\}$ is a binary value predicted by the policy network or the $l$-th policy module. By utilizing reinforcement learning or the Gumbel re-parameterization trick, the network can be trainable in an end-to-end way. 

Another research skipping-style line chooses to skip inputs given a long input sequence~\citep{DBLP:conf/acl/YuLL17, DBLP:conf/iclr/CamposJNTC18, DBLP:conf/iclr/Yu0S018} or assign fewer computations to unimportant steps~\citep{DBLP:conf/iclr/JerniteGJM17, DBLP:conf/iclr/SeoMFH18} or exit reading~\citep{DBLP:conf/acl/YuLL17, DBLP:journals/corr/abs-1807-02314, DBLP:conf/iclr/Yu0S018}. 
1) Skipping unimportant inputs is the natural way.
~\citet{DBLP:conf/iclr/CamposJNTC18} introduced Skip-RNN where a binary gate unit was used to learn to skip current input token or not. If the answer is yes, Skip-RNN copies current hidden state to the next time step, saving computations on those unimportant inputs.  LSTM-Jump~\citep{DBLP:conf/acl/YuLL17} achieved the same goal by directly predicting how many steps to jump through, or whether to exit reading inputs. 
Although skipping partial inputs saves computations largely, these models, like Skip-RNN and LSTM-Jump, suffer from missing or repeating outputs at skipped positions thus are not suitable for token-level tasks. 2) To address this problem, assigning fewer computations to unimportant steps is a flexible solution.
To this end, ~\citet{DBLP:conf/iclr/SeoMFH18} proposed Skim-RNN that dynamically decided to update the full-sized hidden state or partial-sized hidden state at each time step.
3) Exiting-style reading is a special kind of skipping reading~\citep{DBLP:conf/kdd/ShenHGC17,DBLP:conf/iclr/Yu0S018,DBLP:journals/corr/abs-1807-02314}. It decides to truncate the next inputs. 
For example, ~\citet{DBLP:journals/corr/abs-1807-02314} applied the exit mechanism to multi-task scenario. ReasoNet~\citep{DBLP:conf/kdd/ShenHGC17} adopted the exit mechanism for machine comprehension tasks. Despite good trade-off between accuracy and inference speed, skipping-style dynamic networks are harder to train, introducing more tuning overhead.

\textbf{Mixture-of-experts-style dynamic networks} are representative dynamic models~\citep{DBLP:conf/iclr/LepikhinLXCFHKS21, DBLP:journals/corr/abs-2103-00823, DBLP:journals/corr/abs-2101-03961}. In those models, a layer contains multiple experts and only part of these experts will be activated for each instance. For example, Switch Transformer~\citep{DBLP:journals/corr/abs-2101-03961} is the representative model that has trillion-level parameters. It replaces the normal 
feed-forward layer in the Transformer with a switch feed-forward layer, consisting of a routing module and multiple structure-identical experts. In each switch layer, only a single expert will be executed for each token. Compared to general dense computation architectures, mixture-of-expert-style networks provide an affordable and practical way to modify and train large models with sparse activation. 

\subsection{Deployment Sharing}

When deploying deep learning models on edge devices, we have to consider realistic constraints, such as storage, memory, computation, latency, and power consumption. Previous researchers have designed lightweight and compact models for mobile devices or other edge devices, such as MobileNets~\citep{DBLP:journals/corr/HowardZCKWWAA17, DBLP:conf/cvpr/SandlerHZZC18, DBLP:conf/iccv/HowardPALSCWCTC19}. 
However, with different hardware resources, the optimal neural network architecture varies significantly~\citep{DBLP:conf/iclr/CaiGWZH20}.  Thus, developing elastic or dynamic models to satisfy different constraints is critical for practical applications. 


In the recent two years, some studies have paid attention to efficient deployment. In these studies, a super-network is trained together with its massive sub-networks by task-specific losses. During inference, the appropriate sub-network is selected to satisfy the resource constraints. By amortizing the only-once training cost, the total cost of specialized designing is reduced from O(N) to O(1). During inference, the model can dynamically choose an appropriate network for different devices.  To be specific, 
\cite{DBLP:conf/iclr/YuYXYH19} proposed slimmable neural networks where several widths are predefined, supporting instant and adaptive accuracy-efficiency trade-offs by selecting corresponding width. Following  this work, \cite{DBLP:conf/iccv/YuH19} further proposed US-Nets to support arbitrary width selection.
\cite{DBLP:conf/iclr/FanGJ20} proposed an elastic network that can select sub-networks of any depth from one large network without having to finetune them.  
Beyond aforementioned studies, the temporal or input length is also an elastic selection. For example, \cite{DBLP:conf/acl/KimC20} proposed length-adaptive Transformer to support arbitrary progressively length deduction. The Length-adaptive Transformer can 
be directly adopted into the downstream task and satisfy any efficiency constraints by searching the corresponding length deduction configurations. 


\section{Compact-architecture Search}
In addition to model design, there are studies working on search efficient networks towards resource-constraint devices, like mobile. They borrow the idea of neural architecture search and apply it to design tiny networks.  For example, ~\citet{DBLP:conf/cvpr/TanCPVSHL19} proposed a neural architecture search approach, which explicitly incorporated model latency into the main objective so that the search can identify a model with  a good trade-off between accuracy and latency. On the ImageNet classification task, this approach achieved 75.2\% top-1 accuracy with 1.8x faster than MobileNet V2. ~\cite{DBLP:conf/iccv/HowardPALSCWCTC19} combined neural architecture search and network design together to develop a stronger mobile net MobileNet V3. ~\citet{DBLP:conf/iclr/CaiZH19} directly learned the architectures on the target task and hardware. ~\citet{DBLP:conf/cvpr/WuDZWSWTVJK19} proposed a differentiable neural architecture search framework that used gradient-based methods to optimize ConvNet architectures towards mobile devices.
\clearpage

\newpage
\chapter{Energy-Efficient Training}
\label{cha:training}

\begin{figure*}[thp]
  \centering
\begin{forest}
  forked edges,
  for tree={
    grow=east,
    reversed=true, 
    anchor=base west,
    parent anchor=east,
    child anchor=west,
    base=left,
    font=\footnotesize,
    rectangle,
    draw=hiddendraw,
    rounded corners, 
    align=left,
    minimum width=2.5em,
    minimum height=1.2em,
    s sep=6pt,
    inner xsep=3pt,
    inner ysep=1pt,
  },
  where level=1{text width=6em}{},
  where level=2{text width=7em,font=\scriptsize}{},
  where level=3{font=\scriptsize}{},
  where level=4{font=\scriptsize}{},
  where level=5{font=\scriptsize}{},
  [Energy-efficient Training 
    [Initialization
      [Random Initialization
        [Kaiming Initialization]
        [Xaiver Initialization]
        [Fixup Initialization]
        [LSUV Initialization]
      ]
      [Pre-trained Models \\ for Initialization
        [Feature based Initialization]
        [Fine-tuning based Initialization]
        [Supervised Initialization]
        [Self-supervised Initialization]
      ]
    ]
    [Normalization
      [Batch Normalization]
      [Layer Normalization]
      [Group Normalization]
     ]
     [Progressive \\ Training
     ]
     [Efficient AutoML
      [Search Space
      [Continuous]
      [Discrete]
      [Cell block]
      [Meta-architecture]
      ]
      [Search Method
      [RL-based Search]
       [Evolution-based Search]
       [Differentiable Search]
      ]
      [Evaluation Method
      [Early Stop]
      [Weight Sharing]
      [Hypernetworks]
      ]
     ]
    ]
\end{forest}
\caption{Taxonomy of energy-efficient training  with representative examples.}
\label{taxonomy_of_training_design}
\end{figure*}
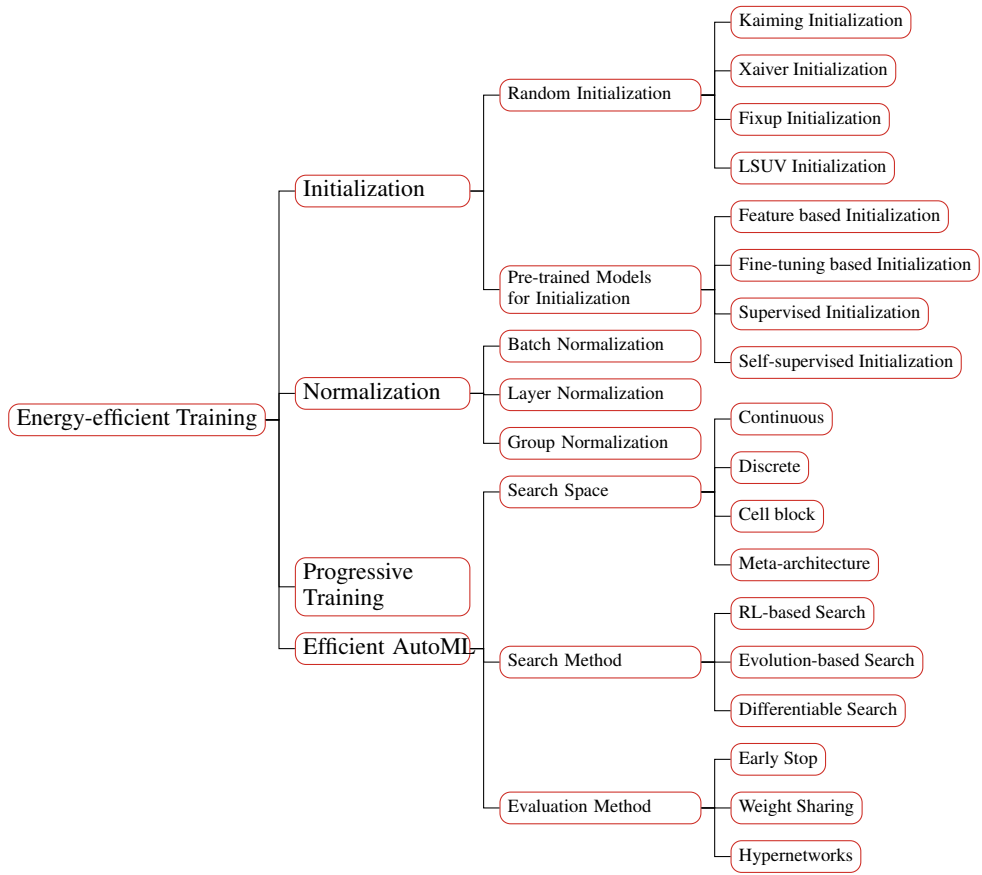

Many advanced approaches have been proposed to reduce training costs for deep learning. In Chapter 2, we describe efficient networks that can reduce computations in a single execution.  In this chapter, we  focus on the computations required during the whole training, including weight tuning and hyper-parameter tuning. To be specific, we survey approaches that aim to accelerate weight tuning/hyper-parameter tuning by using fewer iterations, including initialization,  normalization, progressive training, and efficient NAS. An overview is shown in Figure~\ref{taxonomy_of_training_design}. 

 \section{ Initialization}
 
The training of deep learning starts from architecture design and parameter initialization. We have explored efficient architecture design in Section 2. In this part, we focus on how weight initialization affects model training. 

\begin{table*}[ht]
\renewcommand\arraystretch{1.5}
\caption{Summarization of two common initialization approaches. $d$ and $u$ mean the dimensions of weight matrix $W$. $Uniform$ and $Normal$ mean the uniform distribution and Gassuian distribution. 
}
\label{tab:initialization}
\centering
\resizebox{\textwidth}{!}{
\begin{tabular}{|c|c|c|}
\hline \textbf{Initialization Approach} & \textbf{Description} & \textbf{Formulation} \\
\hline 

\makecell*[c]{Kaiming initialization\\ \citep{DBLP:conf/iccv/HeZRS15}} & 
\makecell*[c]{ Distribution of standard deviation of $\sqrt{\frac{2}{d}}$ } &
\makecell*[c]{$W \sim Normal(0,\frac{2}{d})$ or $W \sim Uniform(-\sqrt{\frac{6}{d}},\sqrt{\frac{6}{d}})$}   \\
\hline 

\makecell*[c]{Xaiver initialization\\ \citep{DBLP:journals/jmlr/GlorotB10}} & 
\makecell*[c]{ Distribution of standard deviation of $\sqrt{\frac{2}{d+u}}$  } &
\makecell*[c]{$W \sim Normal(0,\frac{2}{d+u})$ or $W \sim Uniform(-\sqrt{\frac{6}{d+u}},\sqrt{\frac{6}{d+u}})$} \\
\hline 



\end{tabular}}
\end{table*}

\subsection{Random Initialization}
It is widely accepted that good initialization of weights in a neural network is critical to convergence~\citep{DBLP:journals/jmlr/GlorotB10, DBLP:conf/nips/KrizhevskySH12, DBLP:conf/iccv/HeZRS15, DBLP:journals/corr/MishkinM15, DBLP:journals/corr/Kumar17}. At the beginning, deep networks are usually initialized via random weights drawn from uniform distributions or Gaussian distributions. Many previous studies found that these kinds of initialization failed in handling very deep models~\citep{DBLP:journals/jmlr/GlorotB10,DBLP:journals/corr/SaxeMG13, DBLP:journals/corr/RomeroBKCGB14,DBLP:conf/nips/HaninR18}. The problem is caused that the mean/variance of activations and gradients exponentially with the depth. To enable training with a very deep model, some advanced initialization solutions, like Kaiming initialization~\citep{DBLP:conf/iccv/HeZRS15}, Xaiver initialization~\citep{DBLP:journals/jmlr/GlorotB10}, LSUV initialization~\citep{DBLP:journals/corr/MishkinM15} and Fixup initialization~\citep{DBLP:conf/iclr/ZhangDM19}, have been proposed. The key idea is to normalize  the variance of weights to make the variance of activation in each layer to be around 1. We list the details of two widely-used initialization approaches in Table~\ref{tab:initialization}. Compared to the naive baseline, these initialization approaches can achieve better performance and faster convergence~\citep{DBLP:journals/corr/MishkinM15}.

\subsection{Pre-trained Models for Initialization}

In addition to random initialization, many approaches borrow models pre-trained from other domains (or other tasks) as initialization.  It is widely believed that initialization from existing models is an effective technique to improve the generalization ability with fewer training iterations. We split these pre-training initialization into different categories according to different dimensions. First, according to whether the borrowed parameters keep unchanged, these methods can be classified into feature-based initialization, and fine-tuning-based initialization. Second, according to the knowledge source of pre-trained parameters, these methods can be classified into supervised initialization and self-supervised initialization. 


\textbf{Feature-based initialization} borrows the parameters (usually from low-layers or mid-layers) as initialization from other domains/tasks while these parameters keep fixed during training. Generally speaking, feature based initialization can keep the generalization ability of the borrowed parameters better and thus is more suitable for  extremely few-shot settings.  

\textbf{Fine-tuning-based initialization} uses the target data to train all parameters, including new parameters and borrowed parameters.  Fine-tuning based initialization can further optimize the target objectives via fine-tuning all parameters and thus can better fit training data. It is the most popular solution nowadays for the NLP field.

\textbf{Supervised initialization} is widely investigated in the earlier stage of deep learning. A common solution is to pre-train the target model on similar tasks/datasets, and then reuse the pre-trained parameters as initialization for the target task~\citep{DBLP:conf/icassp/HuangLYDG13,DBLP:conf/cvpr/OquabBLS14,DBLP:conf/nips/YosinskiCBL14, DBLP:conf/acl/DuongCBC15, DBLP:conf/nips/LongZ0J16}. This solution is especially popular for low-resource settings and is extensively studied on domain adaptation/transfer learning.

The most representative example for supervised initialization is the pre-training of deep CNN backbones~\citep{DBLP:journals/corr/SimonyanZ14a,DBLP:conf/nips/RenHGS15,DBLP:conf/cvpr/HeZRS16,DBLP:journals/corr/SimonRD16,DBLP:conf/iccv/HeGDG17,DBLP:journals/corr/abs-1801-05746}. Fine-tuning pre-trained CNNs on different downstream datasets usually leads to improved performance compared to training from scratch and also reduces the number of training steps. Previous researches have explored advanced initialization methods. It is widely accepted that layers near to inputs usually are responsible to capture local features~\citep{DBLP:conf/eccv/ZeilerF14}. Therefore, many studies focus on transferring knowledge via initialization from low-layer features and mid-layer features. Interestingly, with the increase of large-scale training data, current trends directly adopt the simplest solution that uses all parameters for initialization~\citep{DBLP:conf/aaai/LiDFGJ20}.

Supervised initialization is also successfully applied to NLP~\citep{DBLP:conf/emnlp/SocherPWCMNP13}.  It is widely-accepted that features computed in higher layers of the network usually depend on the specific dataset and task. Following this belief, many studies  borrow the parameters of low-level layers and mid-level layers from other domains as initialization~\citep{DBLP:conf/acl/DongWHYW15,DBLP:journals/corr/LuongLSVK15,DBLP:conf/iclr/YangSC17,DBLP:conf/acl/StoyanovJLY18,DBLP:conf/aaai/LiuSRXG0018}. Recently, the trend of large-scale networks enables researchers to reuse all parameters from pre-trained networks. Similar with CV, the widely-adopt setting in NLP is directly reusing all parameters~\citep{DBLP:journals/tacl/JohnsonSLKWCTVW17,DBLP:conf/naacl/AharoniJF19,DBLP:conf/iclr/TanRHQZL19,DBLP:conf/emnlp/BapnaF19,DBLP:conf/emnlp/LinPWQFZL20}.

\textbf{Self-supervised initialization} is also a popular direction. With the increasing parameters in state-of-the-art DNNs, more and more training data are required to achieve better generalization results. To reduce the requirements of supervised data, previous studies investigated self-supervised pre-training that exploited unlabeled data to construct supervision signals to learn representations. Since self-supervised pre-training does not require any human-annotated labels, it is easy to get sufficient training data. To this end, researchers designed various methods to construct self-supervised training signals with unlabeled data. Here we take CV and NLP as examples to review recent self-supervised pre-training models.

For the NLP fields, using a model pre-trained on self-supervised data as initialization is the most popular solution. At the start, researchers use language modeling to pre-train word embeddings which are then used to initialize downstream word embeddings~\citep{DBLP:conf/emnlp/JoshiTPBC16,DBLP:conf/naacl/QiSFPN18,DBLP:journals/jair/RuderVS19}. Glove is one widely-used word embedding toolkit~\citep{DBLP:conf/emnlp/PenningtonSM14} which trains word embeddings based on global word-word co-occurrence counts. With the development of representation learning, researchers begin to explore and reuse contextualized models. Contextualized models define that the representation of a word depends on its contexts and each word has two representations, fixed word embeddings and contextualized representations.  ~\citet{DBLP:conf/naacl/PetersNIGCLZ18} proposed the first widely-used contextualized representations, ELMo.  Following this work, many advanced contextualized representation models begin to spring up, like BERT~\citep{DBLP:conf/naacl/DevlinCLT19}, GPT~\citep{DBLP:conf/naacl/SchickS21}, T5~\citep{DBLP:journals/jmlr/RaffelSRLNMZLL20}. 
The development of pre-trained networks also affect the application of CV. In recent years, CV began to explore large-scale self-supervised models for initialization~\citep{DBLP:conf/nips/LuBPL19,DBLP:conf/aaai/LiDFGJ20,DBLP:journals/corr/abs-2012-00364}. The learning objective is similar with NLP's pre-trained networks, either recovering masked/noised regions, or generating the original image from scratch.

Empirical results demonstrate that these pre-trained networks for initialization can achieve better performance and faster convergence. However, since current pre-trained networks generally require downstream tasks to use the exactly same networks, the training time still depends on architecture execution in addition to convergence speed.  Therefore, it should be considered case by case to conclude whether pre-trained models for initialization reduce downstream training costs in implementation.

\section{Normalization}

In addition to initialization approaches, normalization is another solution to accelerate training. Strictly speaking, normalization is a special component. Considering that it can accelerate convergence~\citep{DBLP:conf/nips/BjorckGSW18,DBLP:conf/nips/SanturkarTIM18,DBLP:conf/iclr/ZhangDM19}, we describe normalization in this chapter.

Normalization is a technique to normalize hidden outputs in deep neural networks. Batch normalization~\citep{DBLP:conf/icml/IoffeS15} is the first widely-used normalization for deep models. The key idea is to normalize the hidden vectors of neural networks to the distribution with mean $\mu=0$ and standard deviation $\sigma=1$. The hidden vectors usually are tensors and batch normalization is applied on the batch dimension. To be specific, it generates the output given the hidden output $h$:
\begin{equation}
    y_i = \frac{y_i-\mu}{\sigma + \epsilon}, \mu = \frac{1}{|B|}\sum_{i=1}^{|B|}h_{b,i}, \sigma = \frac{1}{|B|}\sum_{i=1}^{|B|}\sqrt{(h_{b,i}-\mu)^2}
\end{equation}
where $h$ is a intermediate tensor where the first is batch dimension. $|B|$ is the batch size. $y$ and $h$ are the output and input of the normalization component. ~\citet{DBLP:conf/icml/IoffeS15} fond that applied to a state-of-the-art image classification model, batch normalization achieved the same accuracy with 14 times fewer training steps, and beat the original model by a significant margin.

Following batch normalization, many normalization variants have been proposed, like layer normalization~\citep{DBLP:journals/corr/BaKH16}, group normalization~\citep{,DBLP:journals/ijcv/WuH20}, weight normalization~\citep{DBLP:conf/nips/SalimansK16}. These variants have almost the same calculation process except they are applied to different dimensions or different objectives.  

Despite good performance, it is still controversial where the benefits of normalization come. At the start, normalization is proposed to address internal covariate shift by normalizing layer inputs. Internal covariate shift is a phenomenon where the distribution of each layer’s inputs changes during
training. The parameters of the higher layer are required to continuously fit for the new distribution of lower layers, which slows down the training. To keep distribution steady, normalization is proposed to fix the distribution of input to a standard distribution. However, ~\citet{DBLP:conf/nips/SanturkarTIM18} overturned this belief and they found that the distributional stability of layer inputs had little to do with the success of batch normalization. Instead, normalization
makes the optimization landscape significantly smoother. This smoothness induces more predictive and stable gradients, allowing for faster training. Motivated by this paper, ~\citet{DBLP:conf/nips/Xu0ZZL19} proved that normalization indeed normalized backward gradients, which plays an important role in deciding the success of normalization.







\section{Progressive Training}
Progressive training is another strategy to effectively train DNNs. The key idea is constructively adding layers.   Compared to full training, progressive training does not require full gradients to all parameters, thus can largely reduce computations required for training. In addition, the well-trained lower layers also accelerate the training of higher layers.  ~\citet{DBLP:journals/neco/HintonOT06} applied progressive training to deep belief networks. They trained layers sequentially starting from bottom layers in a greedy, layer-wise fashion. It is based on an assumption that upper layers represent more ``abstract'' concepts
 whereas lower layers extract ``low-level features''.  This method is unsupervised because each layer learns a higher-level representation of the layer below and the training criterion does not depend on the labels. Following this work,  ~\citet{DBLP:conf/nips/BengioLPL06} extended this method to handle continuous inputs. 

With the development of deep learning, layer-wise progressive training methods are exploited to train CNNs~\citep{DBLP:conf/iccci/Rueda-PlataRG15,DBLP:journals/corr/KulkarniK17,DBLP:conf/icml/BelilovskyEO19} and RNNs~\citep{DBLP:conf/icmcs/XuSYTM18}.  Recently, ~\citet{DBLP:conf/icml/GongHLQWL19} and~\citet{DBLP:journals/corr/abs-2011-13635} extended the idea of layer-wise training to large-scale  NLP models by progressively stacking new layers on top of previously trained layers. Their experimental results show that layer-wise training can successfully improve the efficiency of training large transformer language models with huge amounts of data. To be specific, experimental results show that such progressive training policy can achieve more than 110\% training speedup without significant performance degradation.


\section{Efficient Hyper-parameter Optimization}
During training, hyper-parameter optimization (HPO) is a common and fundamental step for AI participants to find better model settings. Hyper-parameters keep fixed during training, including but not limited to optimization settings (e.g., learning rate, batch size) and model settings (e.g., the number of layers).  Since deep learning performs like a black-box model and the learning landscape is  non-convex, current optimization approaches usually find a random local minimum. 
Due to the uncertainty, AI engineers tend to taking a lot of computations to find better hyper-parameter settings on real-world applications ~\citep{DBLP:journals/corr/abs-2003-05689}. HPO or autoML is a field to automatically find the optimal settings. Considering that previous approaches mainly study architecture settings, we take efficient neural architecture search (NAS) as an example in this survey to review recent progress. Following previous studies, we split NAS methods into three components: search space, search strategy, and architecture evaluation. In this survey, we give an overview of efficient NAS. There are also surveys describing more details of NAS~\citep{DBLP:journals/corr/abs-1808-05377}.

\begin{figure}[ht]
    \centering
    \includegraphics[width=0.8\linewidth]{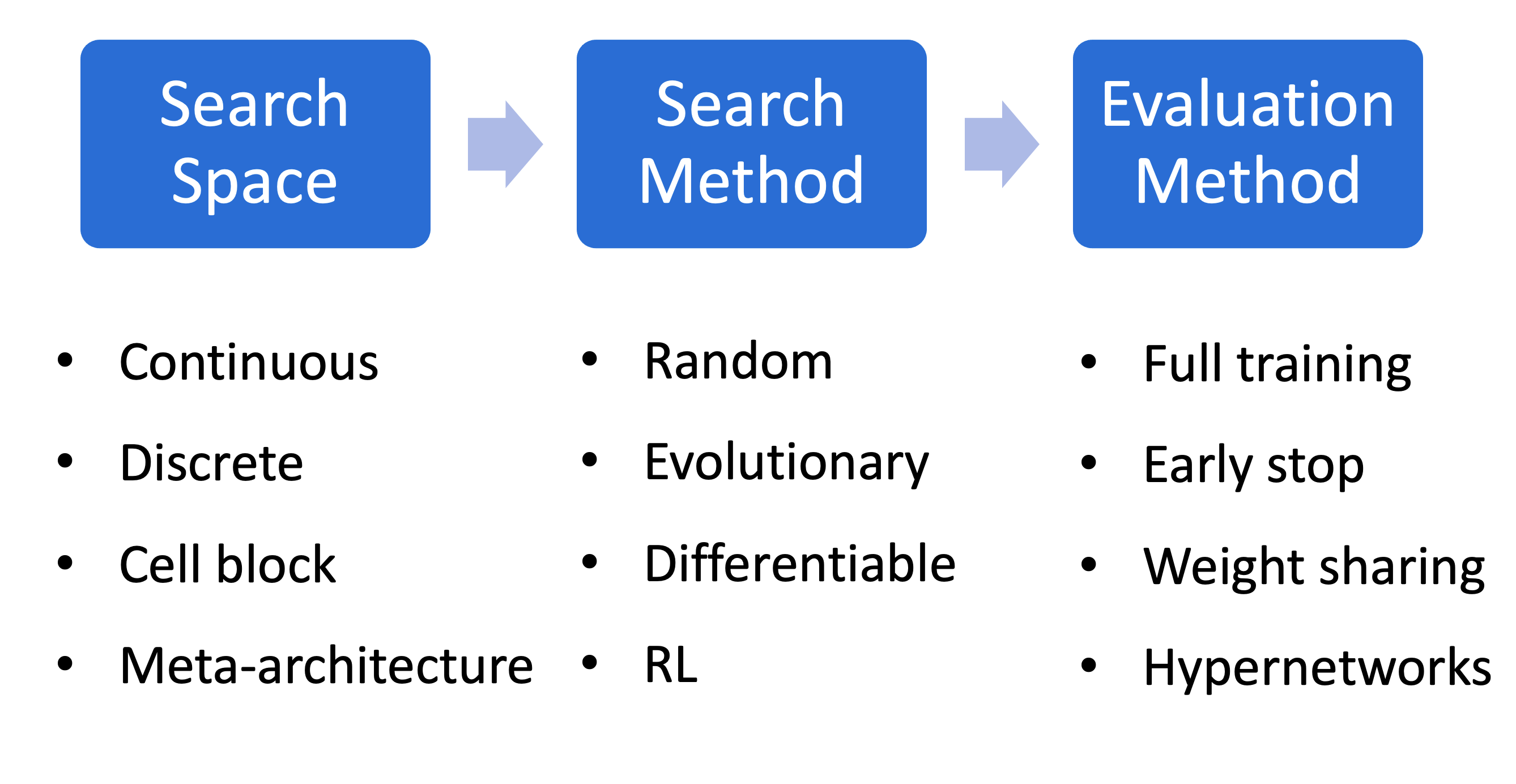}
    \caption{An overview of NAS components.}
    \label{fig:nas}
\end{figure}

The search space defines all architecture candidates. At the first, the search space is defined as a discrete space, including structured space or unstructured space. Considering that network candidates in unstructured space is too massive, researchers usually incorporate inductive bias to build a structured search space~\citep{DBLP:conf/eccv/LiuZNSHLFYHM18,DBLP:conf/iclr/Dong020,DBLP:conf/iclr/Shu0C20}. One of the representative methods is cell-based search space. Cell-based search space assumes that each architecture contains repetitions of fixed structures. In this way, the search space can be limited to the cell space where the number of candidates is largely reduced. In addition, to enable faster search, differentiable approaches~\citep{DBLP:conf/emnlp/JiangHXZZ19} adopt a continuous search space where edge weights are considered.

The search strategy defines a policy to explore search space. \textit{Random search} is one of traditional search approaches. The key idea is to randomly evaluate architectures and to select the best one based on their validation results. To reduce wasted evaluation costs,  researchers proposed \textit{reinforcement learning based search} policies~\citep{DBLP:conf/icml/YingKCR0H19}. It is a direction that introduces an architecture generator to generate well-performing architectures. Since random search and reinforcement learning require validation accuracy as search criterion, these methods usually need expensive computations.  To reduce search costs, \textit{evolution based search} has been proposed~\citep{DBLP:conf/aaai/RealAHL19}.  It is a two-stage search approach. The first stage selects several well-performing parent architectures. The second stage applies mutation on these parent architectures to select the best one. The second stage starts from pre-trained parent networks and does not require too much computations to train child networks.  Recently, ~\citet{DBLP:conf/icml/SoLL19} applied the evolution search on Transformer networks and achieved new state-of-the-art results on machine translation and language modeling tasks. Although these approaches can reduce exploration costs, the dependence on validation accuracy still leads to considerable computation costs. To fully get rid of the dependence on validation accuracy, several studies~\citep{DBLP:conf/emnlp/JiangHXZZ19,DBLP:conf/iclr/LiuSY19,DBLP:conf/cvpr/DongY19,DBLP:conf/iclr/ZelaESMBH20,DBLP:journals/corr/abs-2005-03566} proposed \textit{differentiable search} that re-formulated the task in a differentiable manner and allowed efficient search using gradient descent. In addition, another research line aims to represent a model into a continuous space where there is a mapping between structures and results. In this way, the model only learns how to predict the performance of architectures based on their continuous representations where the downstream training is not required~\citep{DBLP:conf/nips/LuoTQCL18}.
To further reduce training costs, researchers proposed training-free NAS approaches that directly extracted features from randomly-initialized models and used these features as evaluation criterion to select networks~\citep{DBLP:journals/corr/abs-2006-04647,DBLP:journals/corr/abs-2102-11535,DBLP:journals/corr/abs-2101-08134,DBLP:conf/icml/XuZLG0Y21}.  

Architecture evaluation takes almost all computations in NAS approaches. At the first, full training is required to evaluate the performance of a network, which is very heavy. Early stop is a widely used trick to estimate the results of a network. Besides, parameter-sharing is also a popular solution that network candidates can share parameters with each other~\citep{pham2018efficient}. In this way, the model can reuse pre-trained blocks during downstream training. 

\paragraph{Discussion} Current NAS solutions are well-explored in the CV field, and widely-used benchmarks are also based on CV datasets. In the future, it is a promising direction to apply NAS in other fields, like NLP, to address more real-world problems. In addition, existing models focus more on models towards a single task for simplification. The multi-task/domain/lingual model is attracting more attention. Therefore, how to use NAS to search a shared multi-task/domain/lingual model is also a promising direction. Furthermore, the essential question of NAS is how model architecture affects downstream results. More understanding studies are expected to reveal the fundamental connection between architecture and performance.








\clearpage


\chapter{Energy-Efficient Inference}
\label{cha:infer}



\begin{figure*}[ht]
  \centering
\begin{forest}
  forked edges,
  for tree={
    grow=east, 
    reversed=true, 
    anchor=base west, 
    parent anchor=east, 
    child anchor=west, 
    base=left, 
    font=\footnotesize,
    rectangle,
    draw=hiddendraw,
    rounded corners,align=left,
    minimum width=2.5em,
    minimum height=1.2em,
    s sep=6pt,
    inner xsep=3pt,
    inner ysep=1pt,
  },
  where level=1{text width=5em}{},
  where level=2{text width=5.8em,font=\scriptsize}{},
  where level=3{font=\scriptsize}{},
  where level=4{font=\scriptsize}{},
  where level=5{font=\scriptsize}{},
  [Efficient Inference
    [Pruning
      [Pruning Unit
        [Unstructured
          [Neurons{,} Connections]]
        [Structured
          [Filters{,} Channels{,} Layers]]
      ]
      [Scoring Function
        [Magnitude]
        [Important Coefficients]
        [Gradient-based]
        [Movement Pruning]
      ]
      [Scheduling
        [Single-step Pruning]
        [Iterative Pruning]
        [Lottery Ticket]
      ]
    ]
    [Low-rank \\
    Factorization
      [Matrix Factorization
        [Low-rank Matrix Factorization{,} SVD]]
      [Tensor Factorization
        [CP{,} VBMF{,} Tucker Decomposition{,} BTD]]
    ]
    [Quantization
      [Deterministic \\ Quantization
        [Rounding{,}  Vector Quantization]]
      [Stochastic \\ Quantization
        [Random Rounding{,} Probabilistic Quantization]]
    ]
    [Knowledge\\
    Distillation
      [Distillation Target
        [Logits-based (Vanilla KD)]
        [Feature-based]
        [Relation-based]
      ]
      [Teacher Numbers
        [Dynamic KD{,} Multi-teacher KD{,} Mutual Learning]]
    ]
  ]
\end{forest}
\caption{Taxonomy of efficient inference methods with representative examples.}
\label{taxonomy_of_PTMs}
\end{figure*}
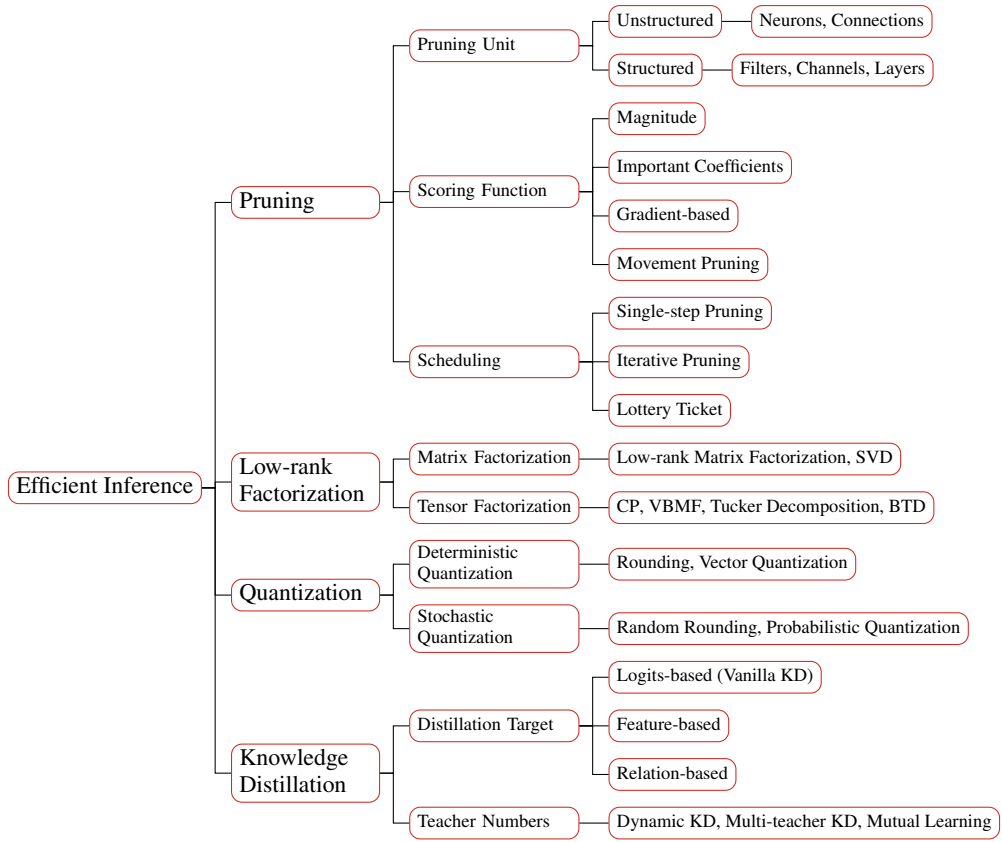

In this chapter, we describe common network surgery methods for reducing inference costs, including pruning, low-rank factorization, quantization, and knowledge distillation.  A brief review of these methods is presented in Figure~\ref{taxonomy_of_PTMs} and Table~\ref{tab:compression_sum}. 




\begin{table*}[ht]
\caption{Different approaches for efficient inference.}
\label{tab:compression_sum}
\centering
\resizebox{\textwidth}{!}{
\begin{tabular}{|c|c|c|}
\hline Approaches & Descriptions & Characteristics \\
\hline Pruning &  Reduce redundant parameters which  & Can be applied to various settings.  \\
 & are not sensitive to results.  & Fine-tuning is optional. \\
\hline Low-rank Factorization & Use matrix/tensor decomposition to  & Matrix decomposition is computationally\\
& approximate the original parameters. & complicated, but can support train  from scratch.\\
\hline Quantization & Reduce the number of 
bits used to & Easy to implement. \\
& represent weights and activations. & Sensitive to hardware architecture. \\
\hline Knowledge Distillation & Train a compact neural network with & Easy to implement. \\
& knowledge distilled from a teacher model. & Sensitive to network parameters.\\
\hline
\end{tabular}}
\end{table*}

\section{Model Pruning}

Model pruning is a popular solution to reduce redundant parameters in DNNs. Back to 1980s,~\citet{DBLP:conf/nips/HansonP88} and~\citet{DBLP:conf/nips/CunDS89} already verified that parameters are not equally important to the final performance. By removing unimportant weights from a network, we can simultaneously reduce the number of parameters, accelerate training/inference, save training examples, and improve generalization. This motivates a great amount of studies on pruning neural networks in the past 30 years. Specifically, given an initial network that is large and accurate, the key idea of pruning is to remove parameters from the original network to produce a smaller network that can retain the accuracy of the original model.


We first provide a formal definition of pruning. Let us define a neural network model as $f(X;\theta)$, which is a function over an input set $X$ and the set of parameters $\theta$. Pruning approaches usually take a model $f(X;\theta)$ as input and then produce a new model $f(X;\theta')$. $\theta'$ is a set of parameters with the size of $\theta'$ being less than that of $\theta$. Usually, $\theta'$ is a subset of $\theta$.

\begin{algorithm}[ht]
\caption{The generic framework of pruning}
\label{alg:prune-after-training}
\begin{algorithmic}
\REQUIRE $N$ is the number of iterations of pruning; $X$ is the dataset
    \STATE $\theta' \gets \text{train-to-convergence}(f(X; \theta))$

    \FOR{$i$ \text{ }in $1$ to $N$}%
        \STATE $\theta' \gets \text{prune}(score(\theta'))$%
        \STATE $ \theta' \gets \text{fine-tune}(f(X; \theta'))$%
    \ENDFOR
    \STATE \textbf{return} $\theta'$
\end{algorithmic}
\end{algorithm}

Previous pruning approaches mainly follow the work of ~\citet{DBLP:conf/nips/HanPTD15} to produce a pruned model $f(X;\theta')$ from an original model $f(X;\theta)$. We show the generic framework in Algorithm~\ref{alg:prune-after-training}. First, the network is trained to convergence to get pre-trained parameters. Second, each parameter or structural element in the network is assigned a score. This score indicates the relative importance to the final performance. The network is then pruned based on these scores. Third, since pruning generally reduces the accuracy of the network, it is a general practice to fine-tune (train further after pruning) the pruned network. The process of pruning and fine-tuning is usually iterated several times.

 We describe some key components of network pruning algorithms:
\begin{itemize}
    \item \textbf{Pruning Unit} refers to the basic unit that the algorithm aims to prune.  According to pruning units, we can classify modern pruning approaches into two categories, unstructured pruning, and structured pruning. Some pruning algorithms prune individual parameters (i.e., unstructured pruning), which produce a sparse neural network. While the resulting sparse network is smaller in terms of the number of parameters, it is hard to yield speedups since the pruned weights are not well arranged. In contrast, structured pruning considers parameters in groups. They keep the dense features of the network by removing entire weight matrices, filters, channels, or layers.
    
    \item \textbf{Scoring Function} defines the metric used to prune parameters. Common practices for parameter scoring usually are based on importance coefficients, network activations, or gradients.  After assigning scores to each part of the parameters, we have two choices to prune networks. First, we can choose to prune a fraction of the parameters with the locally lowest scores within each structural sub-component of the network (e.g., layers). Second, we also can choose parameters with the globally lowest scores within the entire network.
    
    \item \textbf{Scheduling} decides the total step that pruning algorithms use to prune parameters. Some methods prune weights in a single step while another methods use multiple steps to prune parameters where each step only prunes a part of parameters.
    
    \item \textbf{Fine-tuning} is usually required for the pruned network to recover the original accuracy. Many methods choose to fine-tune the pruned network, or re-train the pruned network. 
\end{itemize}

\paragraph{Application} Pruning is first applied to fully-connected networks. For example, ~\citet{DBLP:conf/nips/CunDS89} analyzed the importance of  parameters and showed that small magnitude weights had less impact on training losses. Specifically, they computed the saliency of parameters based on the second derivative. Then, they pruned parameters with small saliency scores. To recover the original performance of the network, the network was fine-tuned after pruning. ~\citet{DBLP:conf/nips/HassibiSW93} extended this idea by using the inverse of the Hessian as saliency score.  In addition to weight pruning,~\citet{DBLP:journals/npl/SuzukiHS01} proposed to prune network connections based on their influence on training losses, and then re-train the network to compensate the performance drop. Unlike these approaches, ~\citet{DBLP:conf/bmvc/SrinivasB15} argued that similar neurons were redundant. They proposed to remove redundant neurons,  instead of removing individual weight connections one by one. 

Currently, many pruning algorithms are applied to CNNs. ~\citet{DBLP:conf/nips/HanPTD15} proposed a simple magnitude-based method to remove unimportant connections in fully-connected layers and convolutional layers.  However, the resulting model, despite being sparse, does not bring significant inference speedups due to the feature of sparsity.  To address this problem, several structured pruning algorithms have been proposed to prune dense blocks, like filters, channels, or layers~\citep{DBLP:conf/iclr/0022KDSG17,DBLP:conf/iclr/MolchanovTKAK17,DBLP:conf/iccv/HeZS17,DBLP:conf/nips/LinRLZ17,DBLP:conf/eccv/HeLLWLH18,DBLP:journals/pami/LuoZZXWL19}.

In addition to CV models, pruning has been successfully applied to NLP tasks. At the early stage, several studies successfully pruned recurrent neural networks (RNNs). ~\citet{DBLP:conf/conll/SeeLM16} used iterative pruning and retraining to prune a recurrent model for neural translation. ~\citet{DBLP:conf/iclr/NarangDSE17} pruned RNNs via magnitude based pruning. ~\citet{DBLP:journals/corr/abs-1711-02782} used iterative ground lasso regularization to induce block sparsity in RNNs. ~\citet{DBLP:conf/iclr/LeeAT19} and~\citet{DBLP:conf/iclr/ZhangS20} proposed one-shot RNN pruning methods based on connection sensitivity and Jacobian spectrum.  More recently, with the success of Transformer, several studies investigated pruning transformer models~\citep{DBLP:conf/nips/MichelLN19,DBLP:conf/acl/VoitaTMST19,mccarley2019structured,DBLP:conf/iclr/FanGJ20,DBLP:conf/emnlp/WangWLT20,DBLP:conf/nips/Sanh0R20}. One trend is to use structured pruning since the transformer architecture is highly parallelized. For instance,~\citet{DBLP:conf/iclr/FanGJ20} introduced LayerDrop to prune transformer layers for efficient inference. ~\citet{DBLP:conf/nips/MichelLN19} and ~\citet{DBLP:conf/acl/VoitaTMST19} revealed that the multi-head attention mechanism in the transformer architecture led to redundant attention heads. Motivated by this finding, they proposed to directly prune attention heads. ~\citet{mccarley2019structured} used structured pruning to compress a BERT-based question answering model. ~\citet{DBLP:conf/emnlp/XuZGWZ20} recently proposed a progressive module replacing approach by replacing a whole module from the original model with a compact module to reduce model size. Unlike these studies, ~\citet{DBLP:journals/corr/abs-2012-07463} proposed an unstructured pruning approach, diff pruning, to compress a multi-task model.  



While most aforementioned pruning algorithms require re-using or re-training the originally trained network, a recent research direction~\citep{DBLP:conf/iclr/FrankleC19} suggested that dense, randomly-initialized, feed-forward networks contained sub-networks (winning tickets), which can reached the test accuracy of the original network. They also found that a standard pruning technique naturally uncovered some of the winning tickets and then proposed an algorithm to identify winning tickets at the early stage of training. Also, ~\citet{DBLP:conf/iclr/FrankleC19} and~\citet{DBLP:conf/iclr/LiuSZHD19} found that once the ``winning ticket'' is found, it can be trained from scratch to get an equivalent or better performance compared to pruned and fine-tuned one. Recently, there are also several studies investigating the lottery ticket hypothesis for BERT-like models~\citep{DBLP:conf/emnlp/PrasannaRR20,DBLP:conf/nips/ChenFC0ZWC20}. For example,  ~\citet{DBLP:conf/emnlp/PrasannaRR20} found that with structured pruning, the ``random” sub-networks are still almost as good as the ``good" ones, and even the ``worst" ones perform on par with a strong baseline. 

\paragraph{Discussion}  Pruning is very effective for reducing the number of parameters in DNNs. With structured pruning, it can accelerate inference and reduce computations. However, there are also several limitations of pruning methods. First, it requires iteratively scoring weights and re-training the network for many iterations. Also, pruning often leads to a non-negligible performance drop when applied to powerful DNNs. The lottery ticket hypothesis provides an interesting direction for more efficient ``pruning'' algorithms.  In this survey, we give an overview of pruning methods. If you are interested in this direction, there are several surveys describing more details of pruning~\citep{DBLP:journals/corr/abs-2101-09671}.

\section{Low-rank Factorization}


Tensor (including matrix) operation is the basic building block and contributes the bulk of most computations and parameters in DNNs. Therefore, compressing tensors or matrices in DNNs is promising for reducing the number of parameters and computation costs. The motivation of low-rank factorization is that a large amount of redundant information exist in large weight matrices. The super-large matrices are generally of low rank and can be factorized into several tiny matrices to save parameters. For example, we can apply singular value decomposition (SVD) to factorize a super large matrix.  SVD factorizes the original weight matrix into three smaller matrices. Formally, for any matrix $A \in \mathbb{R}^{m \times n}$, there exists a factorization, $A = USV^{T}$ where $U \in \mathbb{R}^{m \times r}$ and $V^{T} \in \mathbb{R}^{r \times n}$ are orthogonal matrices, $S \in \mathbb{R}^{r \times r}$ is a diagonal matrix with only singular values of A on the diagonal. With SVD decomposition, the spatial complexity can be reduced from $\mathcal{O}(mn)$ to $\mathcal{O}(r(m + n + 1))$.



\paragraph{Application}  Similar to pruning algorithms, low-rank factorization is successfully applied to CV models. ~\citet{DBLP:conf/icassp/SainathKSAR13}  applied matrix factorization to the final weight layer. If the original weight matrix has the dimension $m \times n$ and rank $r$,  the full rank matrix can be factorized in two weight matrices as $m \times r$ and $r \times n$. Their approach reduced the number of parameters and achieved up to 30–50\% speedup. Similarly,~\citet{DBLP:conf/icassp/XueLYSG14} proposed to use SVD decomposition to compress fully-connected neural networks. ~\citet{DBLP:conf/cvpr/RigamontiSLF13} proposed to approximate trained CNNs with low-rank filters. ~\citet{DBLP:conf/nips/DentonZBLF14} further exploited the linear structure present within the convolutional filters. Their approach was able to reduce the memory requirement of the weights in the first two convolutional layers by 2–3 times.  While low-rank matrix factorization can optimize both the spatial and computational complexity of neural networks, the plane view of a matrix limits the potential for extreme compression.

Tensor factorization algorithms, in contrast, are more flexible and can be employed to achieve an extremely high compression ratio. Among popular tensor factorization methods, classical prolongation (CP)~\citep{DBLP:journals/siamrev/KoldaB09}, where each factor matrix has the same rank and the kernel tensor is a superdialognal tensor, generally achieves better compression performance. ~\citet{DBLP:journals/corr/LebedevGROL14} leveraged CP decomposition to compress CNN weight kernels into several sub-kernels to reduce the number of parameters. Specifically, non-linear least squares were used to compute low-rank CP-decomposition of the 4-D tensor into a sum of rank-one tensors. Using this decomposition, the original 
convolutional layer was replaced by a sequence of 4 convolutional layers with smaller filters. Following this idea, ~\citet{DBLP:journals/corr/KimPYCYS15} introduced a one-shot compression method to compress the whole network. 
In addition, ~\citet{DBLP:conf/ijcai/ChenJKFY18} introduced a collective residual unit based on block term decomposition (BTD), which is a combination of Tucker and CP, to enhance the utilization of parameters in residual CNNs.  ~\citet{DBLP:journals/spic/ZhouLLCZ19} conversely used neural networks to learn an appropriate CP rank for tensor decomposition. 



Apart from the applications on CV models, low-rank factorization has also been applied to NLP models. For example,~\citet{DBLP:conf/premi/GrachevIS17} used low-rank factorization to train RNN language models. ~\citet{DBLP:journals/corr/abs-1908-09982} investigated the use of low-rank factorization as an effective post-processing compression method for LSTMs. They applied low-rank factorization on ELMo, one of  widely-used pre-trained models.  Recently, low-rank factorization has also been applied on Transformer models~\citep{DBLP:conf/nips/MaZZDHZ019}. ~\citet{DBLP:conf/ijcnlp/NoachG20} further proposed a two-stage model-compression method to reduce the inference time cost of BERT, a kind of Transformer-based model. Their approach decomposed the matrices into smaller matrices and then performed feature distillation on the internal representation. Also,~\citet{DBLP:conf/iclr/LanCGGSS20} applied embedding matrix factorization along with layer sharing to reduce the amount of parameters.

\paragraph{Discussion}  Compared with other popular compression methods, low-rank factorization can effectively reduce the size of models with a large compression ratio while preserving the performance well. Low-rank factorization is also relatively flexible. However, low-rank factorization also suffers from the issue of computational efficiency because SVD over large weight matrices can be computationally heavy. Also, compared with the compression ratio in terms of model size, low-rank factorization is less effective for reducing the computational cost and inference time.

\section{Quantization}

The goal of quantization is to compress the original network by reducing the number of bits. The idea of network quantization can be back to early 1990s~\citep{fiesler1990weight,DBLP:journals/nn/BalzerTOK91,229903}. Recently, due to the success of DNNs and their growing sizes, the research of quantization has received increasing attention. In the beginning of 2010s, ~\citet{vanhoucke2011improving} discovered that CNNs encoded with 32-bit can be converted to CNNs encoded with 8-bit, which significantly reduced both storage and computation costs.

Generally speaking, quantization techniques can be classified into two types: \textit{deterministic quantization} and \textit{stochastic quantization}. In deterministic quantization, there is an deterministic mapping between the quantized value and the real value. In stochastic quantization, the quantized value is sampled from discrete distributions~\citep{DBLP:journals/corr/abs-1808-04752}. Usually, post-training quantization is the most simplest solution by applying quantization on a trained model to reduce inference costs. Despite simple, post-training quantization may brings dropped performance. \textit{quantization-aware training} is proposed to address this problem by fine-tuning the quantized model before inference.


\paragraph{Deterministic Quantization} defines a deterministic mapping between real weights and quantized weights. \textit{Rounding quantization} is the simplest mapping function. The key idea of rounding quantization is to map a high-bit floating-point number to its nearest fixed-point low-bit number~\citep{DBLP:conf/icml/GuptaAGN15}. For example, suppose a number $x$ and the target fixed-point representation $\textnormal{[IL, FL]}$. The number of integer bits $\textnormal{IL}$ plus the number of fractional bits $\textnormal{FL}$ yields the total number of bits used to represent the number. This approach considers the following rounding scheme:
\begin{equation}
\label{eq:quant1}
\textnormal{Convert}(x, \textnormal{[IL, FL]}) =  \begin{cases}
-2^{\textnormal{IL-1}} \hskip5em  \textnormal{if}\ x \leq -2^{\textnormal{IL-1}},\\
2^{\textnormal{IL-1}} - 2^{-\textnormal{FL}} \hskip2.6em \textnormal{if}\ x \geq  2^{\textnormal{IL-1}} - 2^{-\textnormal{FL}}, \\
\textnormal{Round}(x) \hskip3.4em\  \textnormal{otherwise}
\end{cases}
\end{equation} 
where

\begin{equation}
\label{eq:quant}
\textnormal{Round}(x) =  \begin{cases}
\floor{x} \hskip2.6em \textnormal{if}\ \floor{x} \le x \le \floor{x} + \frac{\epsilon}{2} ,\\
\floor{x}+\epsilon  \hskip1em \textnormal{if}\ \floor{x}+\frac{\epsilon}{2} < x \le \floor{x} + \epsilon \\
\end{cases}
\end{equation} 
where $\epsilon(=2^{-\textnormal{FL}})$ is the smallest positive number that can be represented in this fixed-point format, $\floor{x}$ is defined as the largest integer
multiple of $\epsilon$ smaller than $x$. Following this approach, more advanced approaches have been proposed~\citep{DBLP:conf/eccv/RastegariORF16,DBLP:conf/iclr/PolinoPA18,DBLP:conf/iclr/WuLCS18}.


In addition to scalar quantization for individual numbers in weight vectors, there is also a research line focusing on \textit{clustering-based quantization}. ~\citet{DBLP:journals/corr/GongLYB14} proposed  to classify weights into groups and to use the centroid of each group to replace the actual weights during inference. ~\citet{DBLP:journals/corr/HanMD15} employed a similar approach but fine-tuned the quantized centroids for better performance. Following this idea, ~\citet{DBLP:conf/iclr/ChoiEL17} further proposed a Hessian weighted k-means clustering approach.

\paragraph{Stochastic Quantization} does not define  one-to-one mapping from real weights to quantized weights. In random rounding, the quantized value is sampled from a discrete distribution parameterized by real values. For example,~\citet{DBLP:conf/nips/CourbariauxBD15} proposed the following random rounding function:
\[
x^b = \begin{cases}
+1\quad \textnormal{with probability}\ p = \sigma(x),\\
-1\quad \textnormal{with probability}\ 1 - p \\
\end{cases}
\]
where $\sigma$ is the ``hard sigmoid" function:
\begin{equation*}
\sigma(x) = \textnormal{clip}(\frac{x+1}{2}, 0, 1) = \max(0, \min(1, \frac{x+1}{2}))
\end{equation*}
If $x$ is a positive value, we will have a high probability to quantize it to $+1$, otherwise to $-1$. This gives us a more flexible quantization scheme. In probabilistic quantization, the weights are assumed to be discretely distributed and a learning algorithm is used to infer the parameters of the distributions.~\citet{DBLP:conf/nips/SoudryHM14} proposed an expectation back-propagation algorithm to train neural networks with binary or ternary weights. They first assumed some discrete prior distribution on the weights and then updated the weights in an online setting based on the Bayesian formula.

\paragraph{Quantization-Aware Training} Recently, quantization-aware training~\citep{DBLP:conf/cvpr/JacobKCZTHAK18,DBLP:conf/bmvc/DongLN17,DBLP:journals/corr/abs-2004-07320} has become the \textit{de facto} approach towards designing robust quantized models. It simulated quantization effects in the forward pass of training and the backward pass was accomplished via straight through estimator~\citep{DBLP:journals/corr/BengioLC13}. It generally relied on techniques like gradient clipping to make the training stable. Recently, several studies analyzed and introduced new quantization-aware training approaches. For example,~\citet{DBLP:journals/corr/abs-2004-07320} and~\citet{DBLP:conf/bmvc/DongLN17}  stochastically applied quantization to a portion of the weights at each training step, while~\citet{DBLP:journals/corr/abs-1803-08607} and~\citet{DBLP:conf/iclr/AlizadehFLG19} re-ordered the blocks or layers.


\paragraph{Applications} Network quantization is first widely applied to CNNs. In addition to CNNs, quantization has been also applied to other models, like RNN, Transformer. ~\citet{DBLP:journals/corr/OttLZLB16a} first investigated RNN quantization and found that weight binarization did not work well on RNNs. For simplification, they proposed to apply weight quantization for RNN weights and to leave activations as floating-point numbers.~\citet{DBLP:journals/jmlr/HubaraCSEB17} explored different combinations of bit-widths for weights and activations. ~\citet{DBLP:journals/corr/HeWZWYZZ16} proposed to quantize the structure of gates and interlinks in LSTM and GRU cells. Recently,~\citet{DBLP:conf/nips/WangXDLWX18} proposed to use a threshold ternary quantization method for weight quantization and a Bernoulli ternary quantization method for activation quantization. 

With the recent success of Transformer, a number of studies investigated the application of quantization on Transformers. For example,~\citet{DBLP:journals/corr/abs-1906-00532} and~\citet{DBLP:conf/emnlp/PratoCR20} showed that 8-bit quantization can successfully reduce the size of a Transformer-based model and accelerate inference without compromising translation quality. Recently, quantization has been applied on Transformer-based language models~\citep{DBLP:journals/corr/abs-1910-06188,DBLP:conf/emnlp/ZhangHYSCJL20,DBLP:journals/corr/abs-2012-15701,DBLP:journals/corr/abs-2101-01321}. ~\citet{DBLP:journals/corr/abs-1910-06188} first applied 8-bit quantization on BERT. Following this work,~\citet{DBLP:journals/corr/abs-2101-01321} proposed I-BERT, which employed lightweight integer-only approximation methods for nonlinear operations to quantize BERT. ~\citet{DBLP:conf/emnlp/ZhangHYSCJL20} proposed TernaryBERT, which ternarized the weights in a fine-tuned BERT model with both approximation-based and loss-aware ternarization methods. 

Very recently, several studies have investigated the application of quantization on GNNs. ~\citet{DBLP:conf/ictai/FengWLYPD20} proposed a GNN-tailored quantization algorithm, and used an automatic bit-selecting approach to pinpoint the most appropriate quantization bits.  ~\citet{DBLP:journals/www/WangLZQHLL21} and~\citet{DBLP:journals/corr/abs-2012-15823} further proposed binarized GNNs. ~\citet{tailor2021degreequant} proposed Degree-Quant, an architecture-agnostic method for quantization-aware training on graphs. Moreover,~\citet{DBLP:journals/corr/abs-2009-09232} proposed to use neural architecture search to find the optimal architecture coupled with the best quantisation strategy for different components in GNNs.

\paragraph{Discussion}  Quantization is very effective for reducing the size of neural networks.  However, post-training quantization often leads to non-neglieable performance drop. In contrast, quantization-aware training can effectively reduces the performance drop. Incorporating knowledge distillation for quantization can also improve the performance of quantized models. In this section, we give an overview of quantization. If you are interested in more details, please refer to surveys ~\citep{DBLP:journals/corr/abs-1808-04752,DBLP:journals/corr/abs-2103-13630}.

\section{Knowledge Distillation}


The idea of knowledge distillation (KD) is exploiting the knowledge inside a large trained ``teacher'' model to help the training of a ``student'' model~\citep{DBLP:conf/kdd/BucilaCN06,DBLP:conf/nips/BaC14,DBLP:journals/corr/HintonVD15}. In this way, we can use a smaller student model to distill a trained model as a replacement for inference.  

The traditional KD solution is to minimize the difference between the output produced by the teacher model and that produced by the student model. Formally, given a labeled dataset $\mathcal{D}$ of $N$ samples $\mathcal{D} = \{\left(x_1, y_1\right), \dots, \left(x_N, y_N\right)\}$, we can write the loss function of the student network during the process of knowledge distillation as follows:
\begin{equation}
    \mathcal{L}_{S}\left(\mathcal{D};  \theta_{S}; \theta_{T} \right)= \frac{1}{N} \sum_{i=1}^N \left[ \alpha\mathcal{L}_{\mathcal{T}}\left(y_i,  S\left(x_i; \theta_{S} \right) \right) + \left(1 - \alpha\right) \mathcal{L}_{\mathit{KD}}\left(T\left(x_i; \theta_{T}\right),S\left(x_i; \theta_{S} \right)\right) \right] \label{eq:loss}
\end{equation}
where $\alpha$ is a hyper-parameter to control the relative importance of the two terms; $\theta_{T}$ and $\theta_{S}$ are the parameters of teacher $T$ and student $S$, respectively. $\mathcal{L}_{\mathcal{T}}$ refers to the task-specific loss 
and $\mathcal{L}_{\mathit{KD}}$ refers to the knowledge distillation loss which measures the similarity of the student and the teacher.

In general, KD exploits the knowledge from the teacher model to help train the student model by minimizing the discrepancy between the knowledge in the teacher model and that in the student model.   According to the source of teacher knowledge, we can classify KD approaches into three categories: \textit{logits-based KD}, \textit{feature-based KD}, and \textit{relation-based KD}.  According to teacher types, we also can classify KD approaches into three categories: \textit{KD with static teacher}, \textit{KD with multiple teachers}, and \textit{KD with dynamic teacher}. 

\paratitle{Logits-based KD} focuses on the output class distribution of the teacher model, also referred as ``soft labels''. This is the vanilla form of knowledge distillation~\citep{DBLP:journals/corr/HintonVD15,DBLP:conf/nips/BaC14}. Soft targets generated by the teacher model provide much more information  than hard targets. Therefore, training the student model to fit soft targets can help the student model generalize better like the teacher model. 

\paratitle{Feature-based KD} exploits intermediate features to teach the student model, which is believed to be important for representation learning~\citep{DBLP:journals/pami/BengioCV13}. The simplest solution is to minimize the distance between intermediate representation of each student layer and its corresponding teacher layer~\citep{DBLP:journals/corr/RomeroBKCGB14,DBLP:conf/emnlp/SunCGL19,DBLP:conf/aaai/AguilarLZYFG20}. It enables the student model to exploit richer information from the teacher model. 
Recently, a number of feature-based KD studies have been proposed. In summary, these studies mainly focused on two key factors: selection of intermediate representations, and distance metric. 
The intuition of investigating better intermediate representations is that the knowledge of teacher should be easy to learn for the student model~\citep{DBLP:journals/corr/abs-1710-09505,DBLP:conf/iclr/ZagoruykoK17,DBLP:journals/corr/HuangW17a,DBLP:conf/cvpr/AhnHDLD19,DBLP:conf/aaai/HeoLY019a,DBLP:conf/emnlp/SunCGL19,DBLP:conf/aaai/AguilarLZYFG20}. For example, ~\citet{DBLP:journals/corr/HuangW17a} proposed to match the distributions of neuron selectivity patterns between the teacher and the student models. ~\citet{DBLP:conf/nips/KimPK18} trained a paraphrase model as $\textnormal{TF}_t$ to extract transferable features from the teacher's intermediate representations, and a translator model as $\textnormal{TF}_s$ to map the student intermediate representation to teacher's representations.
As for distance metrics, $L_2$ distance is the most widely used distance metric. Besides, $L_1$ distance~\citep{DBLP:conf/cvpr/WangYZF19}  and KL-divergence~\citep{DBLP:conf/iclr/LiuPS19,DBLP:conf/aaai/AguilarLZYFG20} are also used in previous KD approaches. 


\paratitle{Relation-based KD} aims to minimize the correlation between feature pairs from the teacher model and the student model~\citep{DBLP:conf/cvpr/YimJBK17,DBLP:conf/icml/SrinivasF18,DBLP:conf/eccv/LeeKS18,DBLP:conf/iccv/TungM19,DBLP:conf/bmvc/LeeS19,DBLP:conf/iccv/PengJLZWLZ019}. Distance can be seen a special kind of relation measure. Recently, many approaches have been proposed to explore better relation measures. 
For example, MHGD~\citep{DBLP:conf/bmvc/LeeS19} employed a multi-head attention network to encode relations between any two feature maps in a certain batch of training instances.  CCKD~\citep{DBLP:conf/iccv/PengJLZWLZ019} transferred correlation between instances with a generalized kernel method based on Taylor series expansion.


\paragraph{KD with Multiple Teachers}
Traditional KD methods focus on transferring the knowledge from one teacher model to the student model. A number of recent studies investigated knowledge transfer from multiple teachers or an ensemble of teachers. The most popular and straightforward way is to learn the ensemble of teacher logits~\citep{DBLP:conf/nips/TarvainenV17,DBLP:conf/kdd/YouX0T17}.  Following this idea, multiple studies have been proposed to model the diversity of teachers using  a weighted average of teacher logits~\citep{DBLP:journals/corr/RuderGB17,DBLP:conf/nips/LanZG18,DBLP:conf/eccv/XiangDH20}.   Apart from averaged logits, using the ensemble of features from multiple teachers is another line of research. ~\citep{DBLP:journals/corr/abs-1909-10754} proposed to train the student's feature map to minimize the gap from the feature maps of multiple teachers with different feature transformation functions. 
While achieving promising performance, traditional KD methods using multiple teachers suffer from expensive computations because they require multiple pre-trained teachers. To alleviate this problem,~\citet{DBLP:conf/cvpr/ZhangXHL18} proposed a deep mutual learning approach, which was an initial form of online KD that has been developed by various studies~\citep{DBLP:conf/nips/LanZG18,DBLP:conf/iclr/AnilPPODH18,DBLP:conf/aaai/ChenMWF020,DBLP:conf/icml/ChungPKK20,DBLP:conf/icpr/KimHCK20}. In online KD, a set of student models, or peers, was trained simultaneously by learning from each other in a peer-teaching fashion. 

\paragraph{KD with Dynamic Teacher}
In traditional KD, the teacher model is fixed during KD. However, this can be sub-optimal because the generalization ability of the student model is dynamic during training. A number of studies explored KD with an evolving teacher model to keep a reasonable capacity difference between student and teacher~\citep{DBLP:conf/aaai/MirzadehFLLMG20,shi2021learning,DBLP:journals/corr/abs-2106-04570,DBLP:journals/corr/abs-2102-07650}. For example, ~\citet{DBLP:conf/iccv/JinPWLLLYH19} designed a sequence of intermediate targets to impose curriculum-based constraint on the optimization path of the student model for improved KD. ~\citet{shi2021learning} and~\citet{DBLP:journals/corr/abs-2102-07650} proposed to jointly update the teacher model and the student model with task specific objectives during KD. 

\paragraph{Discussion} Knowledge distillation is a widely-used approach to get a smaller but more competitive model. However, although the idea of KD is easy to implement, it also has several limitations. First, the performance of KD is very sensitive to the size gap between the teacher model and the student model.  The discrepancy between the expressive power of the models would make it hard to teach the student model. Second, it relies on training data and may not be suitable for few-shot or zero-shot settings.  In addition, recent studies~\citep{DBLP:journals/corr/abs-2109-03228,DBLP:journals/corr/abs-2106-05945} have revealed that while knowledge distillation can effectively improve student generalization, there was still a large discrepancy between the predictive distributions of student and teacher models. It means that there is still a long way to distill full knowledge in a teacher model to a student model. 

\chapter{Efficient Data Usage}



Following the definition of Green deep learning, this chapter mainly explores how to achieve competitive results with fewer data resources, including active learning and pre-training. It is worth noticing that although pre-trained models take massive computations during pre-training, they are widely believed to be a practical solution to release the burden of data in downstream tasks. Therefore, we also include pre-trained models in this chapter. 

\begin{figure*}[ht]
  \centering
\begin{forest}
  forked edges,
  for tree={
    grow=east, 
    reversed=true, 
    anchor=base west, 
    parent anchor=east, 
    child anchor=west, 
    base=left, 
    font=\footnotesize,
    rectangle,
    draw=hiddendraw,
    rounded corners,align=left,
    minimum width=2.5em,
    minimum height=1.2em,
    s sep=6pt,
    inner xsep=3pt,
    inner ysep=1pt,
  },
  where level=1{font=\scriptsize}{},
  where level=2{font=\scriptsize}{},
  where level=3{font=\scriptsize}{},
  where level=4{font=\scriptsize}{},
  where level=5{font=\scriptsize}{},
  [Green Data Usage
    [Active Learning
    [Uncertainty-based]
    [Diversity-based]
    [Expected Model Change]
    ]
    [Pre-training as Few-shot Learners 
      [Self-supervised Pre-training]
      [Contrastive Pre-training]
       [Prompt Pre-training]
     ]
   ]
\end{forest}
\caption{Taxonomy of efficient data usage methods with representative examples.}
\label{taxonomy_of_datausage}
\end{figure*}
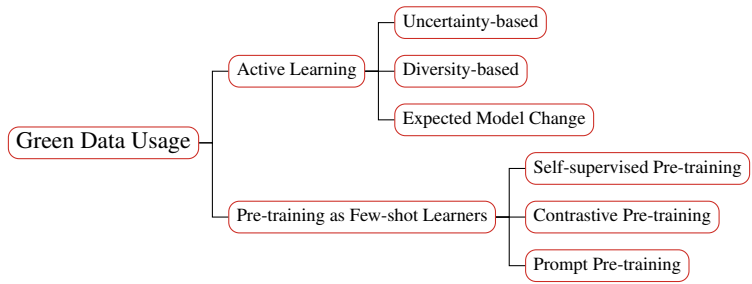

\section{Active Learning}
Active learning is a research direction aiming at using as few samples as possibles to achieve good results. It is initially proposed to reduce annotation costs. Nowadays, pool-based active learning is widely-used to reduce training costs by selecting the most useful examples to train networks. The intuitive behind active learning is quite simple.  The annotated training data do not equally contribute to the final performance. If we can always select the most useful example to train models, the wasted training on unimportant examples can be largely reduced.

Active learning usually starts from a randomly-initialized model or a pre-trained model. It defines several query strategies to select new unlabeled data to query annotation. The new data associated with labels are then used to train the model. This process keeps running until it reaches the maximum number of labelled data or other termination conditions are satisfied. Previous active learning approaches mainly focused on query strategies to improve performance.  Following the work of~\citet{DBLP:conf/cvpr/YooK19}, given a pool of unlabeled data, we classify active learning approaches into three categories according to the selection criteria: uncertainty-based approaches, diversity-based approaches, and expected-model-change based approaches. 

\textit{Uncertainty-based active learning} is used to be the
most common choice in active learning, using uncertainty scores to select data~\citep{DBLP:conf/icip/RanganathanVCP17,DBLP:journals/corr/abs-2107-05687,DBLP:conf/icmcs/HeJDYY19,DBLP:conf/iclr/ShenYLKA18}.
The simplest solution is to utilize class posterior probabilities to define uncertainty. To be specific, ~\citet{DBLP:conf/sigir/LewisG94} used the probability of a predicted class as uncertainty score. ~\citet{DBLP:conf/cvpr/JoshiPP09} defined an entropy of
class posterior probabilities as uncertainty score. Another research line is to train multiple models to construct a committee, and used the committee to evaluate uncertainty~\citep{DBLP:conf/cvpr/BeluchGNK18}.  Recently, uncertainty-based active learning has been widely applied to various fields.
~\citet{DBLP:conf/icip/RanganathanVCP17} applied active learning on image classification which selected the most informative unlabeled samples to train a deep belief network model. ~\citet{DBLP:conf/iclr/ShenYLKA18} applied uncertainty-based active learning to sequence tagging. They selected sentences for which the length-normalized log probability of the current prediction
was the lowest.  ~\citet{DBLP:journals/corr/abs-1802-09841} focused on margin-based active learning for deep networks. 
Despite promising effectiveness, uncertainty-based sampling can easily lead to insufficient diversity of batch query samples. 

\textit{Diversity-based active learning} has been proposed to address the challenges of uncertainty based approaches. For example, ~\citet{DBLP:conf/iclr/SenerS18} defined a core-set to estimate the whole training data. ~\citet{DBLP:conf/icml/NguyenS04} proposed to incorporate clustering into active learning. It first constructed a classifier on the set of the cluster representatives, and then propagated the classification decision to the other samples via a local noise model.

\textit{Expected-model-change based active learning} selected data points that
would cause the greatest change to current model. For example,  ~\citet{roy2001toward} selected examples according to the reduced error rate on future test examples.  ~\citet{DBLP:conf/eccv/FreytagRD14} measured the expected change of model outputs and incorporated the underlying data distribution.  For each example of an unlabeled set, the expected change of model predictions was calculated and marginalized over the unknown label.




\section{Pre-training as Few-shot Learners}
It is widely-believed that pre-trained models as initialization is an effective approach to reduce data requirements in downstream tasks. In this survey, we describe widely-used pre-trained models. 

\subsection{Self-supervised Learning} 
self-supervised learning (SSL) is the most popular solution to get pre-trained models. We take NLP as an example to review recent progress of self-supervised learning. 
Self-supervised objectives can be classified into types including masked language modeling (MLM), language modeling (LM), de-nosing auto-encoding (DAE). In this survey, we give an overview of these models.  We refer the reader to \cite{DBLP:journals/corr/abs-2003-08271} and \cite{DBLP:journals/corr/abs-2106-07139} for more details of pre-trained networks. 



\paragraph{Masked Language Modeling} \citet{DBLP:conf/naacl/DevlinCLT19} proposed to pre-train an Transformer encoder, i.e., BERT, via MLM objective on unlabeled text. MLM builds a corrupted token sequence where 15\% tokens are randomly sampled and replaced with a special token \texttt{[MASK]}, then requires the Transformer to predict the original tokens. 
Formally, given a sequence $\bm{x}_{1:L}=[x_1,x_2,\cdots,x_L]$ and the masked token set $\mathcal{M}$, the MLM objective can be formulated as
\begin{equation}
    \mathcal{L}_\text{MLM} = -\sum_{x_t\in\mathcal{M}} \log p(x_t|\bm{x}_{\setminus\mathcal{M}})
\end{equation}
where $\bm{x}_{\setminus\mathcal{M}}$ indicates the unmasked part of the input sequence. 
With the inherited knowledge, fine-tuned BERT performs well compared to baselines without pre-training on many classification tasks, including single sentence classification~\citep{DBLP:journals/tacl/WarstadtSB19, DBLP:conf/emnlp/SocherPWCMNP13}, sentence pair classification or matching~\citep{DBLP:conf/acl-iwp/DolanB05, DBLP:conf/semeval/CerDALS17}, natural language inference~\citep{DBLP:conf/naacl/WilliamsNB18, DBLP:conf/iclr/WangSMHLB19, DBLP:conf/tac/BentivogliMDDG09, DBLP:conf/kr/LevesqueDM12}, and question answering~\citep{DBLP:conf/emnlp/RajpurkarZLL16, DBLP:conf/acl/RajpurkarJL18}, etc. 

Several studies have made effort to improve MLM through developing more effective MLM-like objectives or exploring more efficient training tricks. SpanBERT~\citep{DBLP:journals/tacl/JoshiCLWZL20}, ERNIE~\citep{DBLP:journals/corr/abs-1904-09223}, and BERT-WWM~\citep{DBLP:journals/corr/abs-1906-08101} proposed to mask adjacent tokens.
 ELECTRA~\citep{DBLP:conf/iclr/ClarkLLM20} proposed a GAN-like replaced token prediction objective which required discriminator to discriminate whether a token is replaced or not. Instead of masking, StructBERT~\citep{DBLP:conf/iclr/0225BYWXBPS20} learned to predict the shuffled span in its original order, which incorporated language structures into pre-training.

\paragraph{Language Modeling} predicts next tokens one by one in an auto-regressive way. Specifically, given a sequence $\bm{x}_{1:L}=[x_1,x_2,\cdots,x_L]$, the joint probability of $\bm{x}$ can be written as 
\begin{equation}
    p(\bm{x}_{1:L})=\prod_{t=1}^Lp(x_t|\bm{x}_{0:t-1})
\end{equation}
\begin{equation}
    \mathcal{L}_\text{LM} = -\sum_t \log p(x_t|\bm{x}_{<t})
\end{equation}
while $x_0$ is the special \texttt{[BOS]} token which indicates the beginning of a sentence. LM is usually implemented with a Transformer decoder. GPT-2~\citep{radford2019language}, and GPT-3~\citep{DBLP:conf/nips/BrownMRSKDNSSAA20} are two representative LM-based pre-trained models. The LM objective directly models the probability of next token given its left context. LM-based pre-trained models as initialization can largely improve conditional natural language generation tasks like  summarization~\citep{DBLP:conf/emnlp/RushCW15, DBLP:conf/acl/SeeLM17} and question answering~\citep{DBLP:conf/acl/RajpurkarJL18, DBLP:journals/tacl/ReddyCM19}. Therefore, LM models usually are used to initialize generative models. 

There are lots of follow-up studies improving the original LM objective from various aspects. For example, inspired by the fact that neural networks could have different reading orders with human, XLNet~\citep{DBLP:conf/nips/YangDYCSL19} proposed to use permuted language modeling to predict a sentence in permuted orders in an auto-regressive way.
UniLM~\citep{DBLP:conf/nips/00040WWLWGZH19} proposed a prefix LM where next tokens are predicted in an auto-regressive way with all context tokens visible to each other.
ProphetNet~\citep{DBLP:conf/emnlp/QiYGLDCZ020} introduced a future $n$-gram
prediction objective to predict the next $n$ tokens simultaneously based on previous context at each time step, through a multi-stream attention similar to XLNet.

\paragraph{De-nosing Auto-encoding} requires a model to recover the original sentence based on the corrupted version. Formally, DAE-like objectives can be formulated as
\begin{equation}
    \mathcal{L}_\text{DAE} = -\sum_t \log p(x_t|\bm{\hat x}, \bm{x}_{<t})
\end{equation}
where $\bm{\hat x}$ is  the corrupted input. By learning to distinguish which part of content is corrupted in a text and recovering it in a natural order, DAE-based pre-trained models as initialization can improve various models, including language understanding and language generation. 

Actually, there are also flexible variants to corrupt a sequence, such as shuffling, masking, dropping, rotating, etc.  BART~\citep{DBLP:conf/acl/LewisLGGMLSZ20} corrupted its input sequence with arbitrary noise transformations. The transformations include token masking, token deletion, sentence permutation, document rotation, and text infilling. 
More recently,~\citet{DBLP:journals/corr/abs-2101-00416} proposed to use text rewriting instead of text infilling (e.g., BART, T5) for improving seq2seq pre-trained transformers.
 
\paragraph{Multilingual Objectives} Another line of SSL objectives is multilingual SSL objectives. For example, mBERT~\citep{DBLP:conf/naacl/DevlinCLT19}~\footnote{\url{https://github.com/google-research/bert/blob/master/multilingual.md}}, mBART~\citep{DBLP:journals/tacl/LiuGGLEGLZ20}, mT5~\citep{DBLP:conf/naacl/XueCRKASBR21} pre-trained the multilingual version BERT, BART, T5 via the multilingual masked language modeling objective. These models highly improve the results of few-shot or low-shot multilingual learning or multilingual generation tasks.  
In addition to these simple variants, recent researchers also designed more sophisticated cross-lingual objectives, such as cross-lingual word recovery, cross-lingual paraphrase classification, and cross-lingual masked language modeling. 

\subsection{Contrastive Learning}
Contrastive learning focuses on pair-wise relationships, aiming at learning closer or similar representation for positive samples while pushing negative samples away. The recent years have witnessed the rapid progress of contrastive-based pre-trained models, especially in CV~\citep{DBLP:journals/corr/abs-1807-03748, DBLP:conf/cvpr/He0WXG20, DBLP:conf/icml/ChenK0H20,  DBLP:journals/corr/abs-2104-02057}. 

The standard contrastive learning utilizes positive and negative pairs at the same time to construct its objective. This kind of loss can back to noise contrastive estimation (NCE) loss~\citep{DBLP:journals/jmlr/GutmannH10} that is defined as
\begin{equation}
    \mathcal{L}_\text{NCE} =
    -\log\frac{\exp (f(q, k^+)/\tau)}
    {\exp (f(q, k^+)/\tau) + \exp (f(q, k^-)/\tau)}
\end{equation}
where $q$ is the anchor sample; $k^+$ and $k^-$ are its positive sample and negative sample; $q$, $k^+$ and $k^-$ are vectors generated by a neural network;  $f(\cdot,\cdot)$ is a similarity function; $\tau$ is the temperature  controlling the concentration of the induced distribution.  
When more than one negative samples exist, the NCE loss becomes the InfoNCE loss:
\begin{equation}
\label{infonce_bound}
    \mathcal{L}_\text{InfoNCE} =
    -\log\frac{\exp (f(q, k^+)/\tau)}
    {\exp (f(q, k^+)/\tau) + 
    \sum_{i=1}^K\exp (f(q, k_i)/\tau)}
\end{equation}
where $k_i$ represents the $i$-th negative sample of anchor/query sample $q$;  $K$ is the size of negative samples. \cite{DBLP:journals/corr/abs-1807-03748} proved that minimizing the InfoNCE loss was equivalent to maximizing the lower bound of mutual information between $q$ and $k^+$:
\begin{equation}
    I(q, k^+)\ge \log(K) - \mathcal{L}_\text{InfoNCE}
\end{equation}
where the more negative samples are, the tighter the lower bound is.

Overall, contrastive losses are easy to implement and contrastive-based pre-trained models generally have strong transfer ability. In the CV field, it largely improve downstream tasks, such as ImageNet classification~\citep{DBLP:conf/cvpr/DengDSLL009}, object detection~\citep{DBLP:journals/ijcv/EveringhamGWWZ10}, and egmentation~\citep{DBLP:conf/eccv/LinMBHPRDZ14}, etc.  MoCo~\citep{DBLP:conf/cvpr/He0WXG20,DBLP:journals/corr/abs-2003-04297,DBLP:journals/corr/abs-2104-02057} is one of representative models which applied advanced contrastive learning methods to train pre-trained models on CV fields. 

Existing state-of-the-art contrastive-based pre-trained models are simple variants of Eq.~\ref{infonce_bound} where positive pairs and negative paris are required. 
 For example, Deep InfoMax applied contrastive losses to learn image representations via maximizing the mutual information between local patches and the whole image;  CMC~\citep{DBLP:conf/eccv/TianKI20} applied contrastive losses to learn representations where different views of the same scene or instance were sampled as positive pairs. 
According to \cite{DBLP:conf/icml/0001I20}, the InfoNCE loss optimized two properties of a representation space, including the alignment of representations between positive samples and the uniformity of the induced distribution of normalized representations on a hypersphere in the representation space.  Nevertheless, the recent studies such as BYOL~\citep{DBLP:conf/nips/GrillSATRBDPGAP20}, SwAV~\citep{DBLP:conf/nips/CaronMMGBJ20}, Simsiam~\citep{DBLP:conf/cvpr/ChenH21} found that contrastive learning also works without negative samples. 
In all, the research of contrastive-based pre-training is still in rapid development. More understanding studies are expected to better explain how contrastive losses work. 

Previous studies also applied contrastive learning to NLP tasks. 
For example, CAPT~\citep{DBLP:journals/corr/abs-2010-06351} taken a sentence $s_i$ and its masked version $\tilde s_i$ as a positive pair. 
CERT~\citep{DBLP:journals/corr/abs-2005-12766} utilized back-translation to generate noised positive samples for the source English sentence; CLEAR~\citep{DBLP:journals/corr/abs-2012-15466} directly integrated four text augmentation mechanisms including word deletion, span deletion, reordering, and substitution, and taken two augmented versions of an sentence as positive pairs; DeCLUTR~\citep{DBLP:conf/acl/GiorgiNWB20} samples nearby even overlapping spans as positive pairs;
SimCSE~\citep{DBLP:journals/corr/abs-2104-08821} added noise to encoded representations via dropout and regard noised representations as positive samples. SimCSE performed pertty well on retrieval tasks and achieved new SoTA on semantic textual similarity tasks.

\subsection{Prompt Learning}\label{prompt}
As the size of pre-trained models grows rapidly, fine-tuning the super large pre-trained model usually requires massive data to get better generalization ability, thus failing on few-shot/zero-shot applications. The objective gap between pre-training and fine-tuning stages are one of important reasons behind failure. Therefore, to improve data efficiency, prompt learning is proposed by extracting the similar/same template for pre-training and fine-tuning stages. ~\citet{DBLP:conf/naacl/ScaoR21} demonstrated that a prompt may be equal to 100 conventional data points, indicating that prompts can greatly improve the sample efficiency. 


In the recent survey about prompt learning~\citep{DBLP:journals/corr/abs-2107-13586}, \textit{prompt} is defined as $\bm{x}'$ = $f_\text{prompt}(\bm{x})$, where $f_\text{prompt}(\cdot)$ is the prompting function to rewrite the input $\bm{x}$ with an task-specific template. Suppose that we have a review ``I dozed off when watching this movie.'' as input $\bm{x}$ associated with sentiment label $\bm{y}$. A prompt example $\bm{x}'$ is ``I dozed off when watching this movie. It is so \texttt{[MASK]}''. In this way, we transfer sentiment prediction to a masked language modeling task. 
As pointed by \citet{DBLP:journals/corr/abs-2109-01652}, language models at scale like GPT-3~\citep{DBLP:conf/nips/BrownMRSKDNSSAA20} contain substantial world knowledge and can perform a range of NLP tasks, which makes large pre-trained models a kind of ``neural knowledge bases''~\citep{DBLP:conf/emnlp/PetroniRRLBWM19}. 
In fact, prompts can be seen as queries for those neural knowledge bases. The difficulty is (1) how to generate appropriate queries to achieve your goals, and (2) how to understand or interpret the response in the format of predicted textual content. 

Generally, prompts are defined as discrete templates. In addition, there are also studies focusing on continuous prompts, rather than textual prompts.    For example, Prefix Tuning~\citep{DBLP:conf/acl/LiL20} and P-tuning~\citep{DBLP:journals/corr/abs-2103-10385} defined virtual tokens where only parameters of virtual tokens were fine-tuned. 
The more recent P-tuning V2~\citep{liu2021ptuning} used multi-task learning objectives and obtained competitive even better results than vanilla fine-tuning. SPoT~\citep{vu2021spot} further utilized transfer learning to support unseen tasks, where the prefix was pre-tuning on related tasks then used for initialization in unseen task. Overall, these methods are ideologically similar to adaptation approaches~\citep{DBLP:conf/icml/HoulsbyGJMLGAG19,DBLP:conf/emnlp/BapnaF19,DBLP:conf/emnlp/PfeifferRPKVRCG20} because both of them do not change the most of parameters of the pre-trained model.  However, the adaptation methods usually insert adaptors between layers, undermining the original model architecture.


\clearpage

\label{cha:data}


\chapter{Conclusions and Future Directions}
\label{conclusion}

We believe that Green deep learning will be an important research direction in the future. With recent advances in deep learning, the community have made significant progress in developing super-large models for downstream tasks,  making it possible to apply AI models on complicated applications. Considering that it is the ultimate goal to deploy AI models on real-world devices with high requirements on resource usage, how to transfer strong models to resource-constraint devices (e.g., mobile) becomes a priority goal.  In this section, we list several challenges for Green deep learning. 


\textbf{Green deep learning theory}. We have harnessed some advanced Green deep learning techniques, but many questions still remain to be explored. For example, 1) If we already have a well-performing model, why was additional training required to transfer knowledge to a small model?   2) How many parameters do we need at least for feasible training and inference? 3) How to design Green learning algorithms to enable efficient zero-shot learning or few-shot learning like human? 4) How our model store knowledge and how to make models to achieve lifelong learning without forgetting learned knowledge during training? 5) Is linear algebra is the only basic theory for deep learning and whether can we develop a new operation system beyond linear algebra? In this survey, we just show limited questions due to page limitation. The community still have a long way to go in the theory of Green deep learning 

\textbf{Green deep learning under extreme computation constraints. }   Deploying models on edge devices enjoys multiple advantages. 
First, it can avoid uploading user information to the cloud amid the tide of protecting user privacy. 
Second, it can empower more applications with high requirements on latency if deployed a light-weight model on edge devices.
While recent advances in machine learning greatly facilitate efficient training and inference in the cloud, edge devices bring new challenges caused by extremely strict memory and computation constraints. Therefore, how to design advanced training and inference algorithms towards tiny devices is also an important and challenging direction. First, algorithm-hardware cooperation is a promising direction to satisfy speed requirements when deploying large models. For example, Lightseq~\citep{wang2021lightseq} and Faster Transformer
are two representative models that use CUDA implementation to accelerate Transformer execution.  Second, edge-cloud cooperation is  also a practical solution by combing powerful and elastic cloud computing and immediate edge computing. An intuitive solution is to design a dynamic network where edge devices can handle simple samples with smaller models and the cloud is responsible for handing complicated samples with larger models. 

\bibliography{ref.bib}

\begin{thebibliography}{402}
\providecommand{\natexlab}[1]{#1}
\providecommand{\url}[1]{\texttt{#1}}
\expandafter\ifx\csname urlstyle\endcsname\relax
  \providecommand{\doi}[1]{doi: #1}\else
  \providecommand{\doi}{doi: \begingroup \urlstyle{rm}\Url}\fi

\bibitem[Abdelfattah et~al.(2021)Abdelfattah, Mehrotra, Dudziak, and
  Lane]{DBLP:journals/corr/abs-2101-08134}
Mohamed~S. Abdelfattah, Abhinav Mehrotra, Lukasz Dudziak, and Nicholas~D. Lane.
\newblock Zero-cost proxies for lightweight {NAS}.
\newblock \emph{ICLR}, 2021.

\bibitem[Aguilar et~al.(2020)Aguilar, Ling, Zhang, Yao, Fan, and
  Guo]{DBLP:conf/aaai/AguilarLZYFG20}
Gustavo Aguilar, Yuan Ling, Yu~Zhang, Benjamin Yao, Xing Fan, and Chenlei Guo.
\newblock Knowledge distillation from internal representations.
\newblock In \emph{The Thirty-Fourth {AAAI} Conference on Artificial
  Intelligence, {AAAI} 2020, The Thirty-Second Innovative Applications of
  Artificial Intelligence Conference, {IAAI} 2020, The Tenth {AAAI} Symposium
  on Educational Advances in Artificial Intelligence, {EAAI} 2020, New York,
  NY, USA, February 7-12, 2020}, pp.\  7350--7357. {AAAI} Press, 2020.

\bibitem[Aharoni et~al.(2019)Aharoni, Johnson, and
  Firat]{DBLP:conf/naacl/AharoniJF19}
Roee Aharoni, Melvin Johnson, and Orhan Firat.
\newblock Massively multilingual neural machine translation.
\newblock In \emph{{NAACL-HLT} {(1)}}, pp.\  3874--3884. Association for
  Computational Linguistics, 2019.

\bibitem[Ahn et~al.(2019)Ahn, Hu, Damianou, Lawrence, and
  Dai]{DBLP:conf/cvpr/AhnHDLD19}
Sungsoo Ahn, Shell~Xu Hu, Andreas~C. Damianou, Neil~D. Lawrence, and Zhenwen
  Dai.
\newblock Variational information distillation for knowledge transfer.
\newblock In \emph{{IEEE} Conference on Computer Vision and Pattern
  Recognition, {CVPR} 2019, Long Beach, CA, USA, June 16-20, 2019}, pp.\
  9163--9171. Computer Vision Foundation / {IEEE}, 2019.

\bibitem[Alizadeh et~al.(2019)Alizadeh, Fern{\'{a}}ndez{-}Marqu{\'{e}}s, Lane,
  and Gal]{DBLP:conf/iclr/AlizadehFLG19}
Milad Alizadeh, Javier Fern{\'{a}}ndez{-}Marqu{\'{e}}s, Nicholas~D. Lane, and
  Yarin Gal.
\newblock An empirical study of binary neural networks' optimisation.
\newblock In \emph{7th International Conference on Learning Representations,
  {ICLR} 2019, New Orleans, LA, USA, May 6-9, 2019}. OpenReview.net, 2019.

\bibitem[Andreas \& Klein(2015)Andreas and Klein]{DBLP:conf/naacl/AndreasK15}
Jacob Andreas and Dan Klein.
\newblock When and why are log-linear models self-normalizing?
\newblock In \emph{{HLT-NAACL}}, pp.\  244--249. The Association for
  Computational Linguistics, 2015.

\bibitem[Anil et~al.(2018)Anil, Pereyra, Passos, Orm{\'{a}}ndi, Dahl, and
  Hinton]{DBLP:conf/iclr/AnilPPODH18}
Rohan Anil, Gabriel Pereyra, Alexandre Passos, R{\'{o}}bert Orm{\'{a}}ndi,
  George~E. Dahl, and Geoffrey~E. Hinton.
\newblock Large scale distributed neural network training through online
  distillation.
\newblock In \emph{6th International Conference on Learning Representations,
  {ICLR} 2018, Vancouver, BC, Canada, April 30 - May 3, 2018, Conference Track
  Proceedings}. OpenReview.net, 2018.

\bibitem[Ba \& Caruana(2014)Ba and Caruana]{DBLP:conf/nips/BaC14}
Jimmy Ba and Rich Caruana.
\newblock Do deep nets really need to be deep?
\newblock In Zoubin Ghahramani, Max Welling, Corinna Cortes, Neil~D. Lawrence,
  and Kilian~Q. Weinberger (eds.), \emph{Advances in Neural Information
  Processing Systems 27: Annual Conference on Neural Information Processing
  Systems 2014, December 8-13 2014, Montreal, Quebec, Canada}, pp.\
  2654--2662, 2014.

\bibitem[Ba et~al.(2016)Ba, Kiros, and Hinton]{DBLP:journals/corr/BaKH16}
Lei~Jimmy Ba, Jamie~Ryan Kiros, and Geoffrey~E. Hinton.
\newblock Layer normalization.
\newblock \emph{CoRR}, abs/1607.06450, 2016.

\bibitem[Bahdanau et~al.(2015)Bahdanau, Cho, and
  Bengio]{DBLP:journals/corr/BahdanauCB14}
Dzmitry Bahdanau, Kyunghyun Cho, and Yoshua Bengio.
\newblock Neural machine translation by jointly learning to align and
  translate.
\newblock In \emph{{ICLR}}, 2015.

\bibitem[Bahri et~al.(2020)Bahri, Bahl, and
  Zafeiriou]{DBLP:journals/corr/abs-2012-15823}
Mehdi Bahri, Ga{\'{e}}tan Bahl, and Stefanos Zafeiriou.
\newblock Binary graph neural networks.
\newblock \emph{CoRR}, abs/2012.15823, 2020.

\bibitem[Bai et~al.(2020)Bai, Zhang, Hou, Shang, Jin, Jiang, Liu, Lyu, and
  King]{DBLP:journals/corr/abs-2012-15701}
Haoli Bai, Wei Zhang, Lu~Hou, Lifeng Shang, Jing Jin, Xin Jiang, Qun Liu,
  Michael~R. Lyu, and Irwin King.
\newblock Binarybert: Pushing the limit of {BERT} quantization.
\newblock \emph{CoRR}, abs/2012.15701, 2020.

\bibitem[Bai et~al.(2019)Bai, Kolter, and Koltun]{DBLP:conf/nips/BaiKK19}
Shaojie Bai, J.~Zico Kolter, and Vladlen Koltun.
\newblock Deep equilibrium models.
\newblock In \emph{NeurIPS}, pp.\  688--699, 2019.

\bibitem[Balzer et~al.(1991)Balzer, Takahashi, Ohta, and
  Kyuma]{DBLP:journals/nn/BalzerTOK91}
Wolfgang Balzer, Masanobu Takahashi, Jun Ohta, and Kazuo Kyuma.
\newblock Weight quantization in boltzmann machines.
\newblock \emph{Neural Networks}, 4\penalty0 (3):\penalty0 405--409, 1991.

\bibitem[Bapna \& Firat(2019)Bapna and Firat]{DBLP:conf/emnlp/BapnaF19}
Ankur Bapna and Orhan Firat.
\newblock Simple, scalable adaptation for neural machine translation.
\newblock In \emph{{EMNLP/IJCNLP} {(1)}}, pp.\  1538--1548. Association for
  Computational Linguistics, 2019.

\bibitem[Belilovsky et~al.(2019)Belilovsky, Eickenberg, and
  Oyallon]{DBLP:conf/icml/BelilovskyEO19}
Eugene Belilovsky, Michael Eickenberg, and Edouard Oyallon.
\newblock Greedy layerwise learning can scale to imagenet.
\newblock In Kamalika Chaudhuri and Ruslan Salakhutdinov (eds.),
  \emph{Proceedings of the 36th International Conference on Machine Learning,
  {ICML} 2019, 9-15 June 2019, Long Beach, California, {USA}}, volume~97 of
  \emph{Proceedings of Machine Learning Research}, pp.\  583--593. {PMLR},
  2019.

\bibitem[Beluch et~al.(2018)Beluch, Genewein, N{\"{u}}rnberger, and
  K{\"{o}}hler]{DBLP:conf/cvpr/BeluchGNK18}
William~H. Beluch, Tim Genewein, Andreas N{\"{u}}rnberger, and Jan~M.
  K{\"{o}}hler.
\newblock The power of ensembles for active learning in image classification.
\newblock In \emph{{CVPR}}, pp.\  9368--9377. Computer Vision Foundation /
  {IEEE} Computer Society, 2018.

\bibitem[Bengio \& Senecal(2003)Bengio and
  Senecal]{DBLP:conf/aistats/BengioS03}
Yoshua Bengio and Jean{-}S{\'{e}}bastien Senecal.
\newblock Quick training of probabilistic neural nets by importance sampling.
\newblock In \emph{{AISTATS}}. Society for Artificial Intelligence and
  Statistics, 2003.

\bibitem[Bengio \& Senecal(2008)Bengio and
  Senecal]{DBLP:journals/tnn/BengioS08}
Yoshua Bengio and Jean{-}S{\'{e}}bastien Senecal.
\newblock Adaptive importance sampling to accelerate training of a neural
  probabilistic language model.
\newblock \emph{{IEEE} Trans. Neural Networks}, 19\penalty0 (4):\penalty0
  713--722, 2008.

\bibitem[Bengio et~al.(2006)Bengio, Lamblin, Popovici, and
  Larochelle]{DBLP:conf/nips/BengioLPL06}
Yoshua Bengio, Pascal Lamblin, Dan Popovici, and Hugo Larochelle.
\newblock Greedy layer-wise training of deep networks.
\newblock In Bernhard Sch{\"{o}}lkopf, John~C. Platt, and Thomas Hofmann
  (eds.), \emph{Advances in Neural Information Processing Systems 19,
  Proceedings of the Twentieth Annual Conference on Neural Information
  Processing Systems, Vancouver, British Columbia, Canada, December 4-7, 2006},
  pp.\  153--160. {MIT} Press, 2006.

\bibitem[Bengio et~al.(2013{\natexlab{a}})Bengio, Courville, and
  Vincent]{DBLP:journals/pami/BengioCV13}
Yoshua Bengio, Aaron~C. Courville, and Pascal Vincent.
\newblock Representation learning: {A} review and new perspectives.
\newblock \emph{{IEEE} Trans. Pattern Anal. Mach. Intell.}, 35\penalty0
  (8):\penalty0 1798--1828, 2013{\natexlab{a}}.

\bibitem[Bengio et~al.(2013{\natexlab{b}})Bengio, L{\'{e}}onard, and
  Courville]{DBLP:journals/corr/BengioLC13}
Yoshua Bengio, Nicholas L{\'{e}}onard, and Aaron~C. Courville.
\newblock Estimating or propagating gradients through stochastic neurons for
  conditional computation.
\newblock \emph{CoRR}, abs/1308.3432, 2013{\natexlab{b}}.

\bibitem[Bentivogli et~al.(2009)Bentivogli, Magnini, Dagan, Dang, and
  Giampiccolo]{DBLP:conf/tac/BentivogliMDDG09}
Luisa Bentivogli, Bernardo Magnini, Ido Dagan, Hoa~Trang Dang, and Danilo
  Giampiccolo.
\newblock The fifth {PASCAL} recognizing textual entailment challenge.
\newblock In \emph{Proceedings of the Second Text Analysis Conference, {TAC}
  2009, Gaithersburg, Maryland, USA, November 16-17, 2009}. {NIST}, 2009.

\bibitem[Bhandare et~al.(2019)Bhandare, Sripathi, Karkada, Menon, Choi, Datta,
  and Saletore]{DBLP:journals/corr/abs-1906-00532}
Aishwarya Bhandare, Vamsi Sripathi, Deepthi Karkada, Vivek Menon, Sun Choi,
  Kushal Datta, and Vikram Saletore.
\newblock Efficient 8-bit quantization of transformer neural machine language
  translation model.
\newblock \emph{CoRR}, abs/1906.00532, 2019.

\bibitem[Bjorck et~al.(2018)Bjorck, Gomes, Selman, and
  Weinberger]{DBLP:conf/nips/BjorckGSW18}
Johan Bjorck, Carla~P. Gomes, Bart Selman, and Kilian~Q. Weinberger.
\newblock Understanding batch normalization.
\newblock In \emph{NeurIPS}, pp.\  7705--7716, 2018.

\bibitem[Blackwood et~al.(2018)Blackwood, Ballesteros, and
  Ward]{DBLP:conf/coling/BlackwoodBW18}
Graeme~W. Blackwood, Miguel Ballesteros, and Todd Ward.
\newblock Multilingual neural machine translation with task-specific attention.
\newblock In \emph{{COLING}}, pp.\  3112--3122. Association for Computational
  Linguistics, 2018.

\bibitem[Bolukbasi et~al.(2017)Bolukbasi, Wang, Dekel, and
  Saligrama]{DBLP:conf/icml/BolukbasiWDS17}
Tolga Bolukbasi, Joseph Wang, Ofer Dekel, and Venkatesh Saligrama.
\newblock Adaptive neural networks for efficient inference.
\newblock In Doina Precup and Yee~Whye Teh (eds.), \emph{Proceedings of the
  34th International Conference on Machine Learning, {ICML} 2017, Sydney, NSW,
  Australia, 6-11 August 2017}, volume~70 of \emph{Proceedings of Machine
  Learning Research}, pp.\  527--536. {PMLR}, 2017.

\bibitem[Brown et~al.(2020)Brown, Mann, Ryder, Subbiah, Kaplan, Dhariwal,
  Neelakantan, Shyam, Sastry, Askell, Agarwal, Herbert{-}Voss, Krueger,
  Henighan, Child, Ramesh, Ziegler, Wu, Winter, Hesse, Chen, Sigler, Litwin,
  Gray, Chess, Clark, Berner, McCandlish, Radford, Sutskever, and
  Amodei]{DBLP:conf/nips/BrownMRSKDNSSAA20}
Tom~B. Brown, Benjamin Mann, Nick Ryder, Melanie Subbiah, Jared Kaplan,
  Prafulla Dhariwal, Arvind Neelakantan, Pranav Shyam, Girish Sastry, Amanda
  Askell, Sandhini Agarwal, Ariel Herbert{-}Voss, Gretchen Krueger, Tom
  Henighan, Rewon Child, Aditya Ramesh, Daniel~M. Ziegler, Jeffrey Wu, Clemens
  Winter, Christopher Hesse, Mark Chen, Eric Sigler, Mateusz Litwin, Scott
  Gray, Benjamin Chess, Jack Clark, Christopher Berner, Sam McCandlish, Alec
  Radford, Ilya Sutskever, and Dario Amodei.
\newblock Language models are few-shot learners.
\newblock In \emph{Advances in Neural Information Processing Systems 33: Annual
  Conference on Neural Information Processing Systems 2020, NeurIPS 2020,
  December 6-12, 2020, virtual}, 2020.

\bibitem[Bucila et~al.(2006)Bucila, Caruana, and
  Niculescu{-}Mizil]{DBLP:conf/kdd/BucilaCN06}
Cristian Bucila, Rich Caruana, and Alexandru Niculescu{-}Mizil.
\newblock Model compression.
\newblock In Tina Eliassi{-}Rad, Lyle~H. Ungar, Mark Craven, and Dimitrios
  Gunopulos (eds.), \emph{Proceedings of the Twelfth {ACM} {SIGKDD}
  International Conference on Knowledge Discovery and Data Mining,
  Philadelphia, PA, USA, August 20-23, 2006}, pp.\  535--541. {ACM}, 2006.

\bibitem[Cai et~al.(2019)Cai, Zhu, and Han]{DBLP:conf/iclr/CaiZH19}
Han Cai, Ligeng Zhu, and Song Han.
\newblock Proxylessnas: Direct neural architecture search on target task and
  hardware.
\newblock In \emph{7th International Conference on Learning Representations,
  {ICLR} 2019, New Orleans, LA, USA, May 6-9, 2019}. OpenReview.net, 2019.

\bibitem[Cai et~al.(2020)Cai, Gan, Wang, Zhang, and
  Han]{DBLP:conf/iclr/CaiGWZH20}
Han Cai, Chuang Gan, Tianzhe Wang, Zhekai Zhang, and Song Han.
\newblock Once-for-all: Train one network and specialize it for efficient
  deployment.
\newblock In \emph{8th International Conference on Learning Representations,
  {ICLR} 2020, Addis Ababa, Ethiopia, April 26-30, 2020}. OpenReview.net, 2020.

\bibitem[Campos et~al.(2018)Campos, Jou, Gir{\'{o}}{-}i{-}Nieto, Torres, and
  Chang]{DBLP:conf/iclr/CamposJNTC18}
V{\'{\i}}ctor Campos, Brendan Jou, Xavier Gir{\'{o}}{-}i{-}Nieto, Jordi Torres,
  and Shih{-}Fu Chang.
\newblock Skip {RNN:} learning to skip state updates in recurrent neural
  networks.
\newblock In \emph{6th International Conference on Learning Representations,
  {ICLR} 2018, Vancouver, BC, Canada, April 30 - May 3, 2018, Conference Track
  Proceedings}. OpenReview.net, 2018.

\bibitem[Caron et~al.(2020)Caron, Misra, Mairal, Goyal, Bojanowski, and
  Joulin]{DBLP:conf/nips/CaronMMGBJ20}
Mathilde Caron, Ishan Misra, Julien Mairal, Priya Goyal, Piotr Bojanowski, and
  Armand Joulin.
\newblock Unsupervised learning of visual features by contrasting cluster
  assignments.
\newblock In Hugo Larochelle, Marc'Aurelio Ranzato, Raia Hadsell,
  Maria{-}Florina Balcan, and Hsuan{-}Tien Lin (eds.), \emph{Advances in Neural
  Information Processing Systems 33: Annual Conference on Neural Information
  Processing Systems 2020, NeurIPS 2020, December 6-12, 2020, virtual}, 2020.

\bibitem[Cer et~al.(2017)Cer, Diab, Agirre, Lopez{-}Gazpio, and
  Specia]{DBLP:conf/semeval/CerDALS17}
Daniel~M. Cer, Mona~T. Diab, Eneko Agirre, I{\~{n}}igo Lopez{-}Gazpio, and
  Lucia Specia.
\newblock Semeval-2017 task 1: Semantic textual similarity multilingual and
  crosslingual focused evaluation.
\newblock In Steven Bethard, Marine Carpuat, Marianna Apidianaki, Saif~M.
  Mohammad, Daniel~M. Cer, and David Jurgens (eds.), \emph{Proceedings of the
  11th International Workshop on Semantic Evaluation, SemEval@ACL 2017,
  Vancouver, Canada, August 3-4, 2017}, pp.\  1--14. Association for
  Computational Linguistics, 2017.

\bibitem[Chen et~al.(2020{\natexlab{a}})Chen, Mei, Wang, Feng, and
  Chen]{DBLP:conf/aaai/ChenMWF020}
Defang Chen, Jian{-}Ping Mei, Can Wang, Yan Feng, and Chun Chen.
\newblock Online knowledge distillation with diverse peers.
\newblock In \emph{The Thirty-Fourth {AAAI} Conference on Artificial
  Intelligence, {AAAI} 2020, The Thirty-Second Innovative Applications of
  Artificial Intelligence Conference, {IAAI} 2020, The Tenth {AAAI} Symposium
  on Educational Advances in Artificial Intelligence, {EAAI} 2020, New York,
  NY, USA, February 7-12, 2020}, pp.\  3430--3437. {AAAI} Press,
  2020{\natexlab{a}}.

\bibitem[Chen et~al.(2020{\natexlab{b}})Chen, Wang, Guo, Xu, Deng, Liu, Ma, Xu,
  Xu, and Gao]{DBLP:journals/corr/abs-2012-00364}
Hanting Chen, Yunhe Wang, Tianyu Guo, Chang Xu, Yiping Deng, Zhenhua Liu, Siwei
  Ma, Chunjing Xu, Chao Xu, and Wen Gao.
\newblock Pre-trained image processing transformer.
\newblock \emph{CoRR}, abs/2012.00364, 2020{\natexlab{b}}.

\bibitem[Chen et~al.(2018{\natexlab{a}})Chen, Si, Li, Chelba, and
  Hsieh]{DBLP:conf/nips/ChenSLCH18}
Patrick~H. Chen, Si~Si, Yang Li, Ciprian Chelba, and Cho{-}Jui Hsieh.
\newblock Groupreduce: Block-wise low-rank approximation for neural language
  model shrinking.
\newblock In \emph{NeurIPS}, pp.\  11011--11021, 2018{\natexlab{a}}.

\bibitem[Chen et~al.(2020{\natexlab{c}})Chen, Frankle, Chang, Liu, Zhang, Wang,
  and Carbin]{DBLP:conf/nips/ChenFC0ZWC20}
Tianlong Chen, Jonathan Frankle, Shiyu Chang, Sijia Liu, Yang Zhang, Zhangyang
  Wang, and Michael Carbin.
\newblock The lottery ticket hypothesis for pre-trained {BERT} networks.
\newblock In Hugo Larochelle, Marc'Aurelio Ranzato, Raia Hadsell,
  Maria{-}Florina Balcan, and Hsuan{-}Tien Lin (eds.), \emph{Advances in Neural
  Information Processing Systems 33: Annual Conference on Neural Information
  Processing Systems 2020, NeurIPS 2020, December 6-12, 2020, virtual},
  2020{\natexlab{c}}.

\bibitem[Chen et~al.(2016{\natexlab{a}})Chen, Xu, Zhang, and
  Guestrin]{DBLP:journals/corr/ChenXZG16}
Tianqi Chen, Bing Xu, Chiyuan Zhang, and Carlos Guestrin.
\newblock Training deep nets with sublinear memory cost.
\newblock \emph{CoRR}, abs/1604.06174, 2016{\natexlab{a}}.

\bibitem[Chen et~al.(2020{\natexlab{d}})Chen, Kornblith, Norouzi, and
  Hinton]{DBLP:conf/icml/ChenK0H20}
Ting Chen, Simon Kornblith, Mohammad Norouzi, and Geoffrey~E. Hinton.
\newblock A simple framework for contrastive learning of visual
  representations.
\newblock In \emph{Proceedings of the 37th International Conference on Machine
  Learning, {ICML} 2020, 13-18 July 2020, Virtual Event}, volume 119 of
  \emph{Proceedings of Machine Learning Research}, pp.\  1597--1607. {PMLR},
  2020{\natexlab{d}}.

\bibitem[Chen et~al.(2016{\natexlab{b}})Chen, Grangier, and
  Auli]{DBLP:conf/acl/ChenGA16}
Wenlin Chen, David Grangier, and Michael Auli.
\newblock Strategies for training large vocabulary neural language models.
\newblock In \emph{{ACL} {(1)}}. The Association for Computer Linguistics,
  2016{\natexlab{b}}.

\bibitem[Chen et~al.(2021{\natexlab{a}})Chen, Gong, and
  Wang]{DBLP:journals/corr/abs-2102-11535}
Wuyang Chen, Xinyu Gong, and Zhangyang Wang.
\newblock Neural architecture search on imagenet in four {GPU} hours: {A}
  theoretically inspired perspective.
\newblock \emph{ICLR}, 2021{\natexlab{a}}.

\bibitem[Chen \& He(2021)Chen and He]{DBLP:conf/cvpr/ChenH21}
Xinlei Chen and Kaiming He.
\newblock Exploring simple siamese representation learning.
\newblock In \emph{{IEEE} Conference on Computer Vision and Pattern
  Recognition, {CVPR} 2021, virtual, June 19-25, 2021}, pp.\  15750--15758.
  Computer Vision Foundation / {IEEE}, 2021.

\bibitem[Chen et~al.(2020{\natexlab{e}})Chen, Fan, Girshick, and
  He]{DBLP:journals/corr/abs-2003-04297}
Xinlei Chen, Haoqi Fan, Ross~B. Girshick, and Kaiming He.
\newblock Improved baselines with momentum contrastive learning.
\newblock \emph{CoRR}, abs/2003.04297, 2020{\natexlab{e}}.

\bibitem[Chen et~al.(2021{\natexlab{b}})Chen, Xie, and
  He]{DBLP:journals/corr/abs-2104-02057}
Xinlei Chen, Saining Xie, and Kaiming He.
\newblock An empirical study of training self-supervised vision transformers.
\newblock \emph{CoRR}, abs/2104.02057, 2021{\natexlab{b}}.

\bibitem[Chen et~al.(2016{\natexlab{c}})Chen, Mou, Xu, Li, and
  Jin]{DBLP:conf/acl/ChenMXLJ16}
Yunchuan Chen, Lili Mou, Yan Xu, Ge~Li, and Zhi Jin.
\newblock Compressing neural language models by sparse word representations.
\newblock In \emph{{ACL} {(1)}}. The Association for Computer Linguistics,
  2016{\natexlab{c}}.

\bibitem[Chen et~al.(2018{\natexlab{b}})Chen, Jin, Kang, Feng, and
  Yan]{DBLP:conf/ijcai/ChenJKFY18}
Yunpeng Chen, Xiaojie Jin, Bingyi Kang, Jiashi Feng, and Shuicheng Yan.
\newblock Sharing residual units through collective tensor factorization to
  improve deep neural networks.
\newblock In \emph{{IJCAI}}, pp.\  635--641. ijcai.org, 2018{\natexlab{b}}.

\bibitem[Child et~al.(2019)Child, Gray, Radford, and
  Sutskever]{DBLP:journals/corr/abs-1904-10509}
Rewon Child, Scott Gray, Alec Radford, and Ilya Sutskever.
\newblock Generating long sequences with sparse transformers.
\newblock \emph{CoRR}, abs/1904.10509, 2019.

\bibitem[Choi et~al.(2017)Choi, El{-}Khamy, and Lee]{DBLP:conf/iclr/ChoiEL17}
Yoojin Choi, Mostafa El{-}Khamy, and Jungwon Lee.
\newblock Towards the limit of network quantization.
\newblock In \emph{5th International Conference on Learning Representations,
  {ICLR} 2017, Toulon, France, April 24-26, 2017, Conference Track
  Proceedings}. OpenReview.net, 2017.

\bibitem[Chollet(2017)]{DBLP:conf/cvpr/Chollet17}
Fran{\c{c}}ois Chollet.
\newblock Xception: Deep learning with depthwise separable convolutions.
\newblock In \emph{{CVPR}}, pp.\  1800--1807. {IEEE} Computer Society, 2017.

\bibitem[Choromanski et~al.(2020)Choromanski, Likhosherstov, Dohan, Song, Gane,
  Sarl{\'{o}}s, Hawkins, Davis, Mohiuddin, Kaiser, Belanger, Colwell, and
  Weller]{DBLP:journals/corr/abs-2009-14794}
Krzysztof Choromanski, Valerii Likhosherstov, David Dohan, Xingyou Song,
  Andreea Gane, Tam{\'{a}}s Sarl{\'{o}}s, Peter Hawkins, Jared Davis, Afroz
  Mohiuddin, Lukasz Kaiser, David Belanger, Lucy Colwell, and Adrian Weller.
\newblock Rethinking attention with performers.
\newblock \emph{CoRR}, abs/2009.14794, 2020.

\bibitem[Chu et~al.(2020)Chu, Zhang, and Li]{DBLP:journals/corr/abs-2005-03566}
Xiangxiang Chu, Bo~Zhang, and Xudong Li.
\newblock Noisy differentiable architecture search.
\newblock \emph{CoRR}, abs/2005.03566, 2020.

\bibitem[Chung et~al.(2020)Chung, Park, Kim, and
  Kwak]{DBLP:conf/icml/ChungPKK20}
Inseop Chung, Seonguk Park, Jangho Kim, and Nojun Kwak.
\newblock Feature-map-level online adversarial knowledge distillation.
\newblock In \emph{Proceedings of the 37th International Conference on Machine
  Learning, {ICML} 2020, 13-18 July 2020, Virtual Event}, volume 119 of
  \emph{Proceedings of Machine Learning Research}, pp.\  2006--2015. {PMLR},
  2020.

\bibitem[Clark et~al.(2020)Clark, Luong, Le, and
  Manning]{DBLP:conf/iclr/ClarkLLM20}
Kevin Clark, Minh{-}Thang Luong, Quoc~V. Le, and Christopher~D. Manning.
\newblock {ELECTRA:} pre-training text encoders as discriminators rather than
  generators.
\newblock In \emph{8th International Conference on Learning Representations,
  {ICLR} 2020, Addis Ababa, Ethiopia, April 26-30, 2020}. OpenReview.net, 2020.

\bibitem[Collobert et~al.(2011)Collobert, Weston, Bottou, Karlen, Kavukcuoglu,
  and Kuksa]{DBLP:journals/jmlr/CollobertWBKKK11}
Ronan Collobert, Jason Weston, L{\'{e}}on Bottou, Michael Karlen, Koray
  Kavukcuoglu, and Pavel~P. Kuksa.
\newblock Natural language processing (almost) from scratch.
\newblock \emph{J. Mach. Learn. Res.}, 12:\penalty0 2493--2537, 2011.

\bibitem[Correia et~al.(2019)Correia, Niculae, and
  Martins]{DBLP:conf/emnlp/CorreiaNM19}
Gon{\c{c}}alo~M. Correia, Vlad Niculae, and Andr{\'{e}} F.~T. Martins.
\newblock Adaptively sparse transformers.
\newblock In \emph{{EMNLP/IJCNLP} {(1)}}, pp.\  2174--2184. Association for
  Computational Linguistics, 2019.

\bibitem[Costa{-}juss{\`{a}} \& Fonollosa(2016)Costa{-}juss{\`{a}} and
  Fonollosa]{DBLP:conf/acl/Costa-JussaF16}
Marta~R. Costa{-}juss{\`{a}} and Jos{\'{e}} A.~R. Fonollosa.
\newblock Character-based neural machine translation.
\newblock In \emph{{ACL} {(2)}}. The Association for Computer Linguistics,
  2016.

\bibitem[Courbariaux et~al.(2015)Courbariaux, Bengio, and
  David]{DBLP:conf/nips/CourbariauxBD15}
Matthieu Courbariaux, Yoshua Bengio, and Jean{-}Pierre David.
\newblock Binaryconnect: Training deep neural networks with binary weights
  during propagations.
\newblock In Corinna Cortes, Neil~D. Lawrence, Daniel~D. Lee, Masashi Sugiyama,
  and Roman Garnett (eds.), \emph{Advances in Neural Information Processing
  Systems 28: Annual Conference on Neural Information Processing Systems 2015,
  December 7-12, 2015, Montreal, Quebec, Canada}, pp.\  3123--3131, 2015.

\bibitem[Cui et~al.(2019)Cui, Che, Liu, Qin, Yang, Wang, and
  Hu]{DBLP:journals/corr/abs-1906-08101}
Yiming Cui, Wanxiang Che, Ting Liu, Bing Qin, Ziqing Yang, Shijin Wang, and
  Guoping Hu.
\newblock Pre-training with whole word masking for chinese {BERT}.
\newblock \emph{CoRR}, abs/1906.08101, 2019.

\bibitem[Dai et~al.(2019)Dai, Yang, Yang, Carbonell, Le, and
  Salakhutdinov]{DBLP:conf/acl/DaiYYCLS19}
Zihang Dai, Zhilin Yang, Yiming Yang, Jaime~G. Carbonell, Quoc~Viet Le, and
  Ruslan Salakhutdinov.
\newblock Transformer-xl: Attentive language models beyond a fixed-length
  context.
\newblock In \emph{{ACL} {(1)}}, pp.\  2978--2988. Association for
  Computational Linguistics, 2019.

\bibitem[Dehghani et~al.(2019)Dehghani, Gouws, Vinyals, Uszkoreit, and
  Kaiser]{DBLP:conf/iclr/DehghaniGVUK19}
Mostafa Dehghani, Stephan Gouws, Oriol Vinyals, Jakob Uszkoreit, and Lukasz
  Kaiser.
\newblock Universal transformers.
\newblock In \emph{7th International Conference on Learning Representations,
  {ICLR} 2019, New Orleans, LA, USA, May 6-9, 2019}. OpenReview.net, 2019.

\bibitem[Deng et~al.(2009)Deng, Dong, Socher, Li, Li, and
  Fei{-}Fei]{DBLP:conf/cvpr/DengDSLL009}
Jia Deng, Wei Dong, Richard Socher, Li{-}Jia Li, Kai Li, and Li~Fei{-}Fei.
\newblock Imagenet: {A} large-scale hierarchical image database.
\newblock In \emph{2009 {IEEE} Computer Society Conference on Computer Vision
  and Pattern Recognition {(CVPR} 2009), 20-25 June 2009, Miami, Florida,
  {USA}}, pp.\  248--255. {IEEE} Computer Society, 2009.

\bibitem[Denton et~al.(2014)Denton, Zaremba, Bruna, LeCun, and
  Fergus]{DBLP:conf/nips/DentonZBLF14}
Emily~L. Denton, Wojciech Zaremba, Joan Bruna, Yann LeCun, and Rob Fergus.
\newblock Exploiting linear structure within convolutional networks for
  efficient evaluation.
\newblock In Zoubin Ghahramani, Max Welling, Corinna Cortes, Neil~D. Lawrence,
  and Kilian~Q. Weinberger (eds.), \emph{Advances in Neural Information
  Processing Systems 27: Annual Conference on Neural Information Processing
  Systems 2014, December 8-13 2014, Montreal, Quebec, Canada}, pp.\
  1269--1277, 2014.

\bibitem[Devlin et~al.(2014)Devlin, Zbib, Huang, Lamar, Schwartz, and
  Makhoul]{DBLP:conf/acl/DevlinZHLSM14}
Jacob Devlin, Rabih Zbib, Zhongqiang Huang, Thomas Lamar, Richard~M. Schwartz,
  and John Makhoul.
\newblock Fast and robust neural network joint models for statistical machine
  translation.
\newblock In \emph{{ACL} {(1)}}, pp.\  1370--1380. The Association for Computer
  Linguistics, 2014.

\bibitem[Devlin et~al.(2019)Devlin, Chang, Lee, and
  Toutanova]{DBLP:conf/naacl/DevlinCLT19}
Jacob Devlin, Ming{-}Wei Chang, Kenton Lee, and Kristina Toutanova.
\newblock {BERT:} pre-training of deep bidirectional transformers for language
  understanding.
\newblock In \emph{Proceedings of the 2019 Conference of the North American
  Chapter of the Association for Computational Linguistics: Human Language
  Technologies, {NAACL-HLT} 2019, Minneapolis, MN, USA, June 2-7, 2019, Volume
  1 (Long and Short Papers)}, pp.\  4171--4186, 2019.

\bibitem[Dolan \& Brockett(2005)Dolan and Brockett]{DBLP:conf/acl-iwp/DolanB05}
William~B. Dolan and Chris Brockett.
\newblock Automatically constructing a corpus of sentential paraphrases.
\newblock In \emph{Proceedings of the Third International Workshop on
  Paraphrasing, IWP@IJCNLP 2005, Jeju Island, Korea, October 2005, 2005}. Asian
  Federation of Natural Language Processing, 2005.

\bibitem[Dong et~al.(2015)Dong, Wu, He, Yu, and Wang]{DBLP:conf/acl/DongWHYW15}
Daxiang Dong, Hua Wu, Wei He, Dianhai Yu, and Haifeng Wang.
\newblock Multi-task learning for multiple language translation.
\newblock In \emph{{ACL} {(1)}}, pp.\  1723--1732. The Association for Computer
  Linguistics, 2015.

\bibitem[Dong et~al.(2019)Dong, Yang, Wang, Wei, Liu, Wang, Gao, Zhou, and
  Hon]{DBLP:conf/nips/00040WWLWGZH19}
Li~Dong, Nan Yang, Wenhui Wang, Furu Wei, Xiaodong Liu, Yu~Wang, Jianfeng Gao,
  Ming Zhou, and Hsiao{-}Wuen Hon.
\newblock Unified language model pre-training for natural language
  understanding and generation.
\newblock In Hanna~M. Wallach, Hugo Larochelle, Alina Beygelzimer, Florence
  d'Alch{\'{e}}{-}Buc, Emily~B. Fox, and Roman Garnett (eds.), \emph{Advances
  in Neural Information Processing Systems 32: Annual Conference on Neural
  Information Processing Systems 2019, NeurIPS 2019, December 8-14, 2019,
  Vancouver, BC, Canada}, pp.\  13042--13054, 2019.

\bibitem[Dong \& Yang(2019)Dong and Yang]{DBLP:conf/cvpr/DongY19}
Xuanyi Dong and Yi~Yang.
\newblock Searching for a robust neural architecture in four {GPU} hours.
\newblock In \emph{{IEEE} Conference on Computer Vision and Pattern
  Recognition, {CVPR} 2019, Long Beach, CA, USA, June 16-20, 2019}, pp.\
  1761--1770, 2019.

\bibitem[Dong \& Yang(2020)Dong and Yang]{DBLP:conf/iclr/Dong020}
Xuanyi Dong and Yi~Yang.
\newblock Nas-bench-201: Extending the scope of reproducible neural
  architecture search.
\newblock In \emph{{ICLR}}. OpenReview.net, 2020.

\bibitem[Dong et~al.(2017)Dong, Li, and Ni]{DBLP:conf/bmvc/DongLN17}
Yinpeng Dong, Jianguo Li, and Renkun Ni.
\newblock Learning accurate low-bit deep neural networks with stochastic
  quantization.
\newblock In \emph{British Machine Vision Conference 2017, {BMVC} 2017, London,
  UK, September 4-7, 2017}. {BMVA} Press, 2017.

\bibitem[Ducoffe \& Precioso(2018)Ducoffe and
  Precioso]{DBLP:journals/corr/abs-1802-09841}
Melanie Ducoffe and Fr{\'{e}}d{\'{e}}ric Precioso.
\newblock Adversarial active learning for deep networks: a margin based
  approach.
\newblock \emph{CoRR}, abs/1802.09841, 2018.

\bibitem[Duong et~al.(2015)Duong, Cohn, Bird, and
  Cook]{DBLP:conf/acl/DuongCBC15}
Long Duong, Trevor Cohn, Steven Bird, and Paul Cook.
\newblock Low resource dependency parsing: Cross-lingual parameter sharing in a
  neural network parser.
\newblock In \emph{{ACL} {(2)}}, pp.\  845--850. The Association for Computer
  Linguistics, 2015.

\bibitem[Elbayad et~al.(2020)Elbayad, Gu, Grave, and
  Auli]{DBLP:conf/iclr/ElbayadGGA20}
Maha Elbayad, Jiatao Gu, Edouard Grave, and Michael Auli.
\newblock Depth-adaptive transformer.
\newblock In \emph{8th International Conference on Learning Representations,
  {ICLR} 2020, Addis Ababa, Ethiopia, April 26-30, 2020}. OpenReview.net, 2020.

\bibitem[Elsken et~al.(2018)Elsken, Metzen, and
  Hutter]{DBLP:journals/corr/abs-1808-05377}
Thomas Elsken, Jan~Hendrik Metzen, and Frank Hutter.
\newblock Neural architecture search: {A} survey.
\newblock \emph{CoRR}, abs/1808.05377, 2018.

\bibitem[Everingham et~al.(2010)Everingham, Gool, Williams, Winn, and
  Zisserman]{DBLP:journals/ijcv/EveringhamGWWZ10}
Mark Everingham, Luc~Van Gool, Christopher K.~I. Williams, John~M. Winn, and
  Andrew Zisserman.
\newblock The pascal visual object classes {(VOC)} challenge.
\newblock \emph{Int. J. Comput. Vis.}, 88\penalty0 (2):\penalty0 303--338,
  2010.

\bibitem[Fan et~al.(2020{\natexlab{a}})Fan, Bhosale, Schwenk, Ma, El{-}Kishky,
  Goyal, Baines, Celebi, Wenzek, Chaudhary, Goyal, Birch, Liptchinsky, Edunov,
  Grave, Auli, and Joulin]{DBLP:journals/corr/abs-2010-11125}
Angela Fan, Shruti Bhosale, Holger Schwenk, Zhiyi Ma, Ahmed El{-}Kishky,
  Siddharth Goyal, Mandeep Baines, Onur Celebi, Guillaume Wenzek, Vishrav
  Chaudhary, Naman Goyal, Tom Birch, Vitaliy Liptchinsky, Sergey Edunov,
  Edouard Grave, Michael Auli, and Armand Joulin.
\newblock Beyond english-centric multilingual machine translation.
\newblock \emph{CoRR}, abs/2010.11125, 2020{\natexlab{a}}.

\bibitem[Fan et~al.(2020{\natexlab{b}})Fan, Grave, and
  Joulin]{DBLP:conf/iclr/FanGJ20}
Angela Fan, Edouard Grave, and Armand Joulin.
\newblock Reducing transformer depth on demand with structured dropout.
\newblock In \emph{{ICLR}}. OpenReview.net, 2020{\natexlab{b}}.

\bibitem[Fan et~al.(2020{\natexlab{c}})Fan, Stock, Graham, Grave, Gribonval,
  J{\'{e}}gou, and Joulin]{DBLP:journals/corr/abs-2004-07320}
Angela Fan, Pierre Stock, Benjamin Graham, Edouard Grave, R{\'{e}}mi Gribonval,
  Herv{\'{e}} J{\'{e}}gou, and Armand Joulin.
\newblock Training with quantization noise for extreme model compression.
\newblock \emph{CoRR}, abs/2004.07320, 2020{\natexlab{c}}.

\bibitem[Fang \& Xie(2020)Fang and Xie]{DBLP:journals/corr/abs-2005-12766}
Hongchao Fang and Pengtao Xie.
\newblock {CERT:} contrastive self-supervised learning for language
  understanding.
\newblock \emph{CoRR}, abs/2005.12766, 2020.

\bibitem[Faruqui et~al.(2015)Faruqui, Tsvetkov, Yogatama, Dyer, and
  Smith]{DBLP:conf/acl/FaruquiTYDS15}
Manaal Faruqui, Yulia Tsvetkov, Dani Yogatama, Chris Dyer, and Noah~A. Smith.
\newblock Sparse overcomplete word vector representations.
\newblock In \emph{{ACL} {(1)}}, pp.\  1491--1500. The Association for Computer
  Linguistics, 2015.

\bibitem[Fedus et~al.(2021)Fedus, Zoph, and
  Shazeer]{DBLP:journals/corr/abs-2101-03961}
William Fedus, Barret Zoph, and Noam Shazeer.
\newblock Switch transformers: Scaling to trillion parameter models with simple
  and efficient sparsity.
\newblock \emph{CoRR}, abs/2101.03961, 2021.

\bibitem[Feng et~al.(2020)Feng, Wang, Li, Yang, Peng, and
  Ding]{DBLP:conf/ictai/FengWLYPD20}
Boyuan Feng, Yuke Wang, Xu~Li, Shu Yang, Xueqiao Peng, and Yufei Ding.
\newblock Sgquant: Squeezing the last bit on graph neural networks with
  specialized quantization.
\newblock In \emph{32nd {IEEE} International Conference on Tools with
  Artificial Intelligence, {ICTAI} 2020, Baltimore, MD, USA, November 9-11,
  2020}, pp.\  1044--1052. {IEEE}, 2020.

\bibitem[Fiesler et~al.(1990)Fiesler, Choudry, and
  Caulfield]{fiesler1990weight}
Emile Fiesler, Amar Choudry, and H.~John Caulfield.
\newblock {Weight discretization paradigm for optical neural networks}.
\newblock In Hartmut Bartelt (ed.), \emph{Optical Interconnections and
  Networks}, volume 1281, pp.\  164 -- 173. International Society for Optics
  and Photonics, SPIE, 1990.

\bibitem[Figurnov et~al.(2017)Figurnov, Collins, Zhu, Zhang, Huang, Vetrov, and
  Salakhutdinov]{DBLP:conf/cvpr/FigurnovCZZHVS17}
Michael Figurnov, Maxwell~D. Collins, Yukun Zhu, Li~Zhang, Jonathan Huang,
  Dmitry~P. Vetrov, and Ruslan Salakhutdinov.
\newblock Spatially adaptive computation time for residual networks.
\newblock In \emph{2017 {IEEE} Conference on Computer Vision and Pattern
  Recognition, {CVPR} 2017, Honolulu, HI, USA, July 21-26, 2017}, pp.\
  1790--1799. {IEEE} Computer Society, 2017.

\bibitem[Firat et~al.(2016)Firat, Cho, and Bengio]{DBLP:conf/naacl/FiratCB16}
Orhan Firat, Kyunghyun Cho, and Yoshua Bengio.
\newblock Multi-way, multilingual neural machine translation with a shared
  attention mechanism.
\newblock In \emph{{HLT-NAACL}}, pp.\  866--875. The Association for
  Computational Linguistics, 2016.

\bibitem[Firat et~al.(2017)Firat, Cho, Sankaran, Yarman{-}Vural, and
  Bengio]{DBLP:journals/csl/FiratCSYB17}
Orhan Firat, Kyunghyun Cho, Baskaran Sankaran, Fatos~T. Yarman{-}Vural, and
  Yoshua Bengio.
\newblock Multi-way, multilingual neural machine translation.
\newblock \emph{Comput. Speech Lang.}, 45:\penalty0 236--252, 2017.

\bibitem[Frankle \& Carbin(2019)Frankle and Carbin]{DBLP:conf/iclr/FrankleC19}
Jonathan Frankle and Michael Carbin.
\newblock The lottery ticket hypothesis: Finding sparse, trainable neural
  networks.
\newblock In \emph{7th International Conference on Learning Representations,
  {ICLR} 2019, New Orleans, LA, USA, May 6-9, 2019}. OpenReview.net, 2019.

\bibitem[Freytag et~al.(2014)Freytag, Rodner, and
  Denzler]{DBLP:conf/eccv/FreytagRD14}
Alexander Freytag, Erik Rodner, and Joachim Denzler.
\newblock Selecting influential examples: Active learning with expected model
  output changes.
\newblock In \emph{{ECCV} {(4)}}, volume 8692 of \emph{Lecture Notes in
  Computer Science}, pp.\  562--577. Springer, 2014.

\bibitem[Gao et~al.(2021)Gao, Yao, and Chen]{DBLP:journals/corr/abs-2104-08821}
Tianyu Gao, Xingcheng Yao, and Danqi Chen.
\newblock Simcse: Simple contrastive learning of sentence embeddings.
\newblock \emph{CoRR}, abs/2104.08821, 2021.

\bibitem[Geng et~al.(2021)Geng, Gao, Fu, and
  Zhang]{DBLP:journals/corr/abs-2101-09755}
Shijie Geng, Peng Gao, Zuohui Fu, and Yongfeng Zhang.
\newblock Romebert: Robust training of multi-exit {BERT}.
\newblock \emph{CoRR}, abs/2101.09755, 2021.

\bibitem[Gholami et~al.(2021)Gholami, Kim, Dong, Yao, Mahoney, and
  Keutzer]{DBLP:journals/corr/abs-2103-13630}
Amir Gholami, Sehoon Kim, Zhen Dong, Zhewei Yao, Michael~W. Mahoney, and Kurt
  Keutzer.
\newblock A survey of quantization methods for efficient neural network
  inference.
\newblock \emph{CoRR}, abs/2103.13630, 2021.

\bibitem[Ghosh et~al.(2011)Ghosh, Johansson, Riccardi, and
  Tonelli]{DBLP:conf/ijcnlp/GhoshJRT11}
Sucheta Ghosh, Richard Johansson, Giuseppe Riccardi, and Sara Tonelli.
\newblock Shallow discourse parsing with conditional random fields.
\newblock In \emph{Fifth International Joint Conference on Natural Language
  Processing, {IJCNLP} 2011, Chiang Mai, Thailand, November 8-13, 2011}, pp.\
  1071--1079. The Association for Computer Linguistics, 2011.

\bibitem[Giorgi et~al.(2021)Giorgi, Nitski, Wang, and
  Bader]{DBLP:conf/acl/GiorgiNWB20}
John~M. Giorgi, Osvald Nitski, Bo~Wang, and Gary~D. Bader.
\newblock Declutr: Deep contrastive learning for unsupervised textual
  representations.
\newblock In Chengqing Zong, Fei Xia, Wenjie Li, and Roberto Navigli (eds.),
  \emph{Proceedings of the 59th Annual Meeting of the Association for
  Computational Linguistics and the 11th International Joint Conference on
  Natural Language Processing, {ACL/IJCNLP} 2021, (Volume 1: Long Papers),
  Virtual Event, August 1-6, 2021}, pp.\  879--895. Association for
  Computational Linguistics, 2021.

\bibitem[Glorot \& Bengio(2010)Glorot and Bengio]{DBLP:journals/jmlr/GlorotB10}
Xavier Glorot and Yoshua Bengio.
\newblock Understanding the difficulty of training deep feedforward neural
  networks.
\newblock In \emph{{AISTATS}}, volume~9 of \emph{{JMLR} Proceedings}, pp.\
  249--256. JMLR.org, 2010.

\bibitem[Gomez et~al.(2017)Gomez, Ren, Urtasun, and
  Grosse]{DBLP:conf/nips/GomezRUG17}
Aidan~N. Gomez, Mengye Ren, Raquel Urtasun, and Roger~B. Grosse.
\newblock The reversible residual network: Backpropagation without storing
  activations.
\newblock In \emph{{NIPS}}, pp.\  2214--2224, 2017.

\bibitem[Gong et~al.(2019)Gong, He, Li, Qin, Wang, and
  Liu]{DBLP:conf/icml/GongHLQWL19}
Linyuan Gong, Di~He, Zhuohan Li, Tao Qin, Liwei Wang, and Tie{-}Yan Liu.
\newblock Efficient training of {BERT} by progressively stacking.
\newblock In Kamalika Chaudhuri and Ruslan Salakhutdinov (eds.),
  \emph{Proceedings of the 36th International Conference on Machine Learning,
  {ICML} 2019, 9-15 June 2019, Long Beach, California, {USA}}, volume~97 of
  \emph{Proceedings of Machine Learning Research}, pp.\  2337--2346. {PMLR},
  2019.

\bibitem[Gong et~al.(2014)Gong, Liu, Yang, and
  Bourdev]{DBLP:journals/corr/GongLYB14}
Yunchao Gong, Liu Liu, Ming Yang, and Lubomir~D. Bourdev.
\newblock Compressing deep convolutional networks using vector quantization.
\newblock \emph{CoRR}, abs/1412.6115, 2014.

\bibitem[Gormez \& Koyuncu(2021)Gormez and
  Koyuncu]{DBLP:journals/corr/abs-2103-01148}
Alperen Gormez and Erdem Koyuncu.
\newblock Class means as an early exit decision mechanism.
\newblock \emph{CoRR}, abs/2103.01148, 2021.

\bibitem[Grachev et~al.(2017)Grachev, Ignatov, and
  Savchenko]{DBLP:conf/premi/GrachevIS17}
Artem~M. Grachev, Dmitry~I. Ignatov, and Andrey~V. Savchenko.
\newblock Neural networks compression for language modeling.
\newblock In \emph{PReMI}, volume 10597 of \emph{Lecture Notes in Computer
  Science}, pp.\  351--357. Springer, 2017.

\bibitem[Graves(2016)]{DBLP:journals/corr/Graves16}
Alex Graves.
\newblock Adaptive computation time for recurrent neural networks.
\newblock \emph{CoRR}, abs/1603.08983, 2016.

\bibitem[Grill et~al.(2020)Grill, Strub, Altch{\'{e}}, Tallec, Richemond,
  Buchatskaya, Doersch, Pires, Guo, Azar, Piot, Kavukcuoglu, Munos, and
  Valko]{DBLP:conf/nips/GrillSATRBDPGAP20}
Jean{-}Bastien Grill, Florian Strub, Florent Altch{\'{e}}, Corentin Tallec,
  Pierre~H. Richemond, Elena Buchatskaya, Carl Doersch, Bernardo~{\'{A}}vila
  Pires, Zhaohan Guo, Mohammad~Gheshlaghi Azar, Bilal Piot, Koray Kavukcuoglu,
  R{\'{e}}mi Munos, and Michal Valko.
\newblock Bootstrap your own latent - {A} new approach to self-supervised
  learning.
\newblock In Hugo Larochelle, Marc'Aurelio Ranzato, Raia Hadsell,
  Maria{-}Florina Balcan, and Hsuan{-}Tien Lin (eds.), \emph{Advances in Neural
  Information Processing Systems 33: Annual Conference on Neural Information
  Processing Systems 2020, NeurIPS 2020, December 6-12, 2020, virtual}, 2020.

\bibitem[Gruslys et~al.(2016)Gruslys, Munos, Danihelka, Lanctot, and
  Graves]{DBLP:conf/nips/GruslysMDLG16}
Audrunas Gruslys, R{\'{e}}mi Munos, Ivo Danihelka, Marc Lanctot, and Alex
  Graves.
\newblock Memory-efficient backpropagation through time.
\newblock In \emph{{NIPS}}, pp.\  4125--4133, 2016.

\bibitem[Guo et~al.(2020)Guo, Rush, and Kim]{DBLP:journals/corr/abs-2012-07463}
Demi Guo, Alexander~M. Rush, and Yoon Kim.
\newblock Parameter-efficient transfer learning with diff pruning.
\newblock \emph{CoRR}, abs/2012.07463, 2020.

\bibitem[Guo(2018)]{DBLP:journals/corr/abs-1808-04752}
Yunhui Guo.
\newblock A survey on methods and theories of quantized neural networks.
\newblock \emph{CoRR}, abs/1808.04752, 2018.

\bibitem[Gupta et~al.(2015)Gupta, Agrawal, Gopalakrishnan, and
  Narayanan]{DBLP:conf/icml/GuptaAGN15}
Suyog Gupta, Ankur Agrawal, Kailash Gopalakrishnan, and Pritish Narayanan.
\newblock Deep learning with limited numerical precision.
\newblock In Francis~R. Bach and David~M. Blei (eds.), \emph{Proceedings of the
  32nd International Conference on Machine Learning, {ICML} 2015, Lille,
  France, 6-11 July 2015}, volume~37 of \emph{{JMLR} Workshop and Conference
  Proceedings}, pp.\  1737--1746. JMLR.org, 2015.

\bibitem[Gutmann \& Hyv{\"{a}}rinen(2010)Gutmann and
  Hyv{\"{a}}rinen]{DBLP:journals/jmlr/GutmannH10}
Michael Gutmann and Aapo Hyv{\"{a}}rinen.
\newblock Noise-contrastive estimation: {A} new estimation principle for
  unnormalized statistical models.
\newblock In Yee~Whye Teh and D.~Mike Titterington (eds.), \emph{Proceedings of
  the Thirteenth International Conference on Artificial Intelligence and
  Statistics, {AISTATS} 2010, Chia Laguna Resort, Sardinia, Italy, May 13-15,
  2010}, volume~9 of \emph{{JMLR} Proceedings}, pp.\  297--304. JMLR.org, 2010.

\bibitem[Ha et~al.(2016)Ha, Niehues, and Waibel]{DBLP:journals/corr/HaNW16}
Thanh{-}Le Ha, Jan Niehues, and Alexander~H. Waibel.
\newblock Toward multilingual neural machine translation with universal encoder
  and decoder.
\newblock \emph{CoRR}, abs/1611.04798, 2016.

\bibitem[Han et~al.(2015)Han, Pool, Tran, and Dally]{DBLP:conf/nips/HanPTD15}
Song Han, Jeff Pool, John Tran, and William~J. Dally.
\newblock Learning both weights and connections for efficient neural network.
\newblock In \emph{{NIPS}}, pp.\  1135--1143, 2015.

\bibitem[Han et~al.(2016)Han, Mao, and Dally]{DBLP:journals/corr/HanMD15}
Song Han, Huizi Mao, and William~J. Dally.
\newblock Deep compression: Compressing deep neural network with pruning,
  trained quantization and huffman coding.
\newblock In Yoshua Bengio and Yann LeCun (eds.), \emph{4th International
  Conference on Learning Representations, {ICLR} 2016, San Juan, Puerto Rico,
  May 2-4, 2016, Conference Track Proceedings}, 2016.

\bibitem[Han et~al.(2021{\natexlab{a}})Han, Zhang, Ding, Gu, Liu, Huo, Qiu,
  Zhang, Han, Huang, Jin, Lan, Liu, Liu, Lu, Qiu, Song, Tang, Wen, Yuan, Zhao,
  and Zhu]{DBLP:journals/corr/abs-2106-07139}
Xu~Han, Zhengyan Zhang, Ning Ding, Yuxian Gu, Xiao Liu, Yuqi Huo, Jiezhong Qiu,
  Liang Zhang, Wentao Han, Minlie Huang, Qin Jin, Yanyan Lan, Yang Liu, Zhiyuan
  Liu, Zhiwu Lu, Xipeng Qiu, Ruihua Song, Jie Tang, Ji{-}Rong Wen, Jinhui Yuan,
  Wayne~Xin Zhao, and Jun Zhu.
\newblock Pre-trained models: Past, present and future.
\newblock \emph{CoRR}, abs/2106.07139, 2021{\natexlab{a}}.

\bibitem[Han et~al.(2021{\natexlab{b}})Han, Huang, Song, Yang, Wang, and
  Wang]{DBLP:journals/corr/abs-2102-04906}
Yizeng Han, Gao Huang, Shiji Song, Le~Yang, Honghui Wang, and Yulin Wang.
\newblock Dynamic neural networks: {A} survey.
\newblock \emph{CoRR}, abs/2102.04906, 2021{\natexlab{b}}.

\bibitem[Hanin \& Rolnick(2018)Hanin and Rolnick]{DBLP:conf/nips/HaninR18}
Boris Hanin and David Rolnick.
\newblock How to start training: The effect of initialization and architecture.
\newblock In \emph{NeurIPS}, pp.\  569--579, 2018.

\bibitem[Hanson \& Pratt(1988)Hanson and Pratt]{DBLP:conf/nips/HansonP88}
Stephen~Jose Hanson and Lorien~Y. Pratt.
\newblock Comparing biases for minimal network construction with
  back-propagation.
\newblock In David~S. Touretzky (ed.), \emph{Advances in Neural Information
  Processing Systems 1, {[NIPS} Conference, Denver, Colorado, USA, 1988]}, pp.\
   177--185. Morgan Kaufmann, 1988.

\bibitem[Hashimoto et~al.(2017)Hashimoto, Xiong, Tsuruoka, and
  Socher]{DBLP:conf/emnlp/HashimotoXTS17}
Kazuma Hashimoto, Caiming Xiong, Yoshimasa Tsuruoka, and Richard Socher.
\newblock A joint many-task model: Growing a neural network for multiple {NLP}
  tasks.
\newblock In \emph{{EMNLP}}, pp.\  1923--1933. Association for Computational
  Linguistics, 2017.

\bibitem[Hassibi et~al.(1993)Hassibi, Stork, and
  Wolff]{DBLP:conf/nips/HassibiSW93}
Babak Hassibi, David~G. Stork, and Gregory~J. Wolff.
\newblock Optimal brain surgeon: Extensions and performance comparison.
\newblock In \emph{{NIPS}}, pp.\  263--270. Morgan Kaufmann, 1993.

\bibitem[He et~al.(2015)He, Zhang, Ren, and Sun]{DBLP:conf/iccv/HeZRS15}
Kaiming He, Xiangyu Zhang, Shaoqing Ren, and Jian Sun.
\newblock Delving deep into rectifiers: Surpassing human-level performance on
  imagenet classification.
\newblock In \emph{{ICCV}}, pp.\  1026--1034. {IEEE} Computer Society, 2015.

\bibitem[He et~al.(2016{\natexlab{a}})He, Zhang, Ren, and
  Sun]{DBLP:conf/cvpr/HeZRS16}
Kaiming He, Xiangyu Zhang, Shaoqing Ren, and Jian Sun.
\newblock Deep residual learning for image recognition.
\newblock In \emph{2016 {IEEE} Conference on Computer Vision and Pattern
  Recognition, {CVPR} 2016, Las Vegas, NV, USA, June 27-30, 2016}, pp.\
  770--778. {IEEE} Computer Society, 2016{\natexlab{a}}.

\bibitem[He et~al.(2017{\natexlab{a}})He, Gkioxari, Doll{\'{a}}r, and
  Girshick]{DBLP:conf/iccv/HeGDG17}
Kaiming He, Georgia Gkioxari, Piotr Doll{\'{a}}r, and Ross~B. Girshick.
\newblock Mask {R-CNN}.
\newblock In \emph{{ICCV}}, pp.\  2980--2988. {IEEE} Computer Society,
  2017{\natexlab{a}}.

\bibitem[He et~al.(2020)He, Fan, Wu, Xie, and
  Girshick]{DBLP:conf/cvpr/He0WXG20}
Kaiming He, Haoqi Fan, Yuxin Wu, Saining Xie, and Ross~B. Girshick.
\newblock Momentum contrast for unsupervised visual representation learning.
\newblock In \emph{2020 {IEEE/CVF} Conference on Computer Vision and Pattern
  Recognition, {CVPR} 2020, Seattle, WA, USA, June 13-19, 2020}, pp.\
  9726--9735. {IEEE}, 2020.

\bibitem[He et~al.(2016{\natexlab{b}})He, Wen, Zhou, Wu, Yao, Zhou, and
  Zou]{DBLP:journals/corr/HeWZWYZZ16}
Qinyao He, He~Wen, Shuchang Zhou, Yuxin Wu, Cong Yao, Xinyu Zhou, and Yuheng
  Zou.
\newblock Effective quantization methods for recurrent neural networks.
\newblock \emph{CoRR}, abs/1611.10176, 2016{\natexlab{b}}.

\bibitem[He et~al.(2019)He, Jin, Ding, Yi, and Yan]{DBLP:conf/icmcs/HeJDYY19}
Tao He, Xiaoming Jin, Guiguang Ding, Lan Yi, and Chenggang Yan.
\newblock Towards better uncertainty sampling: Active learning with multiple
  views for deep convolutional neural network.
\newblock In \emph{{ICME}}, pp.\  1360--1365. {IEEE}, 2019.

\bibitem[He et~al.(2017{\natexlab{b}})He, Zhang, and
  Sun]{DBLP:conf/iccv/HeZS17}
Yihui He, Xiangyu Zhang, and Jian Sun.
\newblock Channel pruning for accelerating very deep neural networks.
\newblock In \emph{{ICCV}}, pp.\  1398--1406. {IEEE} Computer Society,
  2017{\natexlab{b}}.

\bibitem[He et~al.(2018)He, Lin, Liu, Wang, Li, and
  Han]{DBLP:conf/eccv/HeLLWLH18}
Yihui He, Ji~Lin, Zhijian Liu, Hanrui Wang, Li{-}Jia Li, and Song Han.
\newblock {AMC:} automl for model compression and acceleration on mobile
  devices.
\newblock In \emph{Computer Vision - {ECCV} 2018 - 15th European Conference,
  Munich, Germany, September 8-14, 2018, Proceedings, Part {VII}}, pp.\
  815--832, 2018.

\bibitem[Heo et~al.(2019)Heo, Lee, Yun, and Choi]{DBLP:conf/aaai/HeoLY019a}
Byeongho Heo, Minsik Lee, Sangdoo Yun, and Jin~Young Choi.
\newblock Knowledge transfer via distillation of activation boundaries formed
  by hidden neurons.
\newblock In \emph{The Thirty-Third {AAAI} Conference on Artificial
  Intelligence, {AAAI} 2019, The Thirty-First Innovative Applications of
  Artificial Intelligence Conference, {IAAI} 2019, The Ninth {AAAI} Symposium
  on Educational Advances in Artificial Intelligence, {EAAI} 2019, Honolulu,
  Hawaii, USA, January 27 - February 1, 2019}, pp.\  3779--3787. {AAAI} Press,
  2019.

\bibitem[Hinton et~al.(2006)Hinton, Osindero, and
  Teh]{DBLP:journals/neco/HintonOT06}
Geoffrey~E. Hinton, Simon Osindero, and Yee~Whye Teh.
\newblock A fast learning algorithm for deep belief nets.
\newblock \emph{Neural Comput.}, 18\penalty0 (7):\penalty0 1527--1554, 2006.

\bibitem[Hinton et~al.(2015)Hinton, Vinyals, and
  Dean]{DBLP:journals/corr/HintonVD15}
Geoffrey~E. Hinton, Oriol Vinyals, and Jeffrey Dean.
\newblock Distilling the knowledge in a neural network.
\newblock \emph{CoRR}, abs/1503.02531, 2015.

\bibitem[Hoang \& Jo(2018)Hoang and Jo]{DBLP:journals/corr/abs-1811-07083}
Van{-}Thanh Hoang and Kang{-}Hyun Jo.
\newblock Pydmobilenet: Improved version of mobilenets with pyramid depthwise
  separable convolution.
\newblock \emph{CoRR}, abs/1811.07083, 2018.

\bibitem[Hochreiter \& Schmidhuber(1997)Hochreiter and
  Schmidhuber]{DBLP:journals/neco/HochreiterS97}
Sepp Hochreiter and J{\"{u}}rgen Schmidhuber.
\newblock Long short-term memory.
\newblock \emph{Neural Comput.}, 9\penalty0 (8):\penalty0 1735--1780, 1997.

\bibitem[Houlsby et~al.(2019)Houlsby, Giurgiu, Jastrzebski, Morrone,
  de~Laroussilhe, Gesmundo, Attariyan, and
  Gelly]{DBLP:conf/icml/HoulsbyGJMLGAG19}
Neil Houlsby, Andrei Giurgiu, Stanislaw Jastrzebski, Bruna Morrone, Quentin
  de~Laroussilhe, Andrea Gesmundo, Mona Attariyan, and Sylvain Gelly.
\newblock Parameter-efficient transfer learning for {NLP}.
\newblock In \emph{{ICML}}, volume~97 of \emph{Proceedings of Machine Learning
  Research}, pp.\  2790--2799. {PMLR}, 2019.

\bibitem[Howard et~al.(2019)Howard, Pang, Adam, Le, Sandler, Chen, Wang, Chen,
  Tan, Chu, Vasudevan, and Zhu]{DBLP:conf/iccv/HowardPALSCWCTC19}
Andrew Howard, Ruoming Pang, Hartwig Adam, Quoc~V. Le, Mark Sandler, Bo~Chen,
  Weijun Wang, Liang{-}Chieh Chen, Mingxing Tan, Grace Chu, Vijay Vasudevan,
  and Yukun Zhu.
\newblock Searching for mobilenetv3.
\newblock In \emph{2019 {IEEE/CVF} International Conference on Computer Vision,
  {ICCV} 2019, Seoul, Korea (South), October 27 - November 2, 2019}, pp.\
  1314--1324. {IEEE}, 2019.

\bibitem[Howard et~al.(2017)Howard, Zhu, Chen, Kalenichenko, Wang, Weyand,
  Andreetto, and Adam]{DBLP:journals/corr/HowardZCKWWAA17}
Andrew~G. Howard, Menglong Zhu, Bo~Chen, Dmitry Kalenichenko, Weijun Wang,
  Tobias Weyand, Marco Andreetto, and Hartwig Adam.
\newblock Mobilenets: Efficient convolutional neural networks for mobile vision
  applications.
\newblock \emph{CoRR}, abs/1704.04861, 2017.

\bibitem[Huang et~al.(2018)Huang, Chen, Li, Wu, van~der Maaten, and
  Weinberger]{DBLP:conf/iclr/HuangCLWMW18}
Gao Huang, Danlu Chen, Tianhong Li, Felix Wu, Laurens van~der Maaten, and
  Kilian~Q. Weinberger.
\newblock Multi-scale dense networks for resource efficient image
  classification.
\newblock In \emph{6th International Conference on Learning Representations,
  {ICLR} 2018, Vancouver, BC, Canada, April 30 - May 3, 2018, Conference Track
  Proceedings}. OpenReview.net, 2018.

\bibitem[Huang et~al.(2013)Huang, Li, Yu, Deng, and
  Gong]{DBLP:conf/icassp/HuangLYDG13}
Jui{-}Ting Huang, Jinyu Li, Dong Yu, Li~Deng, and Yifan Gong.
\newblock Cross-language knowledge transfer using multilingual deep neural
  network with shared hidden layers.
\newblock In \emph{{ICASSP}}, pp.\  7304--7308. {IEEE}, 2013.

\bibitem[Huang et~al.(2012)Huang, Fayong, and Guo]{DBLP:conf/naacl/HuangFG12}
Liang Huang, Suphan Fayong, and Yang Guo.
\newblock Structured perceptron with inexact search.
\newblock In \emph{Human Language Technologies: Conference of the North
  American Chapter of the Association of Computational Linguistics,
  Proceedings, June 3-8, 2012, Montr{\'{e}}al, Canada}, pp.\  142--151. The
  Association for Computational Linguistics, 2012.

\bibitem[Huang \& Wang(2017)Huang and Wang]{DBLP:journals/corr/HuangW17a}
Zehao Huang and Naiyan Wang.
\newblock Like what you like: Knowledge distill via neuron selectivity
  transfer.
\newblock \emph{CoRR}, abs/1707.01219, 2017.

\bibitem[Hubara et~al.(2017)Hubara, Courbariaux, Soudry, El{-}Yaniv, and
  Bengio]{DBLP:journals/jmlr/HubaraCSEB17}
Itay Hubara, Matthieu Courbariaux, Daniel Soudry, Ran El{-}Yaniv, and Yoshua
  Bengio.
\newblock Quantized neural networks: Training neural networks with low
  precision weights and activations.
\newblock \emph{J. Mach. Learn. Res.}, 18:\penalty0 187:1--187:30, 2017.

\bibitem[Iandola et~al.(2014)Iandola, Moskewicz, Karayev, Girshick, Darrell,
  and Keutzer]{DBLP:journals/corr/IandolaMKGDK14}
Forrest~N. Iandola, Matthew~W. Moskewicz, Sergey Karayev, Ross~B. Girshick,
  Trevor Darrell, and Kurt Keutzer.
\newblock Densenet: Implementing efficient convnet descriptor pyramids.
\newblock \emph{CoRR}, abs/1404.1869, 2014.

\bibitem[Iandola et~al.(2016)Iandola, Moskewicz, Ashraf, Han, Dally, and
  Keutzer]{DBLP:journals/corr/IandolaMAHDK16}
Forrest~N. Iandola, Matthew~W. Moskewicz, Khalid Ashraf, Song Han, William~J.
  Dally, and Kurt Keutzer.
\newblock Squeezenet: Alexnet-level accuracy with 50x fewer parameters and
  {\textless}1mb model size.
\newblock \emph{CoRR}, abs/1602.07360, 2016.

\bibitem[Iglovikov \& Shvets(2018)Iglovikov and
  Shvets]{DBLP:journals/corr/abs-1801-05746}
Vladimir Iglovikov and Alexey Shvets.
\newblock Ternausnet: U-net with {VGG11} encoder pre-trained on imagenet for
  image segmentation.
\newblock \emph{CoRR}, abs/1801.05746, 2018.

\bibitem[Ioffe \& Szegedy(2015)Ioffe and Szegedy]{DBLP:conf/icml/IoffeS15}
Sergey Ioffe and Christian Szegedy.
\newblock Batch normalization: Accelerating deep network training by reducing
  internal covariate shift.
\newblock In \emph{{ICML}}, volume~37 of \emph{{JMLR} Workshop and Conference
  Proceedings}, pp.\  448--456. JMLR.org, 2015.

\bibitem[Jacob et~al.(2018)Jacob, Kligys, Chen, Zhu, Tang, Howard, Adam, and
  Kalenichenko]{DBLP:conf/cvpr/JacobKCZTHAK18}
Benoit Jacob, Skirmantas Kligys, Bo~Chen, Menglong Zhu, Matthew Tang, Andrew~G.
  Howard, Hartwig Adam, and Dmitry Kalenichenko.
\newblock Quantization and training of neural networks for efficient
  integer-arithmetic-only inference.
\newblock In \emph{2018 {IEEE} Conference on Computer Vision and Pattern
  Recognition, {CVPR} 2018, Salt Lake City, UT, USA, June 18-22, 2018}, pp.\
  2704--2713. {IEEE} Computer Society, 2018.

\bibitem[Jain et~al.(2019)Jain, Mo, Jain, Tumanov, Gonzalez, and
  Stoica]{DBLP:journals/corr/abs-1901-10008}
Paras Jain, Xiangxi Mo, Ajay Jain, Alexey Tumanov, Joseph~E. Gonzalez, and Ion
  Stoica.
\newblock The ooo {VLIW} {JIT} compiler for {GPU} inference.
\newblock \emph{CoRR}, abs/1901.10008, 2019.

\bibitem[Jean et~al.(2015)Jean, Cho, Memisevic, and
  Bengio]{DBLP:conf/acl/JeanCMB15}
S{\'{e}}bastien Jean, KyungHyun Cho, Roland Memisevic, and Yoshua Bengio.
\newblock On using very large target vocabulary for neural machine translation.
\newblock In \emph{{ACL} {(1)}}, pp.\  1--10. The Association for Computer
  Linguistics, 2015.

\bibitem[Jernite et~al.(2017)Jernite, Grave, Joulin, and
  Mikolov]{DBLP:conf/iclr/JerniteGJM17}
Yacine Jernite, Edouard Grave, Armand Joulin, and Tom{\'{a}}s Mikolov.
\newblock Variable computation in recurrent neural networks.
\newblock In \emph{5th International Conference on Learning Representations,
  {ICLR} 2017, Toulon, France, April 24-26, 2017, Conference Track
  Proceedings}. OpenReview.net, 2017.

\bibitem[Jiang et~al.(2019)Jiang, Hu, Xiao, Zhang, and
  Zhu]{DBLP:conf/emnlp/JiangHXZZ19}
Yufan Jiang, Chi Hu, Tong Xiao, Chunliang Zhang, and Jingbo Zhu.
\newblock Improved differentiable architecture search for language modeling and
  named entity recognition.
\newblock In \emph{Proceedings of the 2019 Conference on Empirical Methods in
  Natural Language Processing and the 9th International Joint Conference on
  Natural Language Processing, {EMNLP-IJCNLP} 2019, Hong Kong, China, November
  3-7, 2019}, pp.\  3583--3588, 2019.

\bibitem[Jin et~al.(2015)Jin, Dundar, and
  Culurciello]{DBLP:journals/corr/JinDC14}
Jonghoon Jin, Aysegul Dundar, and Eugenio Culurciello.
\newblock Flattened convolutional neural networks for feedforward acceleration.
\newblock In \emph{{ICLR} (Workshop)}, 2015.

\bibitem[Jin et~al.(2019)Jin, Peng, Wu, Liu, Liu, Liang, Yan, and
  Hu]{DBLP:conf/iccv/JinPWLLLYH19}
Xiao Jin, Baoyun Peng, Yichao Wu, Yu~Liu, Jiaheng Liu, Ding Liang, Junjie Yan,
  and Xiaolin Hu.
\newblock Knowledge distillation via route constrained optimization.
\newblock In \emph{2019 {IEEE/CVF} International Conference on Computer Vision,
  {ICCV} 2019, Seoul, Korea (South), October 27 - November 2, 2019}, pp.\
  1345--1354. {IEEE}, 2019.

\bibitem[Johnson et~al.(2017)Johnson, Schuster, Le, Krikun, Wu, Chen, Thorat,
  Vi{\'{e}}gas, Wattenberg, Corrado, Hughes, and
  Dean]{DBLP:journals/tacl/JohnsonSLKWCTVW17}
Melvin Johnson, Mike Schuster, Quoc~V. Le, Maxim Krikun, Yonghui Wu, Zhifeng
  Chen, Nikhil Thorat, Fernanda~B. Vi{\'{e}}gas, Martin Wattenberg, Greg
  Corrado, Macduff Hughes, and Jeffrey Dean.
\newblock Google's multilingual neural machine translation system: Enabling
  zero-shot translation.
\newblock \emph{Trans. Assoc. Comput. Linguistics}, 5:\penalty0 339--351, 2017.

\bibitem[Joshi et~al.(2016)Joshi, Tripathi, Patel, Bhattacharyya, and
  Carman]{DBLP:conf/emnlp/JoshiTPBC16}
Aditya Joshi, Vaibhav Tripathi, Kevin Patel, Pushpak Bhattacharyya, and
  Mark~James Carman.
\newblock Are word embedding-based features useful for sarcasm detection?
\newblock In \emph{{EMNLP}}, pp.\  1006--1011. The Association for
  Computational Linguistics, 2016.

\bibitem[Joshi et~al.(2009)Joshi, Porikli, and
  Papanikolopoulos]{DBLP:conf/cvpr/JoshiPP09}
Ajay~J. Joshi, Fatih Porikli, and Nikolaos Papanikolopoulos.
\newblock Multi-class active learning for image classification.
\newblock In \emph{{CVPR}}, pp.\  2372--2379. {IEEE} Computer Society, 2009.

\bibitem[Joshi et~al.(2019)Joshi, Choi, Levy, Weld, and
  Zettlemoyer]{DBLP:conf/naacl/JoshiCLWZ19}
Mandar Joshi, Eunsol Choi, Omer Levy, Daniel~S. Weld, and Luke Zettlemoyer.
\newblock pair2vec: Compositional word-pair embeddings for cross-sentence
  inference.
\newblock In \emph{{NAACL-HLT} {(1)}}, pp.\  3597--3608. Association for
  Computational Linguistics, 2019.

\bibitem[Joshi et~al.(2020)Joshi, Chen, Liu, Weld, Zettlemoyer, and
  Levy]{DBLP:journals/tacl/JoshiCLWZL20}
Mandar Joshi, Danqi Chen, Yinhan Liu, Daniel~S. Weld, Luke Zettlemoyer, and
  Omer Levy.
\newblock Spanbert: Improving pre-training by representing and predicting
  spans.
\newblock \emph{Trans. Assoc. Comput. Linguistics}, 8:\penalty0 64--77, 2020.

\bibitem[J{\'{o}}zefowicz et~al.(2016)J{\'{o}}zefowicz, Vinyals, Schuster,
  Shazeer, and Wu]{DBLP:journals/corr/JozefowiczVSSW16}
Rafal J{\'{o}}zefowicz, Oriol Vinyals, Mike Schuster, Noam Shazeer, and Yonghui
  Wu.
\newblock Exploring the limits of language modeling.
\newblock \emph{CoRR}, abs/1602.02410, 2016.

\bibitem[Kaya et~al.(2019)Kaya, Hong, and Dumitras]{DBLP:conf/icml/KayaHD19}
Yigitcan Kaya, Sanghyun Hong, and Tudor Dumitras.
\newblock Shallow-deep networks: Understanding and mitigating network
  overthinking.
\newblock In Kamalika Chaudhuri and Ruslan Salakhutdinov (eds.),
  \emph{Proceedings of the 36th International Conference on Machine Learning,
  {ICML} 2019, 9-15 June 2019, Long Beach, California, {USA}}, volume~97 of
  \emph{Proceedings of Machine Learning Research}, pp.\  3301--3310. {PMLR},
  2019.

\bibitem[Kim \& Cho(2021)Kim and Cho]{DBLP:conf/acl/KimC20}
Gyuwan Kim and Kyunghyun Cho.
\newblock Length-adaptive transformer: Train once with length drop, use anytime
  with search.
\newblock In Chengqing Zong, Fei Xia, Wenjie Li, and Roberto Navigli (eds.),
  \emph{Proceedings of the 59th Annual Meeting of the Association for
  Computational Linguistics and the 11th International Joint Conference on
  Natural Language Processing, {ACL/IJCNLP} 2021, (Volume 1: Long Papers),
  Virtual Event, August 1-6, 2021}, pp.\  6501--6511. Association for
  Computational Linguistics, 2021.

\bibitem[Kim et~al.(2018)Kim, Park, and Kwak]{DBLP:conf/nips/KimPK18}
Jangho Kim, Seonguk Park, and Nojun Kwak.
\newblock Paraphrasing complex network: Network compression via factor
  transfer.
\newblock In Samy Bengio, Hanna~M. Wallach, Hugo Larochelle, Kristen Grauman,
  Nicol{\`{o}} Cesa{-}Bianchi, and Roman Garnett (eds.), \emph{Advances in
  Neural Information Processing Systems 31: Annual Conference on Neural
  Information Processing Systems 2018, NeurIPS 2018, December 3-8, 2018,
  Montr{\'{e}}al, Canada}, pp.\  2765--2774, 2018.

\bibitem[Kim et~al.(2020)Kim, Hyun, Chung, and Kwak]{DBLP:conf/icpr/KimHCK20}
Jangho Kim, Minsung Hyun, Inseop Chung, and Nojun Kwak.
\newblock Feature fusion for online mutual knowledge distillation.
\newblock In \emph{25th International Conference on Pattern Recognition, {ICPR}
  2020, Virtual Event / Milan, Italy, January 10-15, 2021}, pp.\  4619--4625.
  {IEEE}, 2020.

\bibitem[Kim et~al.(2021)Kim, Gholami, Yao, Mahoney, and
  Keutzer]{DBLP:journals/corr/abs-2101-01321}
Sehoon Kim, Amir Gholami, Zhewei Yao, Michael~W. Mahoney, and Kurt Keutzer.
\newblock {I-BERT:} integer-only {BERT} quantization.
\newblock \emph{CoRR}, abs/2101.01321, 2021.

\bibitem[Kim et~al.(2016{\natexlab{a}})Kim, Park, Yoo, Choi, Yang, and
  Shin]{DBLP:journals/corr/KimPYCYS15}
Yong{-}Deok Kim, Eunhyeok Park, Sungjoo Yoo, Taelim Choi, Lu~Yang, and Dongjun
  Shin.
\newblock Compression of deep convolutional neural networks for fast and low
  power mobile applications.
\newblock In \emph{{ICLR} (Poster)}, 2016{\natexlab{a}}.

\bibitem[Kim et~al.(2016{\natexlab{b}})Kim, Jernite, Sontag, and
  Rush]{DBLP:conf/aaai/KimJSR16}
Yoon Kim, Yacine Jernite, David~A. Sontag, and Alexander~M. Rush.
\newblock Character-aware neural language models.
\newblock In \emph{{AAAI}}, pp.\  2741--2749. {AAAI} Press, 2016{\natexlab{b}}.

\bibitem[Kitaev et~al.(2020)Kitaev, Kaiser, and
  Levskaya]{DBLP:conf/iclr/KitaevKL20}
Nikita Kitaev, Lukasz Kaiser, and Anselm Levskaya.
\newblock Reformer: The efficient transformer.
\newblock In \emph{{ICLR}}. OpenReview.net, 2020.

\bibitem[Kolda \& Bader(2009)Kolda and Bader]{DBLP:journals/siamrev/KoldaB09}
Tamara~G. Kolda and Brett~W. Bader.
\newblock Tensor decompositions and applications.
\newblock \emph{{SIAM} Review}, 51\penalty0 (3):\penalty0 455--500, 2009.

\bibitem[Krizhevsky et~al.(2012)Krizhevsky, Sutskever, and
  Hinton]{DBLP:conf/nips/KrizhevskySH12}
Alex Krizhevsky, Ilya Sutskever, and Geoffrey~E. Hinton.
\newblock Imagenet classification with deep convolutional neural networks.
\newblock In \emph{Advances in Neural Information Processing Systems 25: 26th
  Annual Conference on Neural Information Processing Systems 2012. Proceedings
  of a meeting held December 3-6, 2012, Lake Tahoe, Nevada, United States},
  pp.\  1106--1114, 2012.

\bibitem[Kulkarni \& Karande(2017)Kulkarni and
  Karande]{DBLP:journals/corr/KulkarniK17}
Mandar Kulkarni and Shirish~Subhash Karande.
\newblock Layer-wise training of deep networks using kernel similarity.
\newblock \emph{CoRR}, abs/1703.07115, 2017.

\bibitem[Kumar(2017)]{DBLP:journals/corr/Kumar17}
Siddharth~Krishna Kumar.
\newblock On weight initialization in deep neural networks.
\newblock \emph{CoRR}, abs/1704.08863, 2017.

\bibitem[Lacoste et~al.(2019)Lacoste, Luccioni, Schmidt, and
  Dandres]{DBLP:journals/corr/abs-1910-09700}
Alexandre Lacoste, Alexandra Luccioni, Victor Schmidt, and Thomas Dandres.
\newblock Quantifying the carbon emissions of machine learning.
\newblock \emph{CoRR}, abs/1910.09700, 2019.

\bibitem[Lam(2018)]{DBLP:journals/corr/abs-1803-05651}
Maximilian Lam.
\newblock Word2bits - quantized word vectors.
\newblock \emph{CoRR}, abs/1803.05651, 2018.

\bibitem[Lan et~al.(2018)Lan, Zhu, and Gong]{DBLP:conf/nips/LanZG18}
Xu~Lan, Xiatian Zhu, and Shaogang Gong.
\newblock Knowledge distillation by on-the-fly native ensemble.
\newblock In Samy Bengio, Hanna~M. Wallach, Hugo Larochelle, Kristen Grauman,
  Nicol{\`{o}} Cesa{-}Bianchi, and Roman Garnett (eds.), \emph{Advances in
  Neural Information Processing Systems 31: Annual Conference on Neural
  Information Processing Systems 2018, NeurIPS 2018, December 3-8, 2018,
  Montr{\'{e}}al, Canada}, pp.\  7528--7538, 2018.

\bibitem[Lan et~al.(2020)Lan, Chen, Goodman, Gimpel, Sharma, and
  Soricut]{DBLP:conf/iclr/LanCGGSS20}
Zhenzhong Lan, Mingda Chen, Sebastian Goodman, Kevin Gimpel, Piyush Sharma, and
  Radu Soricut.
\newblock {ALBERT:} {A} lite {BERT} for self-supervised learning of language
  representations.
\newblock In \emph{{ICLR}}. OpenReview.net, 2020.

\bibitem[Lebedev et~al.(2015)Lebedev, Ganin, Rakhuba, Oseledets, and
  Lempitsky]{DBLP:journals/corr/LebedevGROL14}
Vadim Lebedev, Yaroslav Ganin, Maksim Rakhuba, Ivan~V. Oseledets, and Victor~S.
  Lempitsky.
\newblock Speeding-up convolutional neural networks using fine-tuned
  cp-decomposition.
\newblock In \emph{{ICLR} (Poster)}, 2015.

\bibitem[LeCun et~al.(1989)LeCun, Denker, and Solla]{DBLP:conf/nips/CunDS89}
Yann LeCun, John~S. Denker, and Sara~A. Solla.
\newblock Optimal brain damage.
\newblock In \emph{{NIPS}}, pp.\  598--605. Morgan Kaufmann, 1989.

\bibitem[LeCun et~al.(1998)LeCun, Bottou, Bengio, and
  Haffner]{lecun1998gradient}
Yann LeCun, L{\'e}on Bottou, Yoshua Bengio, and Patrick Haffner.
\newblock Gradient-based learning applied to document recognition.
\newblock \emph{Proceedings of the IEEE}, 86\penalty0 (11):\penalty0
  2278--2324, 1998.

\bibitem[Lee et~al.(2019)Lee, Ajanthan, and Torr]{DBLP:conf/iclr/LeeAT19}
Namhoon Lee, Thalaiyasingam Ajanthan, and Philip H.~S. Torr.
\newblock Snip: single-shot network pruning based on connection sensitivity.
\newblock In \emph{7th International Conference on Learning Representations,
  {ICLR} 2019, New Orleans, LA, USA, May 6-9, 2019}, 2019.

\bibitem[Lee et~al.(2018)Lee, Kim, and Song]{DBLP:conf/eccv/LeeKS18}
Seung~Hyun Lee, Dae~Ha Kim, and Byung~Cheol Song.
\newblock Self-supervised knowledge distillation using singular value
  decomposition.
\newblock In Vittorio Ferrari, Martial Hebert, Cristian Sminchisescu, and Yair
  Weiss (eds.), \emph{Computer Vision - {ECCV} 2018 - 15th European Conference,
  Munich, Germany, September 8-14, 2018, Proceedings, Part {VI}}, volume 11210
  of \emph{Lecture Notes in Computer Science}, pp.\  339--354. Springer, 2018.

\bibitem[Lee \& Song(2019)Lee and Song]{DBLP:conf/bmvc/LeeS19}
Seunghyun Lee and Byung~Cheol Song.
\newblock Graph-based knowledge distillation by multi-head attention network.
\newblock In \emph{30th British Machine Vision Conference 2019, {BMVC} 2019,
  Cardiff, UK, September 9-12, 2019}, pp.\  141. {BMVA} Press, 2019.

\bibitem[Lepikhin et~al.(2021)Lepikhin, Lee, Xu, Chen, Firat, Huang, Krikun,
  Shazeer, and Chen]{DBLP:conf/iclr/LepikhinLXCFHKS21}
Dmitry Lepikhin, HyoukJoong Lee, Yuanzhong Xu, Dehao Chen, Orhan Firat, Yanping
  Huang, Maxim Krikun, Noam Shazeer, and Zhifeng Chen.
\newblock Gshard: Scaling giant models with conditional computation and
  automatic sharding.
\newblock In \emph{9th International Conference on Learning Representations,
  {ICLR} 2021, Virtual Event, Austria, May 3-7, 2021}. OpenReview.net, 2021.

\bibitem[Levesque et~al.(2012)Levesque, Davis, and
  Morgenstern]{DBLP:conf/kr/LevesqueDM12}
Hector~J. Levesque, Ernest Davis, and Leora Morgenstern.
\newblock The winograd schema challenge.
\newblock In Gerhard Brewka, Thomas Eiter, and Sheila~A. McIlraith (eds.),
  \emph{Principles of Knowledge Representation and Reasoning: Proceedings of
  the Thirteenth International Conference, {KR} 2012, Rome, Italy, June 10-14,
  2012}. {AAAI} Press, 2012.

\bibitem[Lewis \& Gale(1994)Lewis and Gale]{DBLP:conf/sigir/LewisG94}
David~D. Lewis and William~A. Gale.
\newblock A sequential algorithm for training text classifiers.
\newblock In \emph{{SIGIR}}, pp.\  3--12. ACM/Springer, 1994.

\bibitem[Lewis et~al.(2020)Lewis, Liu, Goyal, Ghazvininejad, Mohamed, Levy,
  Stoyanov, and Zettlemoyer]{DBLP:conf/acl/LewisLGGMLSZ20}
Mike Lewis, Yinhan Liu, Naman Goyal, Marjan Ghazvininejad, Abdelrahman Mohamed,
  Omer Levy, Veselin Stoyanov, and Luke Zettlemoyer.
\newblock {BART:} denoising sequence-to-sequence pre-training for natural
  language generation, translation, and comprehension.
\newblock In Dan Jurafsky, Joyce Chai, Natalie Schluter, and Joel~R. Tetreault
  (eds.), \emph{Proceedings of the 58th Annual Meeting of the Association for
  Computational Linguistics, {ACL} 2020, Online, July 5-10, 2020}, pp.\
  7871--7880. Association for Computational Linguistics, 2020.

\bibitem[Li et~al.(2020{\natexlab{a}})Li, Duan, Fang, Gong, and
  Jiang]{DBLP:conf/aaai/LiDFGJ20}
Gen Li, Nan Duan, Yuejian Fang, Ming Gong, and Daxin Jiang.
\newblock Unicoder-vl: {A} universal encoder for vision and language by
  cross-modal pre-training.
\newblock In \emph{{AAAI}}, pp.\  11336--11344. {AAAI} Press,
  2020{\natexlab{a}}.

\bibitem[Li et~al.(2017)Li, Kadav, Durdanovic, Samet, and
  Graf]{DBLP:conf/iclr/0022KDSG17}
Hao Li, Asim Kadav, Igor Durdanovic, Hanan Samet, and Hans~Peter Graf.
\newblock Pruning filters for efficient convnets.
\newblock In \emph{5th International Conference on Learning Representations,
  {ICLR} 2017, Toulon, France, April 24-26, 2017, Conference Track
  Proceedings}, 2017.

\bibitem[Li et~al.(2020{\natexlab{b}})Li, Lin, Ren, Chen, Ren, Li, Zhou, and
  Sun]{DBLP:journals/corr/abs-2012-14682}
Lei Li, Yankai Lin, Shuhuai Ren, Deli Chen, Xuancheng Ren, Peng Li, Jie Zhou,
  and Xu~Sun.
\newblock Accelerating pre-trained language models via calibrated cascade.
\newblock \emph{CoRR}, abs/2012.14682, 2020{\natexlab{b}}.

\bibitem[Li \& Ji(2014)Li and Ji]{DBLP:conf/acl/LiJ14}
Qi~Li and Heng Ji.
\newblock Incremental joint extraction of entity mentions and relations.
\newblock In \emph{Proceedings of the 52nd Annual Meeting of the Association
  for Computational Linguistics, {ACL} 2014, June 22-27, 2014, Baltimore, MD,
  USA, Volume 1: Long Papers}, pp.\  402--412. The Association for Computer
  Linguistics, 2014.

\bibitem[Li \& Liang(2021)Li and Liang]{DBLP:conf/acl/LiL20}
Xiang~Lisa Li and Percy Liang.
\newblock Prefix-tuning: Optimizing continuous prompts for generation.
\newblock In Chengqing Zong, Fei Xia, Wenjie Li, and Roberto Navigli (eds.),
  \emph{Proceedings of the 59th Annual Meeting of the Association for
  Computational Linguistics and the 11th International Joint Conference on
  Natural Language Processing, {ACL/IJCNLP} 2021, (Volume 1: Long Papers),
  Virtual Event, August 1-6, 2021}, pp.\  4582--4597. Association for
  Computational Linguistics, 2021.

\bibitem[Li et~al.(2021)Li, Shao, Sun, Yan, Qiu, and
  Huang]{DBLP:conf/acl/LiSSYQH20}
Xiaonan Li, Yunfan Shao, Tianxiang Sun, Hang Yan, Xipeng Qiu, and Xuanjing
  Huang.
\newblock Accelerating {BERT} inference for sequence labeling via early-exit.
\newblock In Chengqing Zong, Fei Xia, Wenjie Li, and Roberto Navigli (eds.),
  \emph{Proceedings of the 59th Annual Meeting of the Association for
  Computational Linguistics and the 11th International Joint Conference on
  Natural Language Processing, {ACL/IJCNLP} 2021, (Volume 1: Long Papers),
  Virtual Event, August 1-6, 2021}, pp.\  189--199. Association for
  Computational Linguistics, 2021.

\bibitem[Liang et~al.(2021)Liang, Glossner, Wang, and
  Shi]{DBLP:journals/corr/abs-2101-09671}
Tailin Liang, John Glossner, Lei Wang, and Shaobo Shi.
\newblock Pruning and quantization for deep neural network acceleration: {A}
  survey.
\newblock \emph{CoRR}, abs/2101.09671, 2021.

\bibitem[Liao et~al.(2021)Liao, Zhang, Ren, Su, Sun, and
  He]{DBLP:conf/naacl/LiaoZRSSH21}
Kaiyuan Liao, Yi~Zhang, Xuancheng Ren, Qi~Su, Xu~Sun, and Bin He.
\newblock A global past-future early exit method for accelerating inference of
  pre-trained language models.
\newblock In Kristina Toutanova, Anna Rumshisky, Luke Zettlemoyer, Dilek
  Hakkani{-}T{\"{u}}r, Iz~Beltagy, Steven Bethard, Ryan Cotterell, Tanmoy
  Chakraborty, and Yichao Zhou (eds.), \emph{Proceedings of the 2021 Conference
  of the North American Chapter of the Association for Computational
  Linguistics: Human Language Technologies, {NAACL-HLT} 2021, Online, June
  6-11, 2021}, pp.\  2013--2023. Association for Computational Linguistics,
  2021.

\bibitem[Lienhart \& Maydt(2002)Lienhart and Maydt]{lienhart2002extended}
Rainer Lienhart and Jochen Maydt.
\newblock An extended set of haar-like features for rapid object detection.
\newblock In \emph{Proceedings. international conference on image processing},
  volume~1, pp.\  I--I. IEEE, 2002.

\bibitem[Lin et~al.(2017)Lin, Rao, Lu, and Zhou]{DBLP:conf/nips/LinRLZ17}
Ji~Lin, Yongming Rao, Jiwen Lu, and Jie Zhou.
\newblock Runtime neural pruning.
\newblock In \emph{Advances in Neural Information Processing Systems 30: Annual
  Conference on Neural Information Processing Systems 2017, December 4-9, 2017,
  Long Beach, CA, {USA}}, pp.\  2181--2191, 2017.

\bibitem[Lin et~al.(2021)Lin, Men, Yang, Zhou, Ding, Zhang, Wang, Wang, Jiang,
  Jia, Zhang, Zhang, Zou, Li, Deng, Liu, Xue, Zhou, Ma, Yu, Li, Lin, Zhou,
  Tang, and Yang]{DBLP:journals/corr/abs-2103-00823}
Junyang Lin, Rui Men, An~Yang, Chang Zhou, Ming Ding, Yichang Zhang, Peng Wang,
  Ang Wang, Le~Jiang, Xianyan Jia, Jie Zhang, Jianwei Zhang, Xu~Zou, Zhikang
  Li, Xiaodong Deng, Jie Liu, Jinbao Xue, Huiling Zhou, Jianxin Ma, Jin Yu,
  Yong Li, Wei Lin, Jingren Zhou, Jie Tang, and Hongxia Yang.
\newblock {M6:} {A} chinese multimodal pretrainer.
\newblock \emph{CoRR}, abs/2103.00823, 2021.

\bibitem[Lin et~al.(2014)Lin, Maire, Belongie, Hays, Perona, Ramanan,
  Doll{\'{a}}r, and Zitnick]{DBLP:conf/eccv/LinMBHPRDZ14}
Tsung{-}Yi Lin, Michael Maire, Serge~J. Belongie, James Hays, Pietro Perona,
  Deva Ramanan, Piotr Doll{\'{a}}r, and C.~Lawrence Zitnick.
\newblock Microsoft {COCO:} common objects in context.
\newblock In David~J. Fleet, Tom{\'{a}}s Pajdla, Bernt Schiele, and Tinne
  Tuytelaars (eds.), \emph{Computer Vision - {ECCV} 2014 - 13th European
  Conference, Zurich, Switzerland, September 6-12, 2014, Proceedings, Part
  {V}}, volume 8693 of \emph{Lecture Notes in Computer Science}, pp.\
  740--755. Springer, 2014.

\bibitem[Lin et~al.(2018)Lin, Yang, Stoyanov, and
  Ji]{DBLP:conf/acl/StoyanovJLY18}
Ying Lin, Shengqi Yang, Veselin Stoyanov, and Heng Ji.
\newblock A multi-lingual multi-task architecture for low-resource sequence
  labeling.
\newblock In \emph{{ACL} {(1)}}, pp.\  799--809. Association for Computational
  Linguistics, 2018.

\bibitem[Lin et~al.(2020)Lin, Pan, Wang, Qiu, Feng, Zhou, and
  Li]{DBLP:conf/emnlp/LinPWQFZL20}
Zehui Lin, Xiao Pan, Mingxuan Wang, Xipeng Qiu, Jiangtao Feng, Hao Zhou, and
  Lei Li.
\newblock Pre-training multilingual neural machine translation by leveraging
  alignment information.
\newblock In \emph{{EMNLP} {(1)}}, pp.\  2649--2663. Association for
  Computational Linguistics, 2020.

\bibitem[Liu et~al.(2018{\natexlab{a}})Liu, Zoph, Neumann, Shlens, Hua, Li,
  Fei{-}Fei, Yuille, Huang, and Murphy]{DBLP:conf/eccv/LiuZNSHLFYHM18}
Chenxi Liu, Barret Zoph, Maxim Neumann, Jonathon Shlens, Wei Hua, Li{-}Jia Li,
  Li~Fei{-}Fei, Alan~L. Yuille, Jonathan Huang, and Kevin Murphy.
\newblock Progressive neural architecture search.
\newblock In \emph{{ECCV} {(1)}}, volume 11205 of \emph{Lecture Notes in
  Computer Science}, pp.\  19--35. Springer, 2018{\natexlab{a}}.

\bibitem[Liu et~al.(2019{\natexlab{a}})Liu, Simonyan, and
  Yang]{DBLP:conf/iclr/LiuSY19}
Hanxiao Liu, Karen Simonyan, and Yiming Yang.
\newblock {DARTS:} differentiable architecture search.
\newblock In \emph{7th International Conference on Learning Representations,
  {ICLR} 2019, New Orleans, LA, USA, May 6-9, 2019}, 2019{\natexlab{a}}.

\bibitem[Liu et~al.(2019{\natexlab{b}})Liu, Peng, and
  Schwing]{DBLP:conf/iclr/LiuPS19}
Iou{-}Jen Liu, Jian Peng, and Alexander~G. Schwing.
\newblock Knowledge flow: Improve upon your teachers.
\newblock In \emph{7th International Conference on Learning Representations,
  {ICLR} 2019, New Orleans, LA, USA, May 6-9, 2019}. OpenReview.net,
  2019{\natexlab{b}}.

\bibitem[Liu et~al.(2018{\natexlab{b}})Liu, Shang, Ren, Xu, Gui, Peng, and
  Han]{DBLP:conf/aaai/LiuSRXG0018}
Liyuan Liu, Jingbo Shang, Xiang Ren, Frank~Fangzheng Xu, Huan Gui, Jian Peng,
  and Jiawei Han.
\newblock Empower sequence labeling with task-aware neural language model.
\newblock In \emph{{AAAI}}, pp.\  5253--5260. {AAAI} Press, 2018{\natexlab{b}}.

\bibitem[Liu et~al.(2021{\natexlab{a}})Liu, Yuan, Fu, Jiang, Hayashi, and
  Neubig]{DBLP:journals/corr/abs-2107-13586}
Pengfei Liu, Weizhe Yuan, Jinlan Fu, Zhengbao Jiang, Hiroaki Hayashi, and
  Graham Neubig.
\newblock Pre-train, prompt, and predict: {A} systematic survey of prompting
  methods in natural language processing.
\newblock \emph{CoRR}, abs/2107.13586, 2021{\natexlab{a}}.

\bibitem[Liu et~al.(2020{\natexlab{a}})Liu, Zhou, Wang, Zhao, Deng, and
  Ju]{DBLP:conf/acl/LiuZWZDJ20}
Weijie Liu, Peng Zhou, Zhiruo Wang, Zhe Zhao, Haotang Deng, and Qi~Ju.
\newblock Fastbert: a self-distilling {BERT} with adaptive inference time.
\newblock In Dan Jurafsky, Joyce Chai, Natalie Schluter, and Joel~R. Tetreault
  (eds.), \emph{Proceedings of the 58th Annual Meeting of the Association for
  Computational Linguistics, {ACL} 2020, Online, July 5-10, 2020}, pp.\
  6035--6044. Association for Computational Linguistics, 2020{\natexlab{a}}.

\bibitem[Liu et~al.(2018{\natexlab{c}})Liu, Mou, Cui, Lu, and
  Song]{DBLP:journals/corr/abs-1807-02314}
Xianggen Liu, Lili Mou, Haotian Cui, Zhengdong Lu, and Sen Song.
\newblock {JUMPER:} learning when to make classification decisions in reading.
\newblock \emph{CoRR}, abs/1807.02314, 2018{\natexlab{c}}.

\bibitem[Liu et~al.(2021{\natexlab{b}})Liu, Ji, Fu, Du, Yang, and
  Tang]{liu2021ptuning}
Xiao Liu, Kaixuan Ji, Yicheng Fu, Zhengxiao Du, Zhilin Yang, and Jie Tang.
\newblock P-tuning v2: Prompt tuning can be comparable to fine-tuning
  universally across scales and tasks, 2021{\natexlab{b}}.

\bibitem[Liu et~al.(2021{\natexlab{c}})Liu, Zheng, Du, Ding, Qian, Yang, and
  Tang]{DBLP:journals/corr/abs-2103-10385}
Xiao Liu, Yanan Zheng, Zhengxiao Du, Ming Ding, Yujie Qian, Zhilin Yang, and
  Jie Tang.
\newblock {GPT} understands, too.
\newblock \emph{CoRR}, abs/2103.10385, 2021{\natexlab{c}}.

\bibitem[Liu et~al.(2020{\natexlab{b}})Liu, Gu, Goyal, Li, Edunov,
  Ghazvininejad, Lewis, and Zettlemoyer]{DBLP:journals/tacl/LiuGGLEGLZ20}
Yinhan Liu, Jiatao Gu, Naman Goyal, Xian Li, Sergey Edunov, Marjan
  Ghazvininejad, Mike Lewis, and Luke Zettlemoyer.
\newblock Multilingual denoising pre-training for neural machine translation.
\newblock \emph{Trans. Assoc. Comput. Linguistics}, 8:\penalty0 726--742,
  2020{\natexlab{b}}.

\bibitem[Liu et~al.(2019{\natexlab{c}})Liu, Sun, Zhou, Huang, and
  Darrell]{DBLP:conf/iclr/LiuSZHD19}
Zhuang Liu, Mingjie Sun, Tinghui Zhou, Gao Huang, and Trevor Darrell.
\newblock Rethinking the value of network pruning.
\newblock In \emph{{ICLR} (Poster)}. OpenReview.net, 2019{\natexlab{c}}.

\bibitem[Long et~al.(2016)Long, Zhu, Wang, and
  Jordan]{DBLP:conf/nips/LongZ0J16}
Mingsheng Long, Han Zhu, Jianmin Wang, and Michael~I. Jordan.
\newblock Unsupervised domain adaptation with residual transfer networks.
\newblock In \emph{{NIPS}}, pp.\  136--144, 2016.

\bibitem[Lu et~al.(2019)Lu, Batra, Parikh, and Lee]{DBLP:conf/nips/LuBPL19}
Jiasen Lu, Dhruv Batra, Devi Parikh, and Stefan Lee.
\newblock Vilbert: Pretraining task-agnostic visiolinguistic representations
  for vision-and-language tasks.
\newblock In \emph{NeurIPS}, pp.\  13--23, 2019.

\bibitem[Luo et~al.(2020)Luo, Yang, Li, Ren, and
  Sun]{DBLP:journals/corr/abs-2010-06351}
Fuli Luo, Pengcheng Yang, Shicheng Li, Xuancheng Ren, and Xu~Sun.
\newblock {CAPT:} contrastive pre-training for learning denoised sequence
  representations.
\newblock \emph{CoRR}, abs/2010.06351, 2020.

\bibitem[Luo et~al.(2019)Luo, Zhang, Zhou, Xie, Wu, and
  Lin]{DBLP:journals/pami/LuoZZXWL19}
Jian{-}Hao Luo, Hao Zhang, Hong{-}Yu Zhou, Chen{-}Wei Xie, Jianxin Wu, and
  Weiyao Lin.
\newblock Thinet: Pruning {CNN} filters for a thinner net.
\newblock \emph{{IEEE} Trans. Pattern Anal. Mach. Intell.}, 41\penalty0
  (10):\penalty0 2525--2538, 2019.

\bibitem[Luo et~al.(2018)Luo, Tian, Qin, Chen, and
  Liu]{DBLP:conf/nips/LuoTQCL18}
Renqian Luo, Fei Tian, Tao Qin, Enhong Chen, and Tie{-}Yan Liu.
\newblock Neural architecture optimization.
\newblock In \emph{NeurIPS}, pp.\  7827--7838, 2018.

\bibitem[Luong et~al.(2016)Luong, Le, Sutskever, Vinyals, and
  Kaiser]{DBLP:journals/corr/LuongLSVK15}
Minh{-}Thang Luong, Quoc~V. Le, Ilya Sutskever, Oriol Vinyals, and Lukasz
  Kaiser.
\newblock Multi-task sequence to sequence learning.
\newblock In \emph{{ICLR} (Poster)}, 2016.

\bibitem[Ma et~al.(2019)Ma, Zhang, Zhang, Duan, Hou, Zhou, and
  Song]{DBLP:conf/nips/MaZZDHZ019}
Xindian Ma, Peng Zhang, Shuai Zhang, Nan Duan, Yuexian Hou, Ming Zhou, and
  Dawei Song.
\newblock A tensorized transformer for language modeling.
\newblock In \emph{NeurIPS}, pp.\  2229--2239, 2019.

\bibitem[MacKay et~al.(2018)MacKay, Vicol, Ba, and
  Grosse]{DBLP:conf/nips/MacKayVBG18}
Matthew MacKay, Paul Vicol, Jimmy Ba, and Roger~B. Grosse.
\newblock Reversible recurrent neural networks.
\newblock In \emph{NeurIPS}, pp.\  9043--9054, 2018.

\bibitem[Martins \& Astudillo(2016)Martins and
  Astudillo]{DBLP:conf/icml/MartinsA16}
Andr{\'{e}} F.~T. Martins and Ram{\'{o}}n~Fernandez Astudillo.
\newblock From softmax to sparsemax: {A} sparse model of attention and
  multi-label classification.
\newblock In \emph{{ICML}}, volume~48 of \emph{{JMLR} Workshop and Conference
  Proceedings}, pp.\  1614--1623. JMLR.org, 2016.

\bibitem[Maruf et~al.(2019)Maruf, Martins, and
  Haffari]{DBLP:conf/naacl/MarufMH19}
Sameen Maruf, Andr{\'{e}} F.~T. Martins, and Gholamreza Haffari.
\newblock Selective attention for context-aware neural machine translation.
\newblock In \emph{{NAACL-HLT} {(1)}}, pp.\  3092--3102. Association for
  Computational Linguistics, 2019.

\bibitem[McCarley et~al.(2019)McCarley, Chakravarti, and
  Sil]{mccarley2019structured}
JS~McCarley, Rishav Chakravarti, and Avirup Sil.
\newblock Structured pruning of a bert-based question answering model.
\newblock \emph{arXiv preprint arXiv:1910.06360}, 2019.

\bibitem[McDonald et~al.(2010)McDonald, Hall, and
  Mann]{DBLP:conf/naacl/McDonaldHM10}
Ryan~T. McDonald, Keith~B. Hall, and Gideon Mann.
\newblock Distributed training strategies for the structured perceptron.
\newblock In \emph{Human Language Technologies: Conference of the North
  American Chapter of the Association of Computational Linguistics,
  Proceedings, June 2-4, 2010, Los Angeles, California, {USA}}, pp.\  456--464.
  The Association for Computational Linguistics, 2010.

\bibitem[Mellor et~al.(2020)Mellor, Turner, Storkey, and
  Crowley]{DBLP:journals/corr/abs-2006-04647}
Joseph Mellor, Jack Turner, Amos~J. Storkey, and Elliot~J. Crowley.
\newblock Neural architecture search without training.
\newblock \emph{CoRR}, abs/2006.04647, 2020.

\bibitem[Michel et~al.(2019)Michel, Levy, and
  Neubig]{DBLP:conf/nips/MichelLN19}
Paul Michel, Omer Levy, and Graham Neubig.
\newblock Are sixteen heads really better than one?
\newblock In \emph{Advances in Neural Information Processing Systems 32: Annual
  Conference on Neural Information Processing Systems 2019, NeurIPS 2019,
  December 8-14, 2019, Vancouver, BC, Canada}, pp.\  14014--14024, 2019.

\bibitem[Mikolov et~al.(2013{\natexlab{a}})Mikolov, Chen, Corrado, and
  Dean]{DBLP:journals/corr/abs-1301-3781}
Tom{\'{a}}s Mikolov, Kai Chen, Greg Corrado, and Jeffrey Dean.
\newblock Efficient estimation of word representations in vector space.
\newblock In \emph{{ICLR} (Workshop Poster)}, 2013{\natexlab{a}}.

\bibitem[Mikolov et~al.(2013{\natexlab{b}})Mikolov, Sutskever, Chen, Corrado,
  and Dean]{DBLP:conf/nips/MikolovSCCD13}
Tom{\'{a}}s Mikolov, Ilya Sutskever, Kai Chen, Gregory~S. Corrado, and Jeffrey
  Dean.
\newblock Distributed representations of words and phrases and their
  compositionality.
\newblock In \emph{{NIPS}}, pp.\  3111--3119, 2013{\natexlab{b}}.

\bibitem[Mirzadeh et~al.(2020)Mirzadeh, Farajtabar, Li, Levine, Matsukawa, and
  Ghasemzadeh]{DBLP:conf/aaai/MirzadehFLLMG20}
Seyed{-}Iman Mirzadeh, Mehrdad Farajtabar, Ang Li, Nir Levine, Akihiro
  Matsukawa, and Hassan Ghasemzadeh.
\newblock Improved knowledge distillation via teacher assistant.
\newblock In \emph{The Thirty-Fourth {AAAI} Conference on Artificial
  Intelligence, {AAAI} 2020, The Thirty-Second Innovative Applications of
  Artificial Intelligence Conference, {IAAI} 2020, The Tenth {AAAI} Symposium
  on Educational Advances in Artificial Intelligence, {EAAI} 2020, New York,
  NY, USA, February 7-12, 2020}, pp.\  5191--5198. {AAAI} Press, 2020.

\bibitem[Mishkin \& Matas(2016)Mishkin and
  Matas]{DBLP:journals/corr/MishkinM15}
Dmytro Mishkin and Jiri Matas.
\newblock All you need is a good init.
\newblock In \emph{{ICLR} (Poster)}, 2016.

\bibitem[Mnih \& Hinton(2008)Mnih and Hinton]{DBLP:conf/nips/MnihH08}
Andriy Mnih and Geoffrey~E. Hinton.
\newblock A scalable hierarchical distributed language model.
\newblock In \emph{{NIPS}}, pp.\  1081--1088. Curran Associates, Inc., 2008.

\bibitem[Molchanov et~al.(2017)Molchanov, Tyree, Karras, Aila, and
  Kautz]{DBLP:conf/iclr/MolchanovTKAK17}
Pavlo Molchanov, Stephen Tyree, Tero Karras, Timo Aila, and Jan Kautz.
\newblock Pruning convolutional neural networks for resource efficient
  inference.
\newblock In \emph{5th International Conference on Learning Representations,
  {ICLR} 2017, Toulon, France, April 24-26, 2017, Conference Track
  Proceedings}, 2017.

\bibitem[Morin \& Bengio(2005)Morin and Bengio]{DBLP:conf/aistats/MorinB05}
Frederic Morin and Yoshua Bengio.
\newblock Hierarchical probabilistic neural network language model.
\newblock In \emph{{AISTATS}}. Society for Artificial Intelligence and
  Statistics, 2005.

\bibitem[Mou et~al.(2016)Mou, Jia, Xu, Li, Zhang, and
  Jin]{DBLP:conf/cikm/MouJXL0J16}
Lili Mou, Ran Jia, Yan Xu, Ge~Li, Lu~Zhang, and Zhi Jin.
\newblock Distilling word embeddings: An encoding approach.
\newblock In \emph{{CIKM}}, pp.\  1977--1980. {ACM}, 2016.

\bibitem[Narang et~al.(2017{\natexlab{a}})Narang, Diamos, Sengupta, and
  Elsen]{DBLP:conf/iclr/NarangDSE17}
Sharan Narang, Greg Diamos, Shubho Sengupta, and Erich Elsen.
\newblock Exploring sparsity in recurrent neural networks.
\newblock In \emph{5th International Conference on Learning Representations,
  {ICLR} 2017, Toulon, France, April 24-26, 2017, Conference Track
  Proceedings}, 2017{\natexlab{a}}.

\bibitem[Narang et~al.(2017{\natexlab{b}})Narang, Undersander, and
  Diamos]{DBLP:journals/corr/abs-1711-02782}
Sharan Narang, Eric Undersander, and Gregory~F. Diamos.
\newblock Block-sparse recurrent neural networks.
\newblock \emph{CoRR}, abs/1711.02782, 2017{\natexlab{b}}.

\bibitem[Nguyen \& Smeulders(2004)Nguyen and
  Smeulders]{DBLP:conf/icml/NguyenS04}
Hieu~Tat Nguyen and Arnold W.~M. Smeulders.
\newblock Active learning using pre-clustering.
\newblock In \emph{{ICML}}, volume~69 of \emph{{ACM} International Conference
  Proceeding Series}. {ACM}, 2004.

\bibitem[Niculae \& Blondel(2017)Niculae and
  Blondel]{DBLP:conf/nips/NiculaeB17}
Vlad Niculae and Mathieu Blondel.
\newblock A regularized framework for sparse and structured neural attention.
\newblock In \emph{{NIPS}}, pp.\  3338--3348, 2017.

\bibitem[Noach \& Goldberg(2020)Noach and Goldberg]{DBLP:conf/ijcnlp/NoachG20}
Matan~Ben Noach and Yoav Goldberg.
\newblock Compressing pre-trained language models by matrix decomposition.
\newblock In Kam{-}Fai Wong, Kevin Knight, and Hua Wu (eds.), \emph{Proceedings
  of the 1st Conference of the Asia-Pacific Chapter of the Association for
  Computational Linguistics and the 10th International Joint Conference on
  Natural Language Processing, {AACL/IJCNLP} 2020, Suzhou, China, December 4-7,
  2020}, pp.\  884--889. Association for Computational Linguistics, 2020.

\bibitem[Oquab et~al.(2014)Oquab, Bottou, Laptev, and
  Sivic]{DBLP:conf/cvpr/OquabBLS14}
Maxime Oquab, L{\'{e}}on Bottou, Ivan Laptev, and Josef Sivic.
\newblock Learning and transferring mid-level image representations using
  convolutional neural networks.
\newblock In \emph{{CVPR}}, pp.\  1717--1724. {IEEE} Computer Society, 2014.

\bibitem[Ott et~al.(2016)Ott, Lin, Zhang, Liu, and
  Bengio]{DBLP:journals/corr/OttLZLB16a}
Joachim Ott, Zhouhan Lin, Ying Zhang, Shih{-}Chii Liu, and Yoshua Bengio.
\newblock Recurrent neural networks with limited numerical precision.
\newblock \emph{CoRR}, abs/1611.07065, 2016.

\bibitem[Park et~al.(2021)Park, Cha, Jeong, Kim, and
  Han]{DBLP:journals/corr/abs-2102-07650}
Dae~Young Park, Moon{-}Hyun Cha, Changwook Jeong, Daesin Kim, and Bohyung Han.
\newblock Learning student-friendly teacher networks for knowledge
  distillation.
\newblock \emph{CoRR}, abs/2102.07650, 2021.

\bibitem[Park et~al.(2015)Park, Kim, Kim, Kim, Kim, Yoon, and
  Yoo]{DBLP:conf/codes/ParkKKKKYY15}
Eunhyeok Park, Dongyoung Kim, Soobeom Kim, Yong{-}Deok Kim, Gunhee Kim, Sungroh
  Yoon, and Sungjoo Yoo.
\newblock Big/little deep neural network for ultra low power inference.
\newblock In Gabriela Nicolescu and Andreas Gerstlauer (eds.), \emph{2015
  International Conference on Hardware/Software Codesign and System Synthesis,
  {CODES+ISSS} 2015, Amsterdam, Netherlands, October 4-9, 2015}, pp.\
  124--132. {IEEE}, 2015.

\bibitem[Park \& Kwak(2019)Park and Kwak]{DBLP:journals/corr/abs-1909-10754}
Seonguk Park and Nojun Kwak.
\newblock {FEED:} feature-level ensemble for knowledge distillation.
\newblock \emph{CoRR}, abs/1909.10754, 2019.

\bibitem[Peng et~al.(2019)Peng, Jin, Li, Zhou, Wu, Liu, Zhang, and
  Liu]{DBLP:conf/iccv/PengJLZWLZ019}
Baoyun Peng, Xiao Jin, Dongsheng Li, Shunfeng Zhou, Yichao Wu, Jiaheng Liu,
  Zhaoning Zhang, and Yu~Liu.
\newblock Correlation congruence for knowledge distillation.
\newblock In \emph{2019 {IEEE/CVF} International Conference on Computer Vision,
  {ICCV} 2019, Seoul, Korea (South), October 27 - November 2, 2019}, pp.\
  5006--5015. {IEEE}, 2019.

\bibitem[Pennington et~al.(2014)Pennington, Socher, and
  Manning]{DBLP:conf/emnlp/PenningtonSM14}
Jeffrey Pennington, Richard Socher, and Christopher~D. Manning.
\newblock Glove: Global vectors for word representation.
\newblock In \emph{{EMNLP}}, pp.\  1532--1543. {ACL}, 2014.

\bibitem[Peters et~al.(2019)Peters, Niculae, and
  Martins]{DBLP:conf/acl/PetersNM19}
Ben Peters, Vlad Niculae, and Andr{\'{e}} F.~T. Martins.
\newblock Sparse sequence-to-sequence models.
\newblock In \emph{{ACL} {(1)}}, pp.\  1504--1519. Association for
  Computational Linguistics, 2019.

\bibitem[Peters et~al.(2018)Peters, Neumann, Iyyer, Gardner, Clark, Lee, and
  Zettlemoyer]{DBLP:conf/naacl/PetersNIGCLZ18}
Matthew~E. Peters, Mark Neumann, Mohit Iyyer, Matt Gardner, Christopher Clark,
  Kenton Lee, and Luke Zettlemoyer.
\newblock Deep contextualized word representations.
\newblock In \emph{{NAACL-HLT}}, pp.\  2227--2237. Association for
  Computational Linguistics, 2018.

\bibitem[Petroni et~al.(2019)Petroni, Rockt{\"{a}}schel, Riedel, Lewis,
  Bakhtin, Wu, and Miller]{DBLP:conf/emnlp/PetroniRRLBWM19}
Fabio Petroni, Tim Rockt{\"{a}}schel, Sebastian Riedel, Patrick S.~H. Lewis,
  Anton Bakhtin, Yuxiang Wu, and Alexander~H. Miller.
\newblock Language models as knowledge bases?
\newblock In Kentaro Inui, Jing Jiang, Vincent Ng, and Xiaojun Wan (eds.),
  \emph{Proceedings of the 2019 Conference on Empirical Methods in Natural
  Language Processing and the 9th International Joint Conference on Natural
  Language Processing, {EMNLP-IJCNLP} 2019, Hong Kong, China, November 3-7,
  2019}, pp.\  2463--2473. Association for Computational Linguistics, 2019.

\bibitem[Pfeiffer et~al.(2020)Pfeiffer, R{\"{u}}ckl{\'{e}}, Poth, Kamath,
  Vulic, Ruder, Cho, and Gurevych]{DBLP:conf/emnlp/PfeifferRPKVRCG20}
Jonas Pfeiffer, Andreas R{\"{u}}ckl{\'{e}}, Clifton Poth, Aishwarya Kamath,
  Ivan Vulic, Sebastian Ruder, Kyunghyun Cho, and Iryna Gurevych.
\newblock Adapterhub: {A} framework for adapting transformers.
\newblock In \emph{{EMNLP} (Demos)}, pp.\  46--54. Association for
  Computational Linguistics, 2020.

\bibitem[Pham et~al.(2018)Pham, Guan, Zoph, Le, and Dean]{pham2018efficient}
Hieu Pham, Melody Guan, Barret Zoph, Quoc Le, and Jeff Dean.
\newblock Efficient neural architecture search via parameters sharing.
\newblock In \emph{International Conference on Machine Learning}, pp.\
  4095--4104. PMLR, 2018.

\bibitem[Pleiss et~al.(2017)Pleiss, Chen, Huang, Li, van~der Maaten, and
  Weinberger]{DBLP:journals/corr/PleissCHLMW17}
Geoff Pleiss, Danlu Chen, Gao Huang, Tongcheng Li, Laurens van~der Maaten, and
  Kilian~Q. Weinberger.
\newblock Memory-efficient implementation of densenets.
\newblock \emph{CoRR}, abs/1707.06990, 2017.

\bibitem[Plummer et~al.(2020)Plummer, Dryden, Frost, Hoefler, and
  Saenko]{plummer2020neural}
Bryan~A Plummer, Nikoli Dryden, Julius Frost, Torsten Hoefler, and Kate Saenko.
\newblock Neural parameter allocation search.
\newblock \emph{arXiv preprint arXiv:2006.10598}, 2020.

\bibitem[Polino et~al.(2018)Polino, Pascanu, and
  Alistarh]{DBLP:conf/iclr/PolinoPA18}
Antonio Polino, Razvan Pascanu, and Dan Alistarh.
\newblock Model compression via distillation and quantization.
\newblock In \emph{6th International Conference on Learning Representations,
  {ICLR} 2018, Vancouver, BC, Canada, April 30 - May 3, 2018, Conference Track
  Proceedings}. OpenReview.net, 2018.

\bibitem[Prasanna et~al.(2020)Prasanna, Rogers, and
  Rumshisky]{DBLP:conf/emnlp/PrasannaRR20}
Sai Prasanna, Anna Rogers, and Anna Rumshisky.
\newblock When {BERT} plays the lottery, all tickets are winning.
\newblock In \emph{{EMNLP} {(1)}}, pp.\  3208--3229. Association for
  Computational Linguistics, 2020.

\bibitem[Prato et~al.(2020)Prato, Charlaix, and
  Rezagholizadeh]{DBLP:conf/emnlp/PratoCR20}
Gabriele Prato, Ella Charlaix, and Mehdi Rezagholizadeh.
\newblock Fully quantized transformer for machine translation.
\newblock In Trevor Cohn, Yulan He, and Yang Liu (eds.), \emph{Proceedings of
  the 2020 Conference on Empirical Methods in Natural Language Processing:
  Findings, {EMNLP} 2020, Online Event, 16-20 November 2020}, pp.\  1--14.
  Association for Computational Linguistics, 2020.

\bibitem[Press \& Wolf(2017)Press and Wolf]{DBLP:conf/eacl/PressW17}
Ofir Press and Lior Wolf.
\newblock Using the output embedding to improve language models.
\newblock In \emph{{EACL} {(2)}}, pp.\  157--163. Association for Computational
  Linguistics, 2017.

\bibitem[Qi et~al.(2020)Qi, Yan, Gong, Liu, Duan, Chen, Zhang, and
  Zhou]{DBLP:conf/emnlp/QiYGLDCZ020}
Weizhen Qi, Yu~Yan, Yeyun Gong, Dayiheng Liu, Nan Duan, Jiusheng Chen, Ruofei
  Zhang, and Ming Zhou.
\newblock Prophetnet: Predicting future n-gram for sequence-to-sequence
  pre-training.
\newblock In Trevor Cohn, Yulan He, and Yang Liu (eds.), \emph{Findings of the
  Association for Computational Linguistics: {EMNLP} 2020, Online Event, 16-20
  November 2020}, volume {EMNLP} 2020 of \emph{Findings of {ACL}}, pp.\
  2401--2410, 2020.

\bibitem[Qi et~al.(2018)Qi, Sachan, Felix, Padmanabhan, and
  Neubig]{DBLP:conf/naacl/QiSFPN18}
Ye~Qi, Devendra~Singh Sachan, Matthieu Felix, Sarguna Padmanabhan, and Graham
  Neubig.
\newblock When and why are pre-trained word embeddings useful for neural
  machine translation?
\newblock In \emph{{NAACL-HLT} {(2)}}, pp.\  529--535. Association for
  Computational Linguistics, 2018.

\bibitem[Qiu et~al.(2020)Qiu, Sun, Xu, Shao, Dai, and
  Huang]{DBLP:journals/corr/abs-2003-08271}
Xipeng Qiu, Tianxiang Sun, Yige Xu, Yunfan Shao, Ning Dai, and Xuanjing Huang.
\newblock Pre-trained models for natural language processing: {A} survey.
\newblock \emph{CoRR}, abs/2003.08271, 2020.

\bibitem[Radford et~al.(2019)Radford, Wu, Child, Luan, Amodei, Sutskever,
  et~al.]{radford2019language}
Alec Radford, Jeffrey Wu, Rewon Child, David Luan, Dario Amodei, Ilya
  Sutskever, et~al.
\newblock Language models are unsupervised multitask learners.
\newblock \emph{OpenAI blog}, 1\penalty0 (8):\penalty0 9, 2019.

\bibitem[Raffel et~al.(2020)Raffel, Shazeer, Roberts, Lee, Narang, Matena,
  Zhou, Li, and Liu]{DBLP:journals/jmlr/RaffelSRLNMZLL20}
Colin Raffel, Noam Shazeer, Adam Roberts, Katherine Lee, Sharan Narang, Michael
  Matena, Yanqi Zhou, Wei Li, and Peter~J. Liu.
\newblock Exploring the limits of transfer learning with a unified text-to-text
  transformer.
\newblock \emph{J. Mach. Learn. Res.}, 21:\penalty0 140:1--140:67, 2020.

\bibitem[Rajbhandari et~al.(2020)Rajbhandari, Rasley, Ruwase, and
  He]{DBLP:conf/sc/RajbhandariRRH20}
Samyam Rajbhandari, Jeff Rasley, Olatunji Ruwase, and Yuxiong He.
\newblock Zero: memory optimizations toward training trillion parameter models.
\newblock In \emph{{SC}}, pp.\ ~20. {IEEE/ACM}, 2020.

\bibitem[Rajpurkar et~al.(2016)Rajpurkar, Zhang, Lopyrev, and
  Liang]{DBLP:conf/emnlp/RajpurkarZLL16}
Pranav Rajpurkar, Jian Zhang, Konstantin Lopyrev, and Percy Liang.
\newblock Squad: 100, 000+ questions for machine comprehension of text.
\newblock In Jian Su, Xavier Carreras, and Kevin Duh (eds.), \emph{Proceedings
  of the 2016 Conference on Empirical Methods in Natural Language Processing,
  {EMNLP} 2016, Austin, Texas, USA, November 1-4, 2016}, pp.\  2383--2392. The
  Association for Computational Linguistics, 2016.

\bibitem[Rajpurkar et~al.(2018)Rajpurkar, Jia, and
  Liang]{DBLP:conf/acl/RajpurkarJL18}
Pranav Rajpurkar, Robin Jia, and Percy Liang.
\newblock Know what you don't know: Unanswerable questions for squad.
\newblock In Iryna Gurevych and Yusuke Miyao (eds.), \emph{Proceedings of the
  56th Annual Meeting of the Association for Computational Linguistics, {ACL}
  2018, Melbourne, Australia, July 15-20, 2018, Volume 2: Short Papers}, pp.\
  784--789. Association for Computational Linguistics, 2018.

\bibitem[Ramsundar et~al.(2015)Ramsundar, Kearnes, Riley, Webster, Konerding,
  and Pande]{DBLP:journals/corr/RamsundarKRWKP15}
Bharath Ramsundar, Steven~M. Kearnes, Patrick Riley, Dale Webster, David~E.
  Konerding, and Vijay~S. Pande.
\newblock Massively multitask networks for drug discovery.
\newblock \emph{CoRR}, abs/1502.02072, 2015.

\bibitem[Ranganathan et~al.(2017)Ranganathan, Venkateswara, Chakraborty, and
  Panchanathan]{DBLP:conf/icip/RanganathanVCP17}
Hiranmayi Ranganathan, Hemanth Venkateswara, Shayok Chakraborty, and Sethuraman
  Panchanathan.
\newblock Deep active learning for image classification.
\newblock In \emph{{ICIP}}, pp.\  3934--3938. {IEEE}, 2017.

\bibitem[Rastegari et~al.(2016)Rastegari, Ordonez, Redmon, and
  Farhadi]{DBLP:conf/eccv/RastegariORF16}
Mohammad Rastegari, Vicente Ordonez, Joseph Redmon, and Ali Farhadi.
\newblock Xnor-net: Imagenet classification using binary convolutional neural
  networks.
\newblock In Bastian Leibe, Jiri Matas, Nicu Sebe, and Max Welling (eds.),
  \emph{Computer Vision - {ECCV} 2016 - 14th European Conference, Amsterdam,
  The Netherlands, October 11-14, 2016, Proceedings, Part {IV}}, volume 9908 of
  \emph{Lecture Notes in Computer Science}, pp.\  525--542. Springer, 2016.

\bibitem[Real et~al.(2019)Real, Aggarwal, Huang, and
  Le]{DBLP:conf/aaai/RealAHL19}
Esteban Real, Alok Aggarwal, Yanping Huang, and Quoc~V. Le.
\newblock Regularized evolution for image classifier architecture search.
\newblock In \emph{The Thirty-Third {AAAI} Conference on Artificial
  Intelligence, {AAAI} 2019, The Thirty-First Innovative Applications of
  Artificial Intelligence Conference, {IAAI} 2019, The Ninth {AAAI} Symposium
  on Educational Advances in Artificial Intelligence, {EAAI} 2019, Honolulu,
  Hawaii, USA, January 27 - February 1, 2019}, pp.\  4780--4789, 2019.

\bibitem[Reddy et~al.(2019)Reddy, Chen, and
  Manning]{DBLP:journals/tacl/ReddyCM19}
Siva Reddy, Danqi Chen, and Christopher~D. Manning.
\newblock Coqa: {A} conversational question answering challenge.
\newblock \emph{Trans. Assoc. Comput. Linguistics}, 7:\penalty0 249--266, 2019.

\bibitem[Ren et~al.(2015)Ren, He, Girshick, and Sun]{DBLP:conf/nips/RenHGS15}
Shaoqing Ren, Kaiming He, Ross~B. Girshick, and Jian Sun.
\newblock Faster {R-CNN:} towards real-time object detection with region
  proposal networks.
\newblock In \emph{{NIPS}}, pp.\  91--99, 2015.

\bibitem[Rigamonti et~al.(2013)Rigamonti, Sironi, Lepetit, and
  Fua]{DBLP:conf/cvpr/RigamontiSLF13}
Roberto Rigamonti, Amos Sironi, Vincent Lepetit, and Pascal Fua.
\newblock Learning separable filters.
\newblock In \emph{2013 {IEEE} Conference on Computer Vision and Pattern
  Recognition, Portland, OR, USA, June 23-28, 2013}, pp.\  2754--2761. {IEEE}
  Computer Society, 2013.

\bibitem[Romero et~al.(2015)Romero, Ballas, Kahou, Chassang, Gatta, and
  Bengio]{DBLP:journals/corr/RomeroBKCGB14}
Adriana Romero, Nicolas Ballas, Samira~Ebrahimi Kahou, Antoine Chassang, Carlo
  Gatta, and Yoshua Bengio.
\newblock Fitnets: Hints for thin deep nets.
\newblock In Yoshua Bengio and Yann LeCun (eds.), \emph{3rd International
  Conference on Learning Representations, {ICLR} 2015, San Diego, CA, USA, May
  7-9, 2015, Conference Track Proceedings}, 2015.

\bibitem[Roy \& McCallum(2001)Roy and McCallum]{roy2001toward}
Nicholas Roy and Andrew McCallum.
\newblock Toward optimal active learning through monte carlo estimation of
  error reduction.
\newblock \emph{ICML, Williamstown}, 2:\penalty0 441--448, 2001.

\bibitem[Ruder(2019)]{ruder2019neural}
Sebastian Ruder.
\newblock \emph{Neural transfer learning for natural language processing}.
\newblock PhD thesis, NUI Galway, 2019.

\bibitem[Ruder et~al.(2017)Ruder, Ghaffari, and
  Breslin]{DBLP:journals/corr/RuderGB17}
Sebastian Ruder, Parsa Ghaffari, and John~G. Breslin.
\newblock Knowledge adaptation: Teaching to adapt.
\newblock \emph{CoRR}, abs/1702.02052, 2017.

\bibitem[Ruder et~al.(2019)Ruder, Vulic, and
  S{\o}gaard]{DBLP:journals/jair/RuderVS19}
Sebastian Ruder, Ivan Vulic, and Anders S{\o}gaard.
\newblock A survey of cross-lingual word embedding models.
\newblock \emph{J. Artif. Intell. Res.}, 65:\penalty0 569--631, 2019.

\bibitem[Rueda{-}Plata et~al.(2015)Rueda{-}Plata, Ramos{-}Poll{\'{a}}n, and
  Gonz{\'{a}}lez]{DBLP:conf/iccci/Rueda-PlataRG15}
Diego Rueda{-}Plata, Ra{\'{u}}l Ramos{-}Poll{\'{a}}n, and Fabio~A.
  Gonz{\'{a}}lez.
\newblock Supervised greedy layer-wise training for deep convolutional networks
  with small datasets.
\newblock In Manuel N{\'{u}}{\~{n}}ez, Ngoc~Thanh Nguyen, David Camacho, and
  Bogdan Trawinski (eds.), \emph{Computational Collective Intelligence - 7th
  International Conference, {ICCCI} 2015, Madrid, Spain, September 21-23, 2015.
  Proceedings, Part {I}}, volume 9329 of \emph{Lecture Notes in Computer
  Science}, pp.\  275--284. Springer, 2015.

\bibitem[Rush et~al.(2015)Rush, Chopra, and Weston]{DBLP:conf/emnlp/RushCW15}
Alexander~M. Rush, Sumit Chopra, and Jason Weston.
\newblock A neural attention model for abstractive sentence summarization.
\newblock In Llu{\'{\i}}s M{\`{a}}rquez, Chris Callison{-}Burch, Jian Su,
  Daniele Pighin, and Yuval Marton (eds.), \emph{Proceedings of the 2015
  Conference on Empirical Methods in Natural Language Processing, {EMNLP} 2015,
  Lisbon, Portugal, September 17-21, 2015}, pp.\  379--389. The Association for
  Computational Linguistics, 2015.

\bibitem[Sainath et~al.(2013)Sainath, Kingsbury, Sindhwani, Arisoy, and
  Ramabhadran]{DBLP:conf/icassp/SainathKSAR13}
Tara~N. Sainath, Brian Kingsbury, Vikas Sindhwani, Ebru Arisoy, and Bhuvana
  Ramabhadran.
\newblock Low-rank matrix factorization for deep neural network training with
  high-dimensional output targets.
\newblock In \emph{{ICASSP}}, pp.\  6655--6659. {IEEE}, 2013.

\bibitem[Salimans \& Kingma(2016)Salimans and
  Kingma]{DBLP:conf/nips/SalimansK16}
Tim Salimans and Diederik~P. Kingma.
\newblock Weight normalization: {A} simple reparameterization to accelerate
  training of deep neural networks.
\newblock In \emph{{NIPS}}, pp.\  901, 2016.

\bibitem[Sandler et~al.(2018)Sandler, Howard, Zhu, Zhmoginov, and
  Chen]{DBLP:conf/cvpr/SandlerHZZC18}
Mark Sandler, Andrew~G. Howard, Menglong Zhu, Andrey Zhmoginov, and
  Liang{-}Chieh Chen.
\newblock Mobilenetv2: Inverted residuals and linear bottlenecks.
\newblock In \emph{2018 {IEEE} Conference on Computer Vision and Pattern
  Recognition, {CVPR} 2018, Salt Lake City, UT, USA, June 18-22, 2018}, pp.\
  4510--4520. Computer Vision Foundation / {IEEE} Computer Society, 2018.

\bibitem[Sanh et~al.(2020)Sanh, Wolf, and Rush]{DBLP:conf/nips/Sanh0R20}
Victor Sanh, Thomas Wolf, and Alexander~M. Rush.
\newblock Movement pruning: Adaptive sparsity by fine-tuning.
\newblock In Hugo Larochelle, Marc'Aurelio Ranzato, Raia Hadsell,
  Maria{-}Florina Balcan, and Hsuan{-}Tien Lin (eds.), \emph{Advances in Neural
  Information Processing Systems 33: Annual Conference on Neural Information
  Processing Systems 2020, NeurIPS 2020, December 6-12, 2020, virtual}, 2020.

\bibitem[Santurkar et~al.(2018)Santurkar, Tsipras, Ilyas, and
  Madry]{DBLP:conf/nips/SanturkarTIM18}
Shibani Santurkar, Dimitris Tsipras, Andrew Ilyas, and Aleksander Madry.
\newblock How does batch normalization help optimization?
\newblock In \emph{NeurIPS}, pp.\  2488--2498, 2018.

\bibitem[Savarese \& Maire(2019)Savarese and Maire]{DBLP:conf/iclr/SavareseM19}
Pedro Savarese and Michael Maire.
\newblock Learning implicitly recurrent cnns through parameter sharing.
\newblock In \emph{{ICLR} (Poster)}. OpenReview.net, 2019.

\bibitem[Saxe et~al.(2014)Saxe, McClelland, and
  Ganguli]{DBLP:journals/corr/SaxeMG13}
Andrew~M. Saxe, James~L. McClelland, and Surya Ganguli.
\newblock Exact solutions to the nonlinear dynamics of learning in deep linear
  neural networks.
\newblock In \emph{{ICLR}}, 2014.

\bibitem[Scao \& Rush(2021)Scao and Rush]{DBLP:conf/naacl/ScaoR21}
Teven~Le Scao and Alexander~M. Rush.
\newblock How many data points is a prompt worth?
\newblock In \emph{{NAACL-HLT}}, pp.\  2627--2636. Association for
  Computational Linguistics, 2021.

\bibitem[Schick \& Sch{\"{u}}tze(2021)Schick and
  Sch{\"{u}}tze]{DBLP:conf/naacl/SchickS21}
Timo Schick and Hinrich Sch{\"{u}}tze.
\newblock It's not just size that matters: Small language models are also
  few-shot learners.
\newblock In \emph{{NAACL-HLT}}, pp.\  2339--2352. Association for
  Computational Linguistics, 2021.

\bibitem[Schr{\"{o}}der et~al.(2021)Schr{\"{o}}der, Niekler, and
  Potthast]{DBLP:journals/corr/abs-2107-05687}
Christopher Schr{\"{o}}der, Andreas Niekler, and Martin Potthast.
\newblock Uncertainty-based query strategies for active learning with
  transformers.
\newblock \emph{CoRR}, abs/2107.05687, 2021.

\bibitem[Schuster et~al.(2021)Schuster, Fisch, Jaakkola, and
  Barzilay]{DBLP:journals/corr/abs-2104-08803}
Tal Schuster, Adam Fisch, Tommi~S. Jaakkola, and Regina Barzilay.
\newblock Consistent accelerated inference via confident adaptive transformers.
\newblock \emph{CoRR}, abs/2104.08803, 2021.

\bibitem[Schwartz et~al.(2020{\natexlab{a}})Schwartz, Dodge, Smith, and
  Etzioni]{DBLP:journals/cacm/SchwartzDSE20}
Roy Schwartz, Jesse Dodge, Noah~A. Smith, and Oren Etzioni.
\newblock Green {AI}.
\newblock \emph{Commun. {ACM}}, 63\penalty0 (12):\penalty0 54--63,
  2020{\natexlab{a}}.

\bibitem[Schwartz et~al.(2020{\natexlab{b}})Schwartz, Stanovsky, Swayamdipta,
  Dodge, and Smith]{DBLP:conf/acl/SchwartzSSDS20}
Roy Schwartz, Gabriel Stanovsky, Swabha Swayamdipta, Jesse Dodge, and Noah~A.
  Smith.
\newblock The right tool for the job: Matching model and instance complexities.
\newblock In Dan Jurafsky, Joyce Chai, Natalie Schluter, and Joel~R. Tetreault
  (eds.), \emph{Proceedings of the 58th Annual Meeting of the Association for
  Computational Linguistics, {ACL} 2020, Online, July 5-10, 2020}, pp.\
  6640--6651. Association for Computational Linguistics, 2020{\natexlab{b}}.

\bibitem[See et~al.(2016)See, Luong, and Manning]{DBLP:conf/conll/SeeLM16}
Abigail See, Minh{-}Thang Luong, and Christopher~D. Manning.
\newblock Compression of neural machine translation models via pruning.
\newblock In \emph{Proceedings of the 20th {SIGNLL} Conference on Computational
  Natural Language Learning, CoNLL 2016, Berlin, Germany, August 11-12, 2016},
  pp.\  291--301, 2016.

\bibitem[See et~al.(2017)See, Liu, and Manning]{DBLP:conf/acl/SeeLM17}
Abigail See, Peter~J. Liu, and Christopher~D. Manning.
\newblock Get to the point: Summarization with pointer-generator networks.
\newblock In Regina Barzilay and Min{-}Yen Kan (eds.), \emph{Proceedings of the
  55th Annual Meeting of the Association for Computational Linguistics, {ACL}
  2017, Vancouver, Canada, July 30 - August 4, Volume 1: Long Papers}, pp.\
  1073--1083. Association for Computational Linguistics, 2017.

\bibitem[Sener \& Savarese(2018)Sener and Savarese]{DBLP:conf/iclr/SenerS18}
Ozan Sener and Silvio Savarese.
\newblock Active learning for convolutional neural networks: {A} core-set
  approach.
\newblock In \emph{{ICLR} (Poster)}. OpenReview.net, 2018.

\bibitem[Sennrich et~al.(2016)Sennrich, Haddow, and
  Birch]{DBLP:conf/acl/SennrichHB16a}
Rico Sennrich, Barry Haddow, and Alexandra Birch.
\newblock Neural machine translation of rare words with subword units.
\newblock In \emph{{ACL} {(1)}}. The Association for Computer Linguistics,
  2016.

\bibitem[Seo et~al.(2018)Seo, Min, Farhadi, and
  Hajishirzi]{DBLP:conf/iclr/SeoMFH18}
Min~Joon Seo, Sewon Min, Ali Farhadi, and Hannaneh Hajishirzi.
\newblock Neural speed reading via skim-rnn.
\newblock In \emph{6th International Conference on Learning Representations,
  {ICLR} 2018, Vancouver, BC, Canada, April 30 - May 3, 2018, Conference Track
  Proceedings}. OpenReview.net, 2018.

\bibitem[Shen et~al.(2018)Shen, Yun, Lipton, Kronrod, and
  Anandkumar]{DBLP:conf/iclr/ShenYLKA18}
Yanyao Shen, Hyokun Yun, Zachary~C. Lipton, Yakov Kronrod, and Animashree
  Anandkumar.
\newblock Deep active learning for named entity recognition.
\newblock In \emph{{ICLR} (Poster)}. OpenReview.net, 2018.

\bibitem[Shen et~al.(2017)Shen, Huang, Gao, and Chen]{DBLP:conf/kdd/ShenHGC17}
Yelong Shen, Po{-}Sen Huang, Jianfeng Gao, and Weizhu Chen.
\newblock Reasonet: Learning to stop reading in machine comprehension.
\newblock In \emph{Proceedings of the 23rd {ACM} {SIGKDD} International
  Conference on Knowledge Discovery and Data Mining, Halifax, NS, Canada,
  August 13 - 17, 2017}, pp.\  1047--1055. {ACM}, 2017.

\bibitem[Sheng et~al.(2018)Sheng, Feng, Zhuo, Zhang, Shen, and
  Aleksic]{DBLP:journals/corr/abs-1803-08607}
Tao Sheng, Chen Feng, Shaojie Zhuo, Xiaopeng Zhang, Liang Shen, and Mickey
  Aleksic.
\newblock A quantization-friendly separable convolution for mobilenets.
\newblock \emph{CoRR}, abs/1803.08607, 2018.

\bibitem[Shi et~al.(2020)Shi, Mudigere, Naumov, and
  Yang]{DBLP:conf/kdd/ShiMNY20}
Hao{-}Jun~Michael Shi, Dheevatsa Mudigere, Maxim Naumov, and Jiyan Yang.
\newblock Compositional embeddings using complementary partitions for
  memory-efficient recommendation systems.
\newblock In \emph{{KDD}}, pp.\  165--175. {ACM}, 2020.

\bibitem[Shi et~al.(2021)Shi, Song, Zhou, Li, and Li]{shi2021learning}
Wenxian Shi, Yuxuan Song, Hao Zhou, Bohan Li, and Lei Li.
\newblock Learning from deep model via exploring local targets, 2021.
\newblock URL \url{https://openreview.net/forum?id=5slGDu_bVc6}.

\bibitem[Shu \& Nakayama(2018)Shu and Nakayama]{DBLP:conf/iclr/ShuN18}
Raphael Shu and Hideki Nakayama.
\newblock Compressing word embeddings via deep compositional code learning.
\newblock In \emph{{ICLR} (Poster)}. OpenReview.net, 2018.

\bibitem[Shu et~al.(2020)Shu, Wang, and Cai]{DBLP:conf/iclr/Shu0C20}
Yao Shu, Wei Wang, and Shaofeng Cai.
\newblock Understanding architectures learnt by cell-based neural architecture
  search.
\newblock In \emph{{ICLR}}. OpenReview.net, 2020.

\bibitem[Simon et~al.(2016)Simon, Rodner, and
  Denzler]{DBLP:journals/corr/SimonRD16}
Marcel Simon, Erik Rodner, and Joachim Denzler.
\newblock Imagenet pre-trained models with batch normalization.
\newblock \emph{CoRR}, abs/1612.01452, 2016.

\bibitem[Simonyan \& Zisserman(2015)Simonyan and
  Zisserman]{DBLP:journals/corr/SimonyanZ14a}
Karen Simonyan and Andrew Zisserman.
\newblock Very deep convolutional networks for large-scale image recognition.
\newblock In \emph{{ICLR}}, 2015.

\bibitem[So et~al.(2019)So, Le, and Liang]{DBLP:conf/icml/SoLL19}
David~R. So, Quoc~V. Le, and Chen Liang.
\newblock The evolved transformer.
\newblock In \emph{Proceedings of the 36th International Conference on Machine
  Learning, {ICML} 2019, 9-15 June 2019, Long Beach, California, {USA}}, pp.\
  5877--5886, 2019.

\bibitem[Socher et~al.(2013)Socher, Perelygin, Wu, Chuang, Manning, Ng, and
  Potts]{DBLP:conf/emnlp/SocherPWCMNP13}
Richard Socher, Alex Perelygin, Jean Wu, Jason Chuang, Christopher~D. Manning,
  Andrew~Y. Ng, and Christopher Potts.
\newblock Recursive deep models for semantic compositionality over a sentiment
  treebank.
\newblock In \emph{Proceedings of the 2013 Conference on Empirical Methods in
  Natural Language Processing, {EMNLP} 2013, 18-21 October 2013, Grand Hyatt
  Seattle, Seattle, Washington, USA, {A} meeting of SIGDAT, a Special Interest
  Group of the {ACL}}, pp.\  1631--1642. {ACL}, 2013.

\bibitem[S{\o}gaard \& Goldberg(2016)S{\o}gaard and
  Goldberg]{DBLP:conf/acl/SogaardG16}
Anders S{\o}gaard and Yoav Goldberg.
\newblock Deep multi-task learning with low level tasks supervised at lower
  layers.
\newblock In \emph{{ACL} {(2)}}. The Association for Computer Linguistics,
  2016.

\bibitem[Soudry et~al.(2014)Soudry, Hubara, and
  Meir]{DBLP:conf/nips/SoudryHM14}
Daniel Soudry, Itay Hubara, and Ron Meir.
\newblock Expectation backpropagation: Parameter-free training of multilayer
  neural networks with continuous or discrete weights.
\newblock In Zoubin Ghahramani, Max Welling, Corinna Cortes, Neil~D. Lawrence,
  and Kilian~Q. Weinberger (eds.), \emph{Advances in Neural Information
  Processing Systems 27: Annual Conference on Neural Information Processing
  Systems 2014, December 8-13 2014, Montreal, Quebec, Canada}, pp.\  963--971,
  2014.

\bibitem[Srinivas \& Babu(2015)Srinivas and Babu]{DBLP:conf/bmvc/SrinivasB15}
Suraj Srinivas and R.~Venkatesh Babu.
\newblock Data-free parameter pruning for deep neural networks.
\newblock In \emph{{BMVC}}, pp.\  31.1--31.12. {BMVA} Press, 2015.

\bibitem[Srinivas \& Fleuret(2018)Srinivas and
  Fleuret]{DBLP:conf/icml/SrinivasF18}
Suraj Srinivas and Fran{\c{c}}ois Fleuret.
\newblock Knowledge transfer with jacobian matching.
\newblock In Jennifer~G. Dy and Andreas Krause (eds.), \emph{Proceedings of the
  35th International Conference on Machine Learning, {ICML} 2018,
  Stockholmsm{\"{a}}ssan, Stockholm, Sweden, July 10-15, 2018}, volume~80 of
  \emph{Proceedings of Machine Learning Research}, pp.\  4730--4738. {PMLR},
  2018.

\bibitem[Stanton et~al.(2021)Stanton, Izmailov, Kirichenko, Alemi, and
  Wilson]{DBLP:journals/corr/abs-2106-05945}
Samuel Stanton, Pavel Izmailov, Polina Kirichenko, Alexander~A. Alemi, and
  Andrew~Gordon Wilson.
\newblock Does knowledge distillation really work?
\newblock \emph{CoRR}, abs/2106.05945, 2021.

\bibitem[Strubell et~al.(2019)Strubell, Ganesh, and
  McCallum]{DBLP:conf/acl/StrubellGM19}
Emma Strubell, Ananya Ganesh, and Andrew McCallum.
\newblock Energy and policy considerations for deep learning in {NLP}.
\newblock In \emph{Proceedings of the 57th Conference of the Association for
  Computational Linguistics, {ACL} 2019, Florence, Italy, July 28- August 2,
  2019, Volume 1: Long Papers}, pp.\  3645--3650, 2019.

\bibitem[Sukhbaatar et~al.(2019)Sukhbaatar, Grave, Bojanowski, and
  Joulin]{DBLP:conf/acl/SukhbaatarGBJ19}
Sainbayar Sukhbaatar, Edouard Grave, Piotr Bojanowski, and Armand Joulin.
\newblock Adaptive attention span in transformers.
\newblock In \emph{{ACL} {(1)}}, pp.\  331--335. Association for Computational
  Linguistics, 2019.

\bibitem[Sun et~al.(2019{\natexlab{a}})Sun, Cheng, Gan, and
  Liu]{DBLP:conf/emnlp/SunCGL19}
Siqi Sun, Yu~Cheng, Zhe Gan, and Jingjing Liu.
\newblock Patient knowledge distillation for {BERT} model compression.
\newblock In Kentaro Inui, Jing Jiang, Vincent Ng, and Xiaojun Wan (eds.),
  \emph{Proceedings of the 2019 Conference on Empirical Methods in Natural
  Language Processing and the 9th International Joint Conference on Natural
  Language Processing, {EMNLP-IJCNLP} 2019, Hong Kong, China, November 3-7,
  2019}, pp.\  4322--4331. Association for Computational Linguistics,
  2019{\natexlab{a}}.

\bibitem[Sun et~al.(2021)Sun, Zhou, Liu, Zhang, Jiang, Cao, Huang, and
  Qiu]{DBLP:journals/corr/abs-2105-13792}
Tianxiang Sun, Yunhua Zhou, Xiangyang Liu, Xinyu Zhang, Hao Jiang, Zhao Cao,
  Xuanjing Huang, and Xipeng Qiu.
\newblock Early exiting with ensemble internal classifiers.
\newblock \emph{CoRR}, abs/2105.13792, 2021.

\bibitem[Sun et~al.(2019{\natexlab{b}})Sun, Wang, Li, Feng, Chen, Zhang, Tian,
  Zhu, Tian, and Wu]{DBLP:journals/corr/abs-1904-09223}
Yu~Sun, Shuohuan Wang, Yu{-}Kun Li, Shikun Feng, Xuyi Chen, Han Zhang, Xin
  Tian, Danxiang Zhu, Hao Tian, and Hua Wu.
\newblock {ERNIE:} enhanced representation through knowledge integration.
\newblock \emph{CoRR}, abs/1904.09223, 2019{\natexlab{b}}.

\bibitem[Sutton \& McCallum(2012)Sutton and
  McCallum]{DBLP:journals/ftml/SuttonM12}
Charles Sutton and Andrew McCallum.
\newblock An introduction to conditional random fields.
\newblock \emph{Found. Trends Mach. Learn.}, 4\penalty0 (4):\penalty0 267--373,
  2012.

\bibitem[Suzuki et~al.(2001)Suzuki, Horiba, and
  Sugie]{DBLP:journals/npl/SuzukiHS01}
Kenji Suzuki, Isao Horiba, and Noboru Sugie.
\newblock A simple neural network pruning algorithm with application to filter
  synthesis.
\newblock \emph{Neural Process. Lett.}, 13\penalty0 (1):\penalty0 43--53, 2001.

\bibitem[Szegedy et~al.(2015)Szegedy, Liu, Jia, Sermanet, Reed, Anguelov,
  Erhan, Vanhoucke, and Rabinovich]{DBLP:conf/cvpr/SzegedyLJSRAEVR15}
Christian Szegedy, Wei Liu, Yangqing Jia, Pierre Sermanet, Scott~E. Reed,
  Dragomir Anguelov, Dumitru Erhan, Vincent Vanhoucke, and Andrew Rabinovich.
\newblock Going deeper with convolutions.
\newblock In \emph{{CVPR}}, pp.\  1--9. {IEEE} Computer Society, 2015.

\bibitem[Szegedy et~al.(2017)Szegedy, Ioffe, Vanhoucke, and
  Alemi]{DBLP:conf/aaai/SzegedyIVA17}
Christian Szegedy, Sergey Ioffe, Vincent Vanhoucke, and Alexander~A. Alemi.
\newblock Inception-v4, inception-resnet and the impact of residual connections
  on learning.
\newblock In \emph{{AAAI}}, pp.\  4278--4284. {AAAI} Press, 2017.

\bibitem[Tailor et~al.(2021)Tailor, Fernandez-Marques, and
  Lane]{tailor2021degreequant}
Shyam~Anil Tailor, Javier Fernandez-Marques, and Nicholas~Donald Lane.
\newblock Degree-quant: Quantization-aware training for graph neural networks.
\newblock In \emph{International Conference on Learning Representations}, 2021.

\bibitem[Takase \& Kiyono(2021)Takase and
  Kiyono]{DBLP:journals/corr/abs-2104-06022}
Sho Takase and Shun Kiyono.
\newblock Lessons on parameter sharing across layers in transformers.
\newblock \emph{CoRR}, abs/2104.06022, 2021.

\bibitem[Tambe et~al.(2021)Tambe, Hooper, Pentecost, Jia, Yang, Donato, Sanh,
  Whatmough, Rush, Brooks, and Wei]{DBLP:conf/micro/TambeHPJYDSWR0W21}
Thierry Tambe, Coleman Hooper, Lillian Pentecost, Tianyu Jia, En{-}Yu Yang,
  Marco Donato, Victor Sanh, Paul~N. Whatmough, Alexander~M. Rush, David
  Brooks, and Gu{-}Yeon Wei.
\newblock Edgebert: Sentence-level energy optimizations for latency-aware
  multi-task {NLP} inference.
\newblock In \emph{{MICRO}}, pp.\  830--844. {ACM}, 2021.

\bibitem[Tan \& Le(2019)Tan and Le]{DBLP:conf/icml/TanL19}
Mingxing Tan and Quoc~V. Le.
\newblock Efficientnet: Rethinking model scaling for convolutional neural
  networks.
\newblock 97:\penalty0 6105--6114, 2019.

\bibitem[Tan et~al.(2019{\natexlab{a}})Tan, Chen, Pang, Vasudevan, Sandler,
  Howard, and Le]{DBLP:conf/cvpr/TanCPVSHL19}
Mingxing Tan, Bo~Chen, Ruoming Pang, Vijay Vasudevan, Mark Sandler, Andrew
  Howard, and Quoc~V. Le.
\newblock Mnasnet: Platform-aware neural architecture search for mobile.
\newblock In \emph{{CVPR}}, pp.\  2820--2828. Computer Vision Foundation /
  {IEEE}, 2019{\natexlab{a}}.

\bibitem[Tan et~al.(2019{\natexlab{b}})Tan, Ren, He, Qin, Zhao, and
  Liu]{DBLP:conf/iclr/TanRHQZL19}
Xu~Tan, Yi~Ren, Di~He, Tao Qin, Zhou Zhao, and Tie{-}Yan Liu.
\newblock Multilingual neural machine translation with knowledge distillation.
\newblock In \emph{{ICLR} (Poster)}. OpenReview.net, 2019{\natexlab{b}}.

\bibitem[Tang \& Kwan(1993)Tang and Kwan]{229903}
C.Z. Tang and H.K. Kwan.
\newblock Multilayer feedforward neural networks with single powers-of-two
  weights.
\newblock \emph{IEEE Transactions on Signal Processing}, 41\penalty0
  (8):\penalty0 2724--2727, 1993.
\newblock \doi{10.1109/78.229903}.

\bibitem[Tarvainen \& Valpola(2017)Tarvainen and
  Valpola]{DBLP:conf/nips/TarvainenV17}
Antti Tarvainen and Harri Valpola.
\newblock Mean teachers are better role models: Weight-averaged consistency
  targets improve semi-supervised deep learning results.
\newblock In Isabelle Guyon, Ulrike von Luxburg, Samy Bengio, Hanna~M. Wallach,
  Rob Fergus, S.~V.~N. Vishwanathan, and Roman Garnett (eds.), \emph{Advances
  in Neural Information Processing Systems 30: Annual Conference on Neural
  Information Processing Systems 2017, December 4-9, 2017, Long Beach, CA,
  {USA}}, pp.\  1195--1204, 2017.

\bibitem[Teerapittayanon et~al.(2016)Teerapittayanon, McDanel, and
  Kung]{DBLP:conf/icpr/Teerapittayanon16}
Surat Teerapittayanon, Bradley McDanel, and H.~T. Kung.
\newblock Branchynet: Fast inference via early exiting from deep neural
  networks.
\newblock In \emph{23rd International Conference on Pattern Recognition, {ICPR}
  2016, Canc{\'{u}}n, Mexico, December 4-8, 2016}, pp.\  2464--2469. {IEEE},
  2016.

\bibitem[Tian et~al.(2020)Tian, Krishnan, and Isola]{DBLP:conf/eccv/TianKI20}
Yonglong Tian, Dilip Krishnan, and Phillip Isola.
\newblock Contrastive multiview coding.
\newblock In Andrea Vedaldi, Horst Bischof, Thomas Brox, and Jan{-}Michael
  Frahm (eds.), \emph{Computer Vision - {ECCV} 2020 - 16th European Conference,
  Glasgow, UK, August 23-28, 2020, Proceedings, Part {XI}}, volume 12356 of
  \emph{Lecture Notes in Computer Science}, pp.\  776--794. Springer, 2020.

\bibitem[Tung \& Mori(2019)Tung and Mori]{DBLP:conf/iccv/TungM19}
Frederick Tung and Greg Mori.
\newblock Similarity-preserving knowledge distillation.
\newblock In \emph{2019 {IEEE/CVF} International Conference on Computer Vision,
  {ICCV} 2019, Seoul, Korea (South), October 27 - November 2, 2019}, pp.\
  1365--1374. {IEEE}, 2019.

\bibitem[Upadhyay et~al.(2016)Upadhyay, Faruqui, Dyer, and
  Roth]{DBLP:conf/acl/UpadhyayFDR16}
Shyam Upadhyay, Manaal Faruqui, Chris Dyer, and Dan Roth.
\newblock Cross-lingual models of word embeddings: An empirical comparison.
\newblock In \emph{{ACL} {(1)}}. The Association for Computer Linguistics,
  2016.

\bibitem[van~den Oord et~al.(2018)van~den Oord, Li, and
  Vinyals]{DBLP:journals/corr/abs-1807-03748}
A{\"{a}}ron van~den Oord, Yazhe Li, and Oriol Vinyals.
\newblock Representation learning with contrastive predictive coding.
\newblock \emph{CoRR}, abs/1807.03748, 2018.

\bibitem[Vanhoucke et~al.(2011)Vanhoucke, Senior, and
  Mao]{vanhoucke2011improving}
Vincent Vanhoucke, Andrew Senior, and Mark~Z Mao.
\newblock {Improving the speed of neural networks on CPUs}.
\newblock In \emph{{Proc. Deep Learning and Unsupervised Feature Learning NIPS
  Workshop}}, 2011.

\bibitem[Vaswani et~al.(2017)Vaswani, Shazeer, Parmar, Uszkoreit, Jones, Gomez,
  Kaiser, and Polosukhin]{DBLP:conf/nips/VaswaniSPUJGKP17}
Ashish Vaswani, Noam Shazeer, Niki Parmar, Jakob Uszkoreit, Llion Jones,
  Aidan~N. Gomez, Lukasz Kaiser, and Illia Polosukhin.
\newblock Attention is all you need.
\newblock In \emph{{NIPS}}, pp.\  5998--6008, 2017.

\bibitem[Veit \& Belongie(2020)Veit and Belongie]{DBLP:journals/ijcv/VeitB20}
Andreas Veit and Serge~J. Belongie.
\newblock Convolutional networks with adaptive inference graphs.
\newblock \emph{Int. J. Comput. Vis.}, 128\penalty0 (3):\penalty0 730--741,
  2020.

\bibitem[Vijayanarasimhan et~al.(2015)Vijayanarasimhan, Shlens, Monga, and
  Yagnik]{DBLP:journals/corr/VijayanarasimhanSMY14}
Sudheendra Vijayanarasimhan, Jonathon Shlens, Rajat Monga, and Jay Yagnik.
\newblock Deep networks with large output spaces.
\newblock In \emph{{ICLR} (Workshop)}, 2015.

\bibitem[Viola \& Jones(2001)Viola and Jones]{viola2001rapid}
Paul Viola and Michael Jones.
\newblock Rapid object detection using a boosted cascade of simple features.
\newblock In \emph{Proceedings of the 2001 IEEE computer society conference on
  computer vision and pattern recognition. CVPR 2001}, volume~1, pp.\  I--I.
  Ieee, 2001.

\bibitem[Viola \& Jones(2004)Viola and Jones]{viola2004robust}
Paul Viola and Michael~J Jones.
\newblock Robust real-time face detection.
\newblock \emph{International journal of computer vision}, 57\penalty0
  (2):\penalty0 137--154, 2004.

\bibitem[Voita et~al.(2019)Voita, Talbot, Moiseev, Sennrich, and
  Titov]{DBLP:conf/acl/VoitaTMST19}
Elena Voita, David Talbot, Fedor Moiseev, Rico Sennrich, and Ivan Titov.
\newblock Analyzing multi-head self-attention: Specialized heads do the heavy
  lifting, the rest can be pruned.
\newblock In \emph{Proceedings of the 57th Conference of the Association for
  Computational Linguistics, {ACL} 2019, Florence, Italy, July 28- August 2,
  2019, Volume 1: Long Papers}, pp.\  5797--5808, 2019.

\bibitem[Vu et~al.(2021)Vu, Lester, Constant, Al-Rfou, and Cer]{vu2021spot}
Tu~Vu, Brian Lester, Noah Constant, Rami Al-Rfou, and Daniel Cer.
\newblock Spot: Better frozen model adaptation through soft prompt transfer.
\newblock \emph{arXiv preprint arXiv:2110.07904}, 2021.

\bibitem[Wang et~al.(2019{\natexlab{a}})Wang, Singh, Michael, Hill, Levy, and
  Bowman]{DBLP:conf/iclr/WangSMHLB19}
Alex Wang, Amanpreet Singh, Julian Michael, Felix Hill, Omer Levy, and
  Samuel~R. Bowman.
\newblock {GLUE:} {A} multi-task benchmark and analysis platform for natural
  language understanding.
\newblock In \emph{7th International Conference on Learning Representations,
  {ICLR} 2019, New Orleans, LA, USA, May 6-9, 2019}. OpenReview.net,
  2019{\natexlab{a}}.

\bibitem[Wang et~al.(2020{\natexlab{a}})Wang, Cho, and
  Gu]{DBLP:conf/aaai/WangCG20}
Changhan Wang, Kyunghyun Cho, and Jiatao Gu.
\newblock Neural machine translation with byte-level subwords.
\newblock In \emph{{AAAI}}, pp.\  9154--9160. {AAAI} Press, 2020{\natexlab{a}}.

\bibitem[Wang et~al.(2021{\natexlab{a}})Wang, Lian, Zhang, Qin, He, Lin, and
  Lin]{DBLP:journals/www/WangLZQHLL21}
Hanchen Wang, Defu Lian, Ying Zhang, Lu~Qin, Xiangjian He, Yiguang Lin, and
  Xuemin Lin.
\newblock Binarized graph neural network.
\newblock \emph{World Wide Web}, 24\penalty0 (3):\penalty0 825--848,
  2021{\natexlab{a}}.

\bibitem[Wang et~al.(2016)Wang, Wu, Xu, Xiao, and
  Yang]{DBLP:conf/ics/WangWXXY16}
Linnan Wang, Wei Wu, Zenglin Xu, Jianxiong Xiao, and Yi~Yang.
\newblock {BLASX:} {A} high performance level-3 {BLAS} library for
  heterogeneous multi-gpu computing.
\newblock In \emph{{ICS}}, pp.\  20:1--20:11. {ACM}, 2016.

\bibitem[Wang et~al.(2018{\natexlab{a}})Wang, Ye, Zhao, Wu, Li, Song, Xu, and
  Kraska]{DBLP:conf/ppopp/WangYZWLSXK18}
Linnan Wang, Jinmian Ye, Yiyang Zhao, Wei Wu, Ang Li, Shuaiwen~Leon Song,
  Zenglin Xu, and Tim Kraska.
\newblock Superneurons: dynamic {GPU} memory management for training deep
  neural networks.
\newblock In \emph{{PPOPP}}, pp.\  41--53. {ACM}, 2018{\natexlab{a}}.

\bibitem[Wang et~al.(2017)Wang, Liu, and Foroosh]{DBLP:conf/iccvw/WangLF17}
Min Wang, Baoyuan Liu, and Hassan Foroosh.
\newblock Factorized convolutional neural networks.
\newblock In \emph{{ICCV} Workshops}, pp.\  545--553. {IEEE} Computer Society,
  2017.

\bibitem[Wang et~al.(2018{\natexlab{b}})Wang, Xie, Deng, Li, Wang, and
  Xie]{DBLP:conf/nips/WangXDLWX18}
Peiqi Wang, Xinfeng Xie, Lei Deng, Guoqi Li, Dongsheng Wang, and Yuan Xie.
\newblock Hitnet: Hybrid ternary recurrent neural network.
\newblock In Samy Bengio, Hanna~M. Wallach, Hugo Larochelle, Kristen Grauman,
  Nicol{\`{o}} Cesa{-}Bianchi, and Roman Garnett (eds.), \emph{Advances in
  Neural Information Processing Systems 31: Annual Conference on Neural
  Information Processing Systems 2018, NeurIPS 2018, December 3-8, 2018,
  Montr{\'{e}}al, Canada}, pp.\  602--612, 2018{\natexlab{b}}.

\bibitem[Wang et~al.(2019{\natexlab{b}})Wang, Yuan, Zhang, and
  Feng]{DBLP:conf/cvpr/WangYZF19}
Tao Wang, Li~Yuan, Xiaopeng Zhang, and Jiashi Feng.
\newblock Distilling object detectors with fine-grained feature imitation.
\newblock In \emph{{IEEE} Conference on Computer Vision and Pattern
  Recognition, {CVPR} 2019, Long Beach, CA, USA, June 16-20, 2019}, pp.\
  4933--4942. Computer Vision Foundation / {IEEE}, 2019{\natexlab{b}}.

\bibitem[Wang \& Isola(2020)Wang and Isola]{DBLP:conf/icml/0001I20}
Tongzhou Wang and Phillip Isola.
\newblock Understanding contrastive representation learning through alignment
  and uniformity on the hypersphere.
\newblock In \emph{Proceedings of the 37th International Conference on Machine
  Learning, {ICML} 2020, 13-18 July 2020, Virtual Event}, volume 119 of
  \emph{Proceedings of Machine Learning Research}, pp.\  9929--9939. {PMLR},
  2020.

\bibitem[Wang et~al.(2020{\natexlab{b}})Wang, Bi, Yan, Wu, Xia, Bao, Peng, and
  Si]{DBLP:conf/iclr/0225BYWXBPS20}
Wei Wang, Bin Bi, Ming Yan, Chen Wu, Jiangnan Xia, Zuyi Bao, Liwei Peng, and
  Luo Si.
\newblock Structbert: Incorporating language structures into pre-training for
  deep language understanding.
\newblock In \emph{8th International Conference on Learning Representations,
  {ICLR} 2020, Addis Ababa, Ethiopia, April 26-30, 2020}. OpenReview.net,
  2020{\natexlab{b}}.

\bibitem[Wang et~al.(2021{\natexlab{b}})Wang, Xiong, Wei, Wang, and
  Li]{wang2021lightseq}
Xiaohui Wang, Ying Xiong, Yang Wei, Mingxuan Wang, and Lei Li.
\newblock {L}ight{S}eq: A high performance inference library for transformers.
\newblock In \emph{Proceedings of the 2021 Conference of the North American
  Chapter of the Association for Computational Linguistics: Human Language
  Technologies: Industry Papers (NAACL-HLT)}, pp.\  113--120. Association for
  Computational Linguistics, June 2021{\natexlab{b}}.

\bibitem[Wang et~al.(2018{\natexlab{c}})Wang, Luo, Crankshaw, Tumanov, Yu, and
  Gonzalez]{DBLP:conf/uai/WangLCTYG18}
Xin Wang, Yujia Luo, Daniel Crankshaw, Alexey Tumanov, Fisher Yu, and Joseph~E.
  Gonzalez.
\newblock {IDK} cascades: Fast deep learning by learning not to overthink.
\newblock In Amir Globerson and Ricardo Silva (eds.), \emph{Proceedings of the
  Thirty-Fourth Conference on Uncertainty in Artificial Intelligence, {UAI}
  2018, Monterey, California, USA, August 6-10, 2018}, pp.\  580--590. {AUAI}
  Press, 2018{\natexlab{c}}.

\bibitem[Wang et~al.(2018{\natexlab{d}})Wang, Yu, Dou, Darrell, and
  Gonzalez]{DBLP:conf/eccv/WangYDDG18}
Xin Wang, Fisher Yu, Zi{-}Yi Dou, Trevor Darrell, and Joseph~E. Gonzalez.
\newblock Skipnet: Learning dynamic routing in convolutional networks.
\newblock In Vittorio Ferrari, Martial Hebert, Cristian Sminchisescu, and Yair
  Weiss (eds.), \emph{Computer Vision - {ECCV} 2018 - 15th European Conference,
  Munich, Germany, September 8-14, 2018, Proceedings, Part {XIII}}, volume
  11217 of \emph{Lecture Notes in Computer Science}, pp.\  420--436. Springer,
  2018{\natexlab{d}}.

\bibitem[Wang et~al.(2020{\natexlab{c}})Wang, Wang, Li, and
  Tu]{DBLP:conf/emnlp/WangWLT20}
Yong Wang, Longyue Wang, Victor O.~K. Li, and Zhaopeng Tu.
\newblock On the sparsity of neural machine translation models.
\newblock In \emph{Proceedings of the 2020 Conference on Empirical Methods in
  Natural Language Processing, {EMNLP} 2020, Online, November 16-20, 2020},
  pp.\  1060--1066, 2020{\natexlab{c}}.

\bibitem[Wang et~al.(2020{\natexlab{d}})Wang, Lv, Huang, Song, Yang, and
  Huang]{DBLP:conf/nips/WangLHSYH20}
Yulin Wang, Kangchen Lv, Rui Huang, Shiji Song, Le~Yang, and Gao Huang.
\newblock Glance and focus: a dynamic approach to reducing spatial redundancy
  in image classification.
\newblock In Hugo Larochelle, Marc'Aurelio Ranzato, Raia Hadsell,
  Maria{-}Florina Balcan, and Hsuan{-}Tien Lin (eds.), \emph{Advances in Neural
  Information Processing Systems 33: Annual Conference on Neural Information
  Processing Systems 2020, NeurIPS 2020, December 6-12, 2020, virtual},
  2020{\natexlab{d}}.

\bibitem[Wang et~al.(2021{\natexlab{c}})Wang, Huang, Song, Huang, and
  Huang]{DBLP:journals/corr/abs-2105-15075}
Yulin Wang, Rui Huang, Shiji Song, Zeyi Huang, and Gao Huang.
\newblock Not all images are worth 16x16 words: Dynamic vision transformers
  with adaptive sequence length.
\newblock \emph{CoRR}, abs/2105.15075, 2021{\natexlab{c}}.

\bibitem[Warstadt et~al.(2019)Warstadt, Singh, and
  Bowman]{DBLP:journals/tacl/WarstadtSB19}
Alex Warstadt, Amanpreet Singh, and Samuel~R. Bowman.
\newblock Neural network acceptability judgments.
\newblock \emph{Trans. Assoc. Comput. Linguistics}, 7:\penalty0 625--641, 2019.

\bibitem[Wei et~al.(2021)Wei, Bosma, Zhao, Guu, Yu, Lester, Du, Dai, and
  Le]{DBLP:journals/corr/abs-2109-01652}
Jason Wei, Maarten Bosma, Vincent~Y. Zhao, Kelvin Guu, Adams~Wei Yu, Brian
  Lester, Nan Du, Andrew~M. Dai, and Quoc~V. Le.
\newblock Finetuned language models are zero-shot learners.
\newblock \emph{CoRR}, abs/2109.01652, 2021.

\bibitem[Williams et~al.(2018)Williams, Nangia, and
  Bowman]{DBLP:conf/naacl/WilliamsNB18}
Adina Williams, Nikita Nangia, and Samuel~R. Bowman.
\newblock A broad-coverage challenge corpus for sentence understanding through
  inference.
\newblock In Marilyn~A. Walker, Heng Ji, and Amanda Stent (eds.),
  \emph{Proceedings of the 2018 Conference of the North American Chapter of the
  Association for Computational Linguistics: Human Language Technologies,
  {NAACL-HLT} 2018, New Orleans, Louisiana, USA, June 1-6, 2018, Volume 1 (Long
  Papers)}, pp.\  1112--1122. Association for Computational Linguistics, 2018.

\bibitem[Winata et~al.(2019)Winata, Madotto, Shin, Barezi, and
  Fung]{DBLP:journals/corr/abs-1908-09982}
Genta~Indra Winata, Andrea Madotto, Jamin Shin, Elham~J. Barezi, and Pascale
  Fung.
\newblock On the effectiveness of low-rank matrix factorization for {LSTM}
  model compression.
\newblock \emph{CoRR}, abs/1908.09982, 2019.

\bibitem[Wu et~al.(2019)Wu, Dai, Zhang, Wang, Sun, Wu, Tian, Vajda, Jia, and
  Keutzer]{DBLP:conf/cvpr/WuDZWSWTVJK19}
Bichen Wu, Xiaoliang Dai, Peizhao Zhang, Yanghan Wang, Fei Sun, Yiming Wu,
  Yuandong Tian, Peter Vajda, Yangqing Jia, and Kurt Keutzer.
\newblock Fbnet: Hardware-aware efficient convnet design via differentiable
  neural architecture search.
\newblock In \emph{{CVPR}}, pp.\  10734--10742. Computer Vision Foundation /
  {IEEE}, 2019.

\bibitem[Wu et~al.(2018{\natexlab{a}})Wu, Li, Chen, and
  Shi]{DBLP:conf/iclr/WuLCS18}
Shuang Wu, Guoqi Li, Feng Chen, and Luping Shi.
\newblock Training and inference with integers in deep neural networks.
\newblock In \emph{6th International Conference on Learning Representations,
  {ICLR} 2018, Vancouver, BC, Canada, April 30 - May 3, 2018, Conference Track
  Proceedings}. OpenReview.net, 2018{\natexlab{a}}.

\bibitem[Wu \& He(2020)Wu and He]{DBLP:journals/ijcv/WuH20}
Yuxin Wu and Kaiming He.
\newblock Group normalization.
\newblock \emph{Int. J. Comput. Vis.}, 128\penalty0 (3):\penalty0 742--755,
  2020.

\bibitem[Wu et~al.(2020)Wu, Wang, Gu, Khabsa, Sun, and
  Ma]{DBLP:journals/corr/abs-2012-15466}
Zhuofeng Wu, Sinong Wang, Jiatao Gu, Madian Khabsa, Fei Sun, and Hao Ma.
\newblock {CLEAR:} contrastive learning for sentence representation.
\newblock \emph{CoRR}, abs/2012.15466, 2020.

\bibitem[Wu et~al.(2018{\natexlab{b}})Wu, Nagarajan, Kumar, Rennie, Davis,
  Grauman, and Feris]{DBLP:conf/cvpr/WuNKRDGF18}
Zuxuan Wu, Tushar Nagarajan, Abhishek Kumar, Steven Rennie, Larry~S. Davis,
  Kristen Grauman, and Rog{\'{e}}rio~Schmidt Feris.
\newblock Blockdrop: Dynamic inference paths in residual networks.
\newblock In \emph{2018 {IEEE} Conference on Computer Vision and Pattern
  Recognition, {CVPR} 2018, Salt Lake City, UT, USA, June 18-22, 2018}, pp.\
  8817--8826. Computer Vision Foundation / {IEEE} Computer Society,
  2018{\natexlab{b}}.

\bibitem[Xiang et~al.(2020)Xiang, Ding, and Han]{DBLP:conf/eccv/XiangDH20}
Liuyu Xiang, Guiguang Ding, and Jungong Han.
\newblock Learning from multiple experts: Self-paced knowledge distillation for
  long-tailed classification.
\newblock In Andrea Vedaldi, Horst Bischof, Thomas Brox, and Jan{-}Michael
  Frahm (eds.), \emph{Computer Vision - {ECCV} 2020 - 16th European Conference,
  Glasgow, UK, August 23-28, 2020, Proceedings, Part {V}}, volume 12350 of
  \emph{Lecture Notes in Computer Science}, pp.\  247--263. Springer, 2020.

\bibitem[Xin et~al.(2020)Xin, Tang, Lee, Yu, and Lin]{DBLP:conf/acl/XinTLYL20}
Ji~Xin, Raphael Tang, Jaejun Lee, Yaoliang Yu, and Jimmy Lin.
\newblock Deebert: Dynamic early exiting for accelerating {BERT} inference.
\newblock In Dan Jurafsky, Joyce Chai, Natalie Schluter, and Joel~R. Tetreault
  (eds.), \emph{Proceedings of the 58th Annual Meeting of the Association for
  Computational Linguistics, {ACL} 2020, Online, July 5-10, 2020}, pp.\
  2246--2251. Association for Computational Linguistics, 2020.

\bibitem[Xin et~al.(2021)Xin, Tang, Yu, and Lin]{DBLP:conf/eacl/XinTYL21}
Ji~Xin, Raphael Tang, Yaoliang Yu, and Jimmy Lin.
\newblock Berxit: Early exiting for {BERT} with better fine-tuning and
  extension to regression.
\newblock In Paola Merlo, J{\"{o}}rg Tiedemann, and Reut Tsarfaty (eds.),
  \emph{Proceedings of the 16th Conference of the European Chapter of the
  Association for Computational Linguistics: Main Volume, {EACL} 2021, Online,
  April 19 - 23, 2021}, pp.\  91--104. Association for Computational
  Linguistics, 2021.

\bibitem[Xu et~al.(2020)Xu, Zhou, Ge, Wei, and Zhou]{DBLP:conf/emnlp/XuZGWZ20}
Canwen Xu, Wangchunshu Zhou, Tao Ge, Furu Wei, and Ming Zhou.
\newblock Bert-of-theseus: Compressing {BERT} by progressive module replacing.
\newblock In Bonnie Webber, Trevor Cohn, Yulan He, and Yang Liu (eds.),
  \emph{Proceedings of the 2020 Conference on Empirical Methods in Natural
  Language Processing, {EMNLP} 2020, Online, November 16-20, 2020}, pp.\
  7859--7869. Association for Computational Linguistics, 2020.

\bibitem[Xu et~al.(2021{\natexlab{a}})Xu, Zhou, Ge, Xu, McAuley, and
  Wei]{DBLP:journals/corr/abs-2109-03228}
Canwen Xu, Wangchunshu Zhou, Tao Ge, Ke~Xu, Julian~J. McAuley, and Furu Wei.
\newblock Beyond preserved accuracy: Evaluating loyalty and robustness of
  {BERT} compression.
\newblock \emph{CoRR}, abs/2109.03228, 2021{\natexlab{a}}.

\bibitem[Xu et~al.(2019)Xu, Sun, Zhang, Zhao, and Lin]{DBLP:conf/nips/Xu0ZZL19}
Jingjing Xu, Xu~Sun, Zhiyuan Zhang, Guangxiang Zhao, and Junyang Lin.
\newblock Understanding and improving layer normalization.
\newblock In \emph{NeurIPS}, pp.\  4383--4393, 2019.

\bibitem[Xu et~al.(2021{\natexlab{b}})Xu, Zhao, Lin, Gao, Sun, and
  Yang]{DBLP:conf/icml/XuZLG0Y21}
Jingjing Xu, Liang Zhao, Junyang Lin, Rundong Gao, Xu~Sun, and Hongxia Yang.
\newblock {KNAS:} green neural architecture search.
\newblock In \emph{{ICML}}, volume 139 of \emph{Proceedings of Machine Learning
  Research}, pp.\  11613--11625. {PMLR}, 2021{\natexlab{b}}.

\bibitem[Xu et~al.(2018)Xu, Shen, Yao, Tian, and Mei]{DBLP:conf/icmcs/XuSYTM18}
Kaisheng Xu, Xu~Shen, Ting Yao, Xinmei Tian, and Tao Mei.
\newblock Greedy layer-wise training of long short term memory networks.
\newblock In \emph{2018 {IEEE} International Conference on Multimedia {\&} Expo
  Workshops, {ICME} Workshops 2018, San Diego, CA, USA, July 23-27, 2018}, pp.\
   1--6. {IEEE} Computer Society, 2018.

\bibitem[Xue et~al.(2014)Xue, Li, Yu, Seltzer, and
  Gong]{DBLP:conf/icassp/XueLYSG14}
Jian Xue, Jinyu Li, Dong Yu, Mike Seltzer, and Yifan Gong.
\newblock Singular value decomposition based low-footprint speaker adaptation
  and personalization for deep neural network.
\newblock In \emph{{IEEE} International Conference on Acoustics, Speech and
  Signal Processing, {ICASSP} 2014, Florence, Italy, May 4-9, 2014}, pp.\
  6359--6363, 2014.

\bibitem[Xue et~al.(2021)Xue, Constant, Roberts, Kale, Al{-}Rfou, Siddhant,
  Barua, and Raffel]{DBLP:conf/naacl/XueCRKASBR21}
Linting Xue, Noah Constant, Adam Roberts, Mihir Kale, Rami Al{-}Rfou, Aditya
  Siddhant, Aditya Barua, and Colin Raffel.
\newblock mt5: {A} massively multilingual pre-trained text-to-text transformer.
\newblock In Kristina Toutanova, Anna Rumshisky, Luke Zettlemoyer, Dilek
  Hakkani{-}T{\"{u}}r, Iz~Beltagy, Steven Bethard, Ryan Cotterell, Tanmoy
  Chakraborty, and Yichao Zhou (eds.), \emph{Proceedings of the 2021 Conference
  of the North American Chapter of the Association for Computational
  Linguistics: Human Language Technologies, {NAACL-HLT} 2021, Online, June
  6-11, 2021}, pp.\  483--498. Association for Computational Linguistics, 2021.

\bibitem[Yang et~al.(2020{\natexlab{a}})Yang, Wang, Yang, Li, He, and
  Zhang]{DBLP:journals/corr/abs-2011-13635}
Cheng Yang, Shengnan Wang, Chao Yang, Yuechuan Li, Ru~He, and Jingqiao Zhang.
\newblock Progressively stacking 2.0: {A} multi-stage layerwise training method
  for {BERT} training speedup.
\newblock \emph{CoRR}, abs/2011.13635, 2020{\natexlab{a}}.

\bibitem[Yang et~al.(2020{\natexlab{b}})Yang, Han, Chen, Song, Dai, and
  Huang]{DBLP:conf/cvpr/YangHCSDH20}
Le~Yang, Yizeng Han, Xi~Chen, Shiji Song, Jifeng Dai, and Gao Huang.
\newblock Resolution adaptive networks for efficient inference.
\newblock In \emph{2020 {IEEE/CVF} Conference on Computer Vision and Pattern
  Recognition, {CVPR} 2020, Seattle, WA, USA, June 13-19, 2020}, pp.\
  2366--2375. Computer Vision Foundation / {IEEE}, 2020{\natexlab{b}}.

\bibitem[Yang et~al.(2017)Yang, Salakhutdinov, and
  Cohen]{DBLP:conf/iclr/YangSC17}
Zhilin Yang, Ruslan Salakhutdinov, and William~W. Cohen.
\newblock Transfer learning for sequence tagging with hierarchical recurrent
  networks.
\newblock In \emph{{ICLR} (Poster)}. OpenReview.net, 2017.

\bibitem[Yang et~al.(2019)Yang, Dai, Yang, Carbonell, Salakhutdinov, and
  Le]{DBLP:conf/nips/YangDYCSL19}
Zhilin Yang, Zihang Dai, Yiming Yang, Jaime~G. Carbonell, Ruslan Salakhutdinov,
  and Quoc~V. Le.
\newblock Xlnet: Generalized autoregressive pretraining for language
  understanding.
\newblock In \emph{Advances in Neural Information Processing Systems 32: Annual
  Conference on Neural Information Processing Systems 2019, NeurIPS 2019,
  December 8-14, 2019, Vancouver, BC, Canada}, pp.\  5754--5764, 2019.

\bibitem[Yim et~al.(2017)Yim, Joo, Bae, and Kim]{DBLP:conf/cvpr/YimJBK17}
Junho Yim, Donggyu Joo, Ji{-}Hoon Bae, and Junmo Kim.
\newblock A gift from knowledge distillation: Fast optimization, network
  minimization and transfer learning.
\newblock In \emph{2017 {IEEE} Conference on Computer Vision and Pattern
  Recognition, {CVPR} 2017, Honolulu, HI, USA, July 21-26, 2017}, pp.\
  7130--7138. {IEEE} Computer Society, 2017.

\bibitem[Ying et~al.(2019)Ying, Klein, Christiansen, Real, Murphy, and
  Hutter]{DBLP:conf/icml/YingKCR0H19}
Chris Ying, Aaron Klein, Eric Christiansen, Esteban Real, Kevin Murphy, and
  Frank Hutter.
\newblock Nas-bench-101: Towards reproducible neural architecture search.
\newblock In \emph{Proceedings of the 36th International Conference on Machine
  Learning, {ICML} 2019, 9-15 June 2019, Long Beach, California, {USA}}, pp.\
  7105--7114, 2019.

\bibitem[Yoo \& Kweon(2019)Yoo and Kweon]{DBLP:conf/cvpr/YooK19}
Donggeun Yoo and In~So Kweon.
\newblock Learning loss for active learning.
\newblock In \emph{{CVPR}}, pp.\  93--102. Computer Vision Foundation / {IEEE},
  2019.

\bibitem[Yosinski et~al.(2014)Yosinski, Clune, Bengio, and
  Lipson]{DBLP:conf/nips/YosinskiCBL14}
Jason Yosinski, Jeff Clune, Yoshua Bengio, and Hod Lipson.
\newblock How transferable are features in deep neural networks?
\newblock In \emph{{NIPS}}, pp.\  3320--3328, 2014.

\bibitem[You et~al.(2017)You, Xu, Xu, and Tao]{DBLP:conf/kdd/YouX0T17}
Shan You, Chang Xu, Chao Xu, and Dacheng Tao.
\newblock Learning from multiple teacher networks.
\newblock In \emph{Proceedings of the 23rd {ACM} {SIGKDD} International
  Conference on Knowledge Discovery and Data Mining, Halifax, NS, Canada,
  August 13 - 17, 2017}, pp.\  1285--1294. {ACM}, 2017.

\bibitem[Yu et~al.(2017)Yu, Lee, and Le]{DBLP:conf/acl/YuLL17}
Adams~Wei Yu, Hongrae Lee, and Quoc~V. Le.
\newblock Learning to skim text.
\newblock In Regina Barzilay and Min{-}Yen Kan (eds.), \emph{Proceedings of the
  55th Annual Meeting of the Association for Computational Linguistics, {ACL}
  2017, Vancouver, Canada, July 30 - August 4, Volume 1: Long Papers}, pp.\
  1880--1890. Association for Computational Linguistics, 2017.

\bibitem[Yu \& Huang(2019)Yu and Huang]{DBLP:conf/iccv/YuH19}
Jiahui Yu and Thomas~S. Huang.
\newblock Universally slimmable networks and improved training techniques.
\newblock In \emph{2019 {IEEE/CVF} International Conference on Computer Vision,
  {ICCV} 2019, Seoul, Korea (South), October 27 - November 2, 2019}, pp.\
  1803--1811. {IEEE}, 2019.

\bibitem[Yu et~al.(2019)Yu, Yang, Xu, Yang, and Huang]{DBLP:conf/iclr/YuYXYH19}
Jiahui Yu, Linjie Yang, Ning Xu, Jianchao Yang, and Thomas~S. Huang.
\newblock Slimmable neural networks.
\newblock In \emph{7th International Conference on Learning Representations,
  {ICLR} 2019, New Orleans, LA, USA, May 6-9, 2019}. OpenReview.net, 2019.

\bibitem[Yu et~al.(2018)Yu, Liu, Schwing, and Peng]{DBLP:conf/iclr/Yu0S018}
Keyi Yu, Yang Liu, Alexander~G. Schwing, and Jian Peng.
\newblock Fast and accurate text classification: Skimming, rereading and early
  stopping.
\newblock In \emph{6th International Conference on Learning Representations,
  {ICLR} 2018, Vancouver, BC, Canada, April 30 - May 3, 2018, Workshop Track
  Proceedings}. OpenReview.net, 2018.

\bibitem[Yu \& Zhu(2020)Yu and Zhu]{DBLP:journals/corr/abs-2003-05689}
Tong Yu and Hong Zhu.
\newblock Hyper-parameter optimization: {A} review of algorithms and
  applications.
\newblock \emph{CoRR}, abs/2003.05689, 2020.

\bibitem[Zafrir et~al.(2019)Zafrir, Boudoukh, Izsak, and
  Wasserblat]{DBLP:journals/corr/abs-1910-06188}
Ofir Zafrir, Guy Boudoukh, Peter Izsak, and Moshe Wasserblat.
\newblock {Q8BERT:} quantized 8bit {BERT}.
\newblock \emph{CoRR}, abs/1910.06188, 2019.

\bibitem[Zagoruyko \& Komodakis(2017)Zagoruyko and
  Komodakis]{DBLP:conf/iclr/ZagoruykoK17}
Sergey Zagoruyko and Nikos Komodakis.
\newblock Paying more attention to attention: Improving the performance of
  convolutional neural networks via attention transfer.
\newblock In \emph{5th International Conference on Learning Representations,
  {ICLR} 2017, Toulon, France, April 24-26, 2017, Conference Track
  Proceedings}. OpenReview.net, 2017.

\bibitem[Zaheer et~al.(2020)Zaheer, Guruganesh, Dubey, Ainslie, Alberti,
  Onta{\~{n}}{\'{o}}n, Pham, Ravula, Wang, Yang, and
  Ahmed]{DBLP:conf/nips/ZaheerGDAAOPRWY20}
Manzil Zaheer, Guru Guruganesh, Kumar~Avinava Dubey, Joshua Ainslie, Chris
  Alberti, Santiago Onta{\~{n}}{\'{o}}n, Philip Pham, Anirudh Ravula, Qifan
  Wang, Li~Yang, and Amr Ahmed.
\newblock Big bird: Transformers for longer sequences.
\newblock In \emph{NeurIPS}, 2020.

\bibitem[Zeiler \& Fergus(2014)Zeiler and Fergus]{DBLP:conf/eccv/ZeilerF14}
Matthew~D. Zeiler and Rob Fergus.
\newblock Visualizing and understanding convolutional networks.
\newblock In \emph{{ECCV} {(1)}}, volume 8689 of \emph{Lecture Notes in
  Computer Science}, pp.\  818--833. Springer, 2014.

\bibitem[Zela et~al.(2020)Zela, Elsken, Saikia, Marrakchi, Brox, and
  Hutter]{DBLP:conf/iclr/ZelaESMBH20}
Arber Zela, Thomas Elsken, Tonmoy Saikia, Yassine Marrakchi, Thomas Brox, and
  Frank Hutter.
\newblock Understanding and robustifying differentiable architecture search.
\newblock In \emph{8th International Conference on Learning Representations,
  {ICLR} 2020, Addis Ababa, Ethiopia, April 26-30, 2020}, 2020.

\bibitem[Zhang et~al.(2019)Zhang, Dauphin, and Ma]{DBLP:conf/iclr/ZhangDM19}
Hongyi Zhang, Yann~N. Dauphin, and Tengyu Ma.
\newblock Fixup initialization: Residual learning without normalization.
\newblock In \emph{{ICLR} (Poster)}. OpenReview.net, 2019.

\bibitem[Zhang \& Stadie(2020)Zhang and Stadie]{DBLP:conf/iclr/ZhangS20}
Matthew~Shunshi Zhang and Bradly~C. Stadie.
\newblock One-shot pruning of recurrent neural networks by jacobian spectrum
  evaluation.
\newblock In \emph{8th International Conference on Learning Representations,
  {ICLR} 2020, Addis Ababa, Ethiopia, April 26-30, 2020}, 2020.

\bibitem[Zhang et~al.(2020)Zhang, Hou, Yin, Shang, Chen, Jiang, and
  Liu]{DBLP:conf/emnlp/ZhangHYSCJL20}
Wei Zhang, Lu~Hou, Yichun Yin, Lifeng Shang, Xiao Chen, Xin Jiang, and Qun Liu.
\newblock Ternarybert: Distillation-aware ultra-low bit {BERT}.
\newblock In Bonnie Webber, Trevor Cohn, Yulan He, and Yang Liu (eds.),
  \emph{Proceedings of the 2020 Conference on Empirical Methods in Natural
  Language Processing, {EMNLP} 2020, Online, November 16-20, 2020}, pp.\
  509--521. Association for Computational Linguistics, 2020.

\bibitem[Zhang et~al.(2018)Zhang, Xiang, Hospedales, and
  Lu]{DBLP:conf/cvpr/ZhangXHL18}
Ying Zhang, Tao Xiang, Timothy~M. Hospedales, and Huchuan Lu.
\newblock Deep mutual learning.
\newblock In \emph{2018 {IEEE} Conference on Computer Vision and Pattern
  Recognition, {CVPR} 2018, Salt Lake City, UT, USA, June 18-22, 2018}, pp.\
  4320--4328. {IEEE} Computer Society, 2018.

\bibitem[Zhang et~al.(2017)Zhang, Ning, and
  He]{DBLP:journals/corr/abs-1710-09505}
Zhi Zhang, Guanghan Ning, and Zhihai He.
\newblock Knowledge projection for deep neural networks.
\newblock \emph{CoRR}, abs/1710.09505, 2017.

\bibitem[Zhao et~al.(2020)Zhao, Wang, Bates, Mullins, Jamnik, and
  Li{\`{o}}]{DBLP:journals/corr/abs-2009-09232}
Yiren Zhao, Duo Wang, Daniel Bates, Robert~D. Mullins, Mateja Jamnik, and
  Pietro Li{\`{o}}.
\newblock Learned low precision graph neural networks.
\newblock \emph{CoRR}, abs/2009.09232, 2020.

\bibitem[Zheng et~al.(2015)Zheng, Jayasumana, Romera{-}Paredes, Vineet, Su, Du,
  Huang, and Torr]{DBLP:conf/iccv/0001JRVSDHT15}
Shuai Zheng, Sadeep Jayasumana, Bernardino Romera{-}Paredes, Vibhav Vineet,
  Zhizhong Su, Dalong Du, Chang Huang, and Philip H.~S. Torr.
\newblock Conditional random fields as recurrent neural networks.
\newblock In \emph{2015 {IEEE} International Conference on Computer Vision,
  {ICCV} 2015, Santiago, Chile, December 7-13, 2015}, pp.\  1529--1537, 2015.

\bibitem[Zhou et~al.(2019)Zhou, Liu, Long, Chen, and
  Zhu]{DBLP:journals/spic/ZhouLLCZ19}
Mingyi Zhou, Yipeng Liu, Zhen Long, Longxi Chen, and Ce~Zhu.
\newblock Tensor rank learning in {CP} decomposition via convolutional neural
  network.
\newblock \emph{Signal Process. Image Commun.}, 73:\penalty0 12--21, 2019.

\bibitem[Zhou et~al.(2020)Zhou, Xu, Ge, McAuley, Xu, and
  Wei]{DBLP:conf/nips/ZhouXGM0W20}
Wangchunshu Zhou, Canwen Xu, Tao Ge, Julian~J. McAuley, Ke~Xu, and Furu Wei.
\newblock {BERT} loses patience: Fast and robust inference with early exit.
\newblock In Hugo Larochelle, Marc'Aurelio Ranzato, Raia Hadsell,
  Maria{-}Florina Balcan, and Hsuan{-}Tien Lin (eds.), \emph{Advances in Neural
  Information Processing Systems 33: Annual Conference on Neural Information
  Processing Systems 2020, NeurIPS 2020, December 6-12, 2020, virtual}, 2020.

\bibitem[Zhou et~al.(2021{\natexlab{a}})Zhou, Ge, Xu, and
  Wei]{DBLP:journals/corr/abs-2101-00416}
Wangchunshu Zhou, Tao Ge, Ke~Xu, and Furu Wei.
\newblock Improving sequence-to-sequence pre-training via sequence span
  rewriting.
\newblock \emph{CoRR}, abs/2101.00416, 2021{\natexlab{a}}.

\bibitem[Zhou et~al.(2021{\natexlab{b}})Zhou, Xu, and
  McAuley]{DBLP:journals/corr/abs-2106-04570}
Wangchunshu Zhou, Canwen Xu, and Julian~J. McAuley.
\newblock Meta learning for knowledge distillation.
\newblock \emph{CoRR}, abs/2106.04570, 2021{\natexlab{b}}.

\bibitem[Zhu(2021)]{DBLP:conf/acl/Zhu20}
Wei Zhu.
\newblock Leebert: Learned early exit for {BERT} with cross-level optimization.
\newblock In Chengqing Zong, Fei Xia, Wenjie Li, and Roberto Navigli (eds.),
  \emph{Proceedings of the 59th Annual Meeting of the Association for
  Computational Linguistics and the 11th International Joint Conference on
  Natural Language Processing, {ACL/IJCNLP} 2021, (Volume 1: Long Papers),
  Virtual Event, August 1-6, 2021}, pp.\  2968--2980. Association for
  Computational Linguistics, 2021.

\end{thebibliography}
\bibliographystyle{style/iclr2019_conference}
\clearpage
\appendix
\end{document}